\documentclass[10pt,journal,compsoc]{IEEEtran}
\ifCLASSOPTIONcompsoc
  \usepackage[nocompress]{cite}
\else
  \usepackage{cite}
\fi
\ifCLASSINFOpdf
\else
\fi

\usepackage{url}
\usepackage[square,sort,numbers]{natbib}
\usepackage{graphicx}
\usepackage{amsfonts}
\usepackage{amsmath}
\usepackage{tikz, tikz-qtree}
\usepackage{threeparttable, booktabs, multirow, makecell}
\usepackage{lineno,hyperref,color}
\modulolinenumbers[5]
\hypersetup{
    colorlinks=true,
    linkcolor=blue,
    urlcolor=blue,
    citecolor=cyan,
}

\begin{document}
\title{Graph-based Facial Affect Analysis: A Review}

\author{Yang~Liu,
        Xingming~Zhang,
        Yante~Li,
        Jinzhao~Zhou,
        Xin~Li,
        and~Guoying~Zhao,~\IEEEmembership{Fellow,~IEEE}% <-this % stops a space
\IEEEcompsocitemizethanks{
\scriptsize
\IEEEcompsocthanksitem Y. Liu, X. Zhang and J. Zhou are with the School of Computer Science and Engineering, South China University of Technology, Guangzhou 510006, China. 
E-mail: liuy17@163.com, cszxm@scut.edu.cn, charlesmzhouscut@gmail.com. 

\IEEEcompsocthanksitem X. Li is with the Department of Electrical and Computer Engineering, Rutgers University, Piscataway, NJ-08854, United States. E-mail: xin.li.ece@rutgers.edu

\IEEEcompsocthanksitem Y. Liu, Y. Li and G. Zhao are with the Center for Machine Vision and Signal Analysis, University of Oulu, Oulu, FI-90014, Finland. E-mail: \{firstname.lastname\}@oulu.fi}% <-this % stops an unwanted space

% \thanks{Manuscript received XX XX, 20XX; revised XX XX, 20XX.}
}


\markboth{Journal of \LaTeX\ Class Files,~Vol.~14, No.~8, August~2015}%
{Shell \MakeLowercase{\textit{et al.}}: Bare Demo of IEEEtran.cls for Computer Society Journals}

% \IEEEpubid{0000--0000/00\$00.00~\copyright~2015 IEEE}


\IEEEtitleabstractindextext{%
\begin{abstract}
  As one of the most important affective signals, facial affect analysis (FAA) is essential for developing human-computer interaction systems. Early methods focus on extracting appearance and geometry features associated with human affects while ignoring the latent semantic information among individual facial changes, leading to limited performance and generalization. Recent work attempts to establish a graph-based representation to model these semantic relationships and develop frameworks to leverage them for various FAA tasks. This paper provides a comprehensive review of graph-based FAA, including the evolution of algorithms and their applications. First, the FAA background knowledge is introduced, especially on the role of the graph. We then discuss approaches widely used for graph-based affective representation in literature and show a trend towards graph construction. For the relational reasoning in graph-based FAA, existing studies are categorized according to their non-deep or deep learning methods, emphasizing the latest graph neural networks. Performance comparisons of the state-of-the-art graph-based FAA methods are also summarized. Finally, we discuss the challenges and potential directions. As far as we know, this is the first survey of graph-based FAA methods. Our findings can serve as a reference for future research in this field.
\end{abstract}

}

\maketitle

\IEEEdisplaynontitleabstractindextext

% \fi
\IEEEpeerreviewmaketitle

\IEEEraisesectionheading{\section{Introduction}\label{sec:intro}}

\IEEEPARstart{F}{acial} affects relate to $55\%$ of messages when people perceive others' feelings and attitudes \cite{mehrabian1971silent,mehrabian1974approach} because it conveys critical information that reflects emotional states and reactions in human communications \cite{darwin1998expression,gazzaniga2014cognitive,picard2001toward}. Many facial affect analysis (FAA) methods have been explored during the past decade, benefiting from interdisciplinary studies of affective computing, computer vision, and psychology \cite{jack2012facial,sariyanidi2014automatic,calvo2018makes}. Some have been extended to many applications, including medical diagnosis \cite{tavakolian2019spatiotemporal}, social media \cite{zhang2018facial2}, and video generation \cite{zhao2020towards}. Meanwhile, competitions such as FERA \cite{valstar2017fera}, EmotiW \cite{dhall2020emotiw}, Aff-Wild \cite{zafeiriou2017aff}, ABAW \cite{kollias2021analysing}, EmotioNet \cite{benitez2017emotionet}, AVEC \cite{ringeval2019avec}, and MuSe \cite{stappen2020muse} are regularly held to evaluate the latest progress and propose frontier research trends. 

Historically, FAA methods have undergone a series of evolutions. Initial studies usually rely on hand-crafted design or classic machine learning to obtain useful affective features without structural information \cite{zhao2007dynamic,sariyanidi2014automatic}. Psychological findings indicate that the human cognition of facial information is realized through a dual system composed of analytic processing and holistic processing \cite{gazzaniga2014cognitive}. The former acquires multi-dimensional cluster features by analyzing local areas, while the latter aims to generate a holistic representation to perceive the overall structure \cite{gazzaniga2014cognitive,friesen1978facial}. Such an analytic-holistic working system is similar to a topology-like structure, so it is reasonable for machine vision researchers to model it into a graph. Accordingly, many state-of-the-art studies have been dedicated to generating a facial graph with local-to-global affective features \cite{zhong2019graph,li2019semantic,zhang2020region,song2021dynamic}. 

If the above evidence reveals the feasibility of using graph-based methods for FAA, research on how facial muscles participate in affective expression further demonstrates its possibility as a necessary condition \cite{ekman1994strong,cohn2015automated,zarins2015anatomy}. There are latent relationships among different facial areas and contexts, which are vital clues \cite{bimler2006facial,ekman2002facial}. A few non-graph-based deep models have partly captured these relationships and improved performance \cite{zhang2017facial,li2020facial,jacob2021facial}. The underlying assumption is that explicit mappings that reflect this relationship can be directly learned \cite{martinez2017automatic}. However, these mappings are not solid enough in the real world because they differ from subject to subject and even from one condition to another \cite{barrett2017emotions,liu2019facial}. Recently, graph-based methods have shown that they represent facial anatomy and simultaneously fit latent relationships in facial affects \cite{cui2020knowledge,fan2020facial,xie2020assisted}. Some pilot studies have also suggested that the graph-based method can even move beyond to deal with challenging tasks such as analyzing occluded faces \cite{dapogny2018confidence,zhou2020learning} and ambiguous facial affects \cite{cui2020label,chen2020label}. 

\begin{figure}[tb]
  \centering
  \includegraphics[width=0.9\columnwidth]{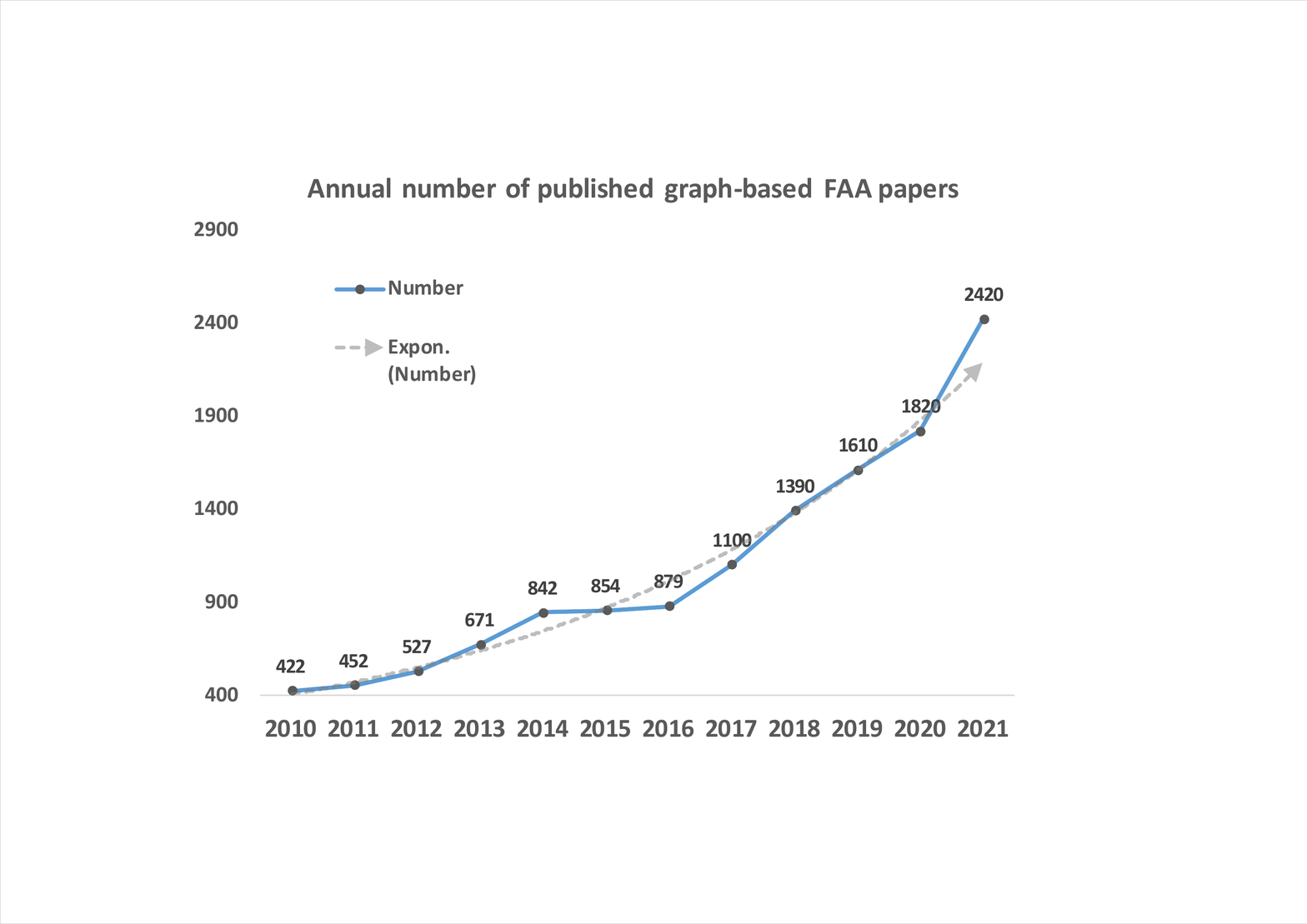}
  \caption{The growth trend of papers related to graph-based FAA.}
  \label{fig:num}\vspace{-10pt}
\end{figure}

By searching on Google Scholar using keywords of \textbf{'graph'} and \textbf{Index Terms} in this survey, we have counted the number of relevant published papers from 2010 to the present. As presented in Fig. \ref{fig:num}, the graph-based FAA has gained increasing attention, especially in the past five years (publications in 2021 increased by 600 year-on-year). 

Based on theoretical support, outstanding performance and quantity of existing work, and potential for future development, it is necessary to review the state of graph-based FAA methods. Although many reviews have discussed FFA's historical evolutions \cite{sariyanidi2014automatic, corneanu2016survey, martinez2017automatic} and recent advances \cite{kollias2019deep, li2020deep, goh2020micro}, including some on specific problems like occluded expression \cite{zhang2018facial}, multi-modal affect \cite{rouast2019deep} and micro-expression \cite{ben2021video}, \textbf{this is the FIRST systematic and in-depth survey for the graph-based FAA field} as far as we know. We emphasize representative research proposed after 2010. The goal is to present a novel perspective on FAA and its latest trends.

This review is organized as follows: Section \ref{sec:bk} provides a brief background on FAA and discusses the unique role of graph-based methods in FAA research. Section \ref{sec:gbar} presents a taxonomy of mainstream graph-based methods for affective representation. Section \ref{sec:agrr} reviews classical and advanced approaches for graph relational reasoning and discusses their pros and cons in FAA tasks. Section \ref{sec:appe} summarizes public databases, main FAA applications, and current challenges based on a detailed comparison of related literature. Finally, Section \ref{sec:od} concludes with a general discussion and identifies potential directions.

\section{Facial Affect Analysis}\label{sec:bk}

\subsection{Affective Desription Model}
\begin{figure}[tb]
  \centering
  \includegraphics[width=0.95\columnwidth]{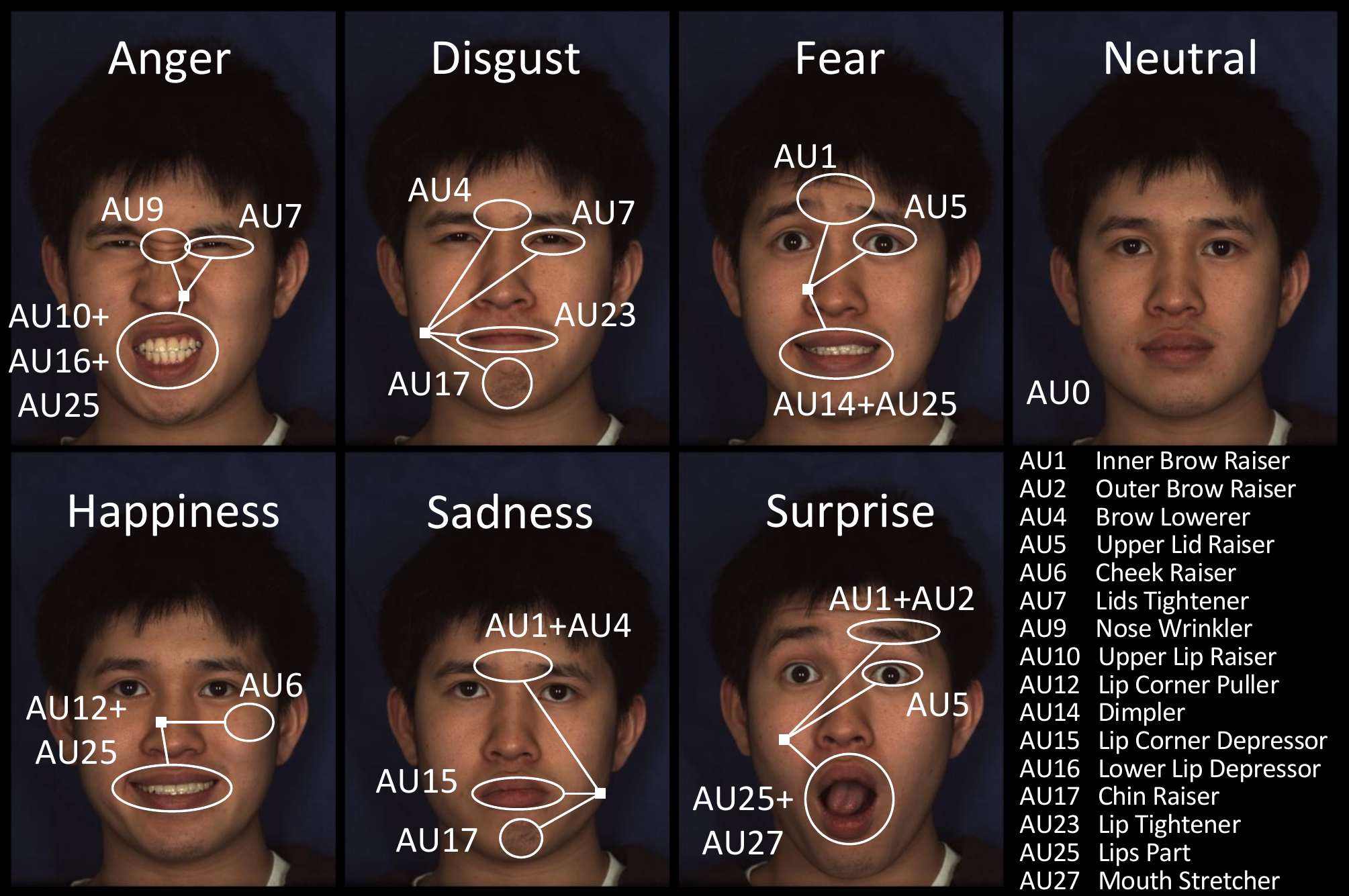}
  \caption{An example of basic facial affects and related AU marks. AU0 denotes no activated AU. All annotations are made by FACS certificated experts. Images are from BU-4DFE \cite{yin2008high} database.}
  \label{fig:fa}
  \vspace{-10pt}
\end{figure}

As early as the 1970s, \citet{ekman1971constants} proposed the definition of six basic affects, i.e., \emph{happiness}, \emph{sadness}, \emph{fear}, \emph{anger}, \emph{disgust}, and \emph{surprise}, based on an assumption of the universality of human affective display \cite{ekman1994strong}. In addition, compound affects \cite{du2014compound} defined by different combinations of basic affects (e.g., \emph{sadly surprise} and \emph{happily surprise}) are proposed to depict more complex affective situations \cite{sethu2019ambiguous}. Another kind of famous description, called Facial Action Coding System (FACS), is designed for a broader range of affects, which consists of a set of atomic Action Units (AUs) \cite{friesen1978facial, ekman2002facial}. Fig. \ref{fig:fa} shows an example of six basic affects plus \emph{neutral} and activated AUs in each facial affect. Besides categorical models, a continuous affective model named VAD Emotional State Model \cite{mehrabian1995framework} is also suggested \cite{plutchik1980general, greenwald1989affective, russell1978evidence}. The VAD model has three dimensions, i.e., valence (how positive or negative an affect is), arousal (the activation intensity of an affect), and dominance (how submissive or in-control a person is in an affective display). Recent studies consider that the continuous model is more appropriate for describing dynamic changes of human affects in the real-world \cite{soleymani2015analysis, kollias2019deep}. Please refer to \cite{sariyanidi2014automatic,corneanu2016survey} for a more detailed discussion about this topic.

\subsection{General Pipeline}
\begin{figure*}[!tb]
  \centering
  \includegraphics[width=1.8\columnwidth]{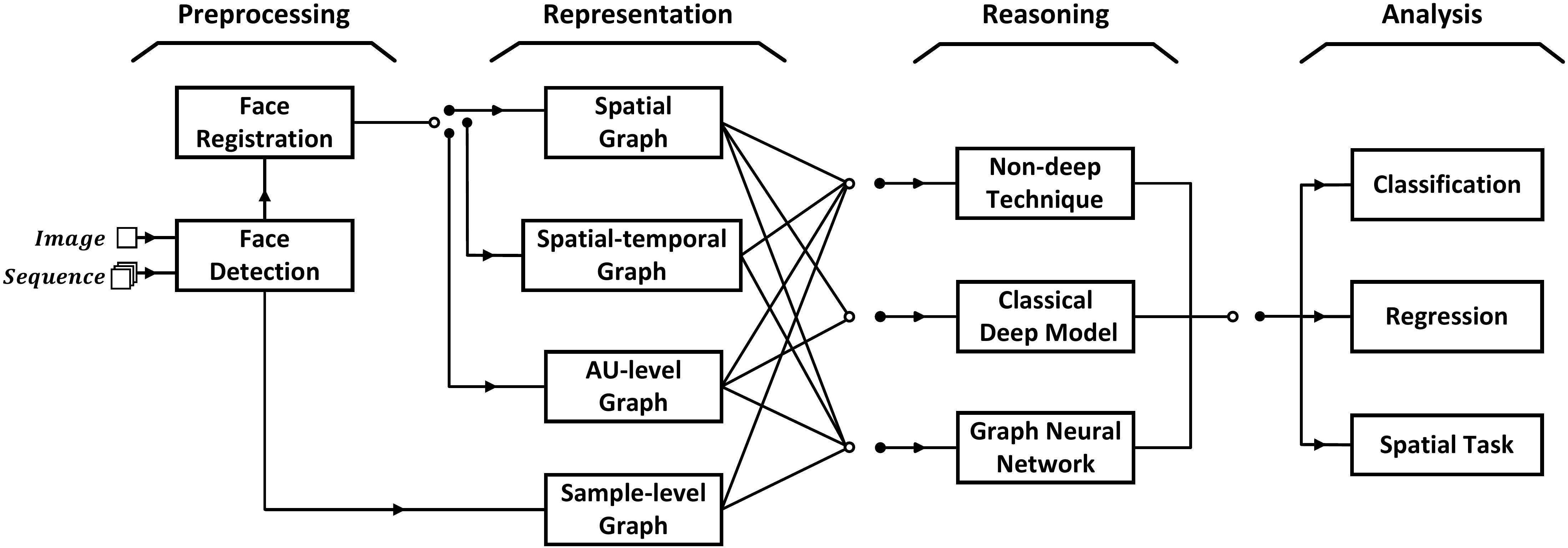}
  \caption{The pipeline of graph-based facial affect analysis methods.}
  \label{fig:pipe}\vspace{-10pt}
\end{figure*}

\begin{figure}[!h]
  \centering
  \includegraphics[width=0.85\columnwidth]{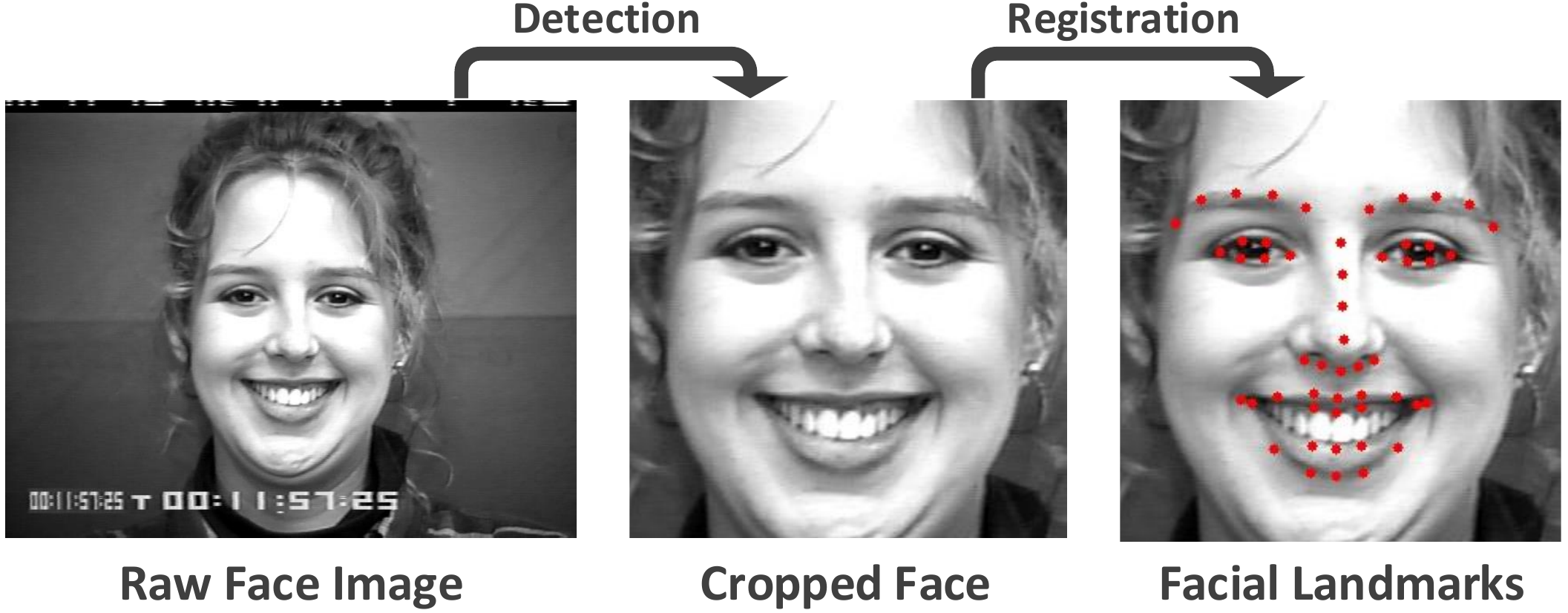}
  \caption{An illustration of face detection and face registration. The \emph{happy} image is from CK+ database \cite{lucey2010extended}.}
  \label{fig:prep}\vspace{-10pt}
\end{figure}

A standard FAA method can be broken down into fundamental components: face preprocessing, affective representation, and task analysis. As a new branch of FAA, the graph-based method also follows this generic pipeline (see Fig. \ref{fig:pipe}). Face detection and registration are two necessary pre-steps that first locate faces and normalize facial variations, sometimes also providing facial landmarks \cite{baltruvsaitis2016openface, baltrusaitis2018openface}. Fig. \ref{fig:prep} presents an illustration of the preprocessing steps. Early methods like \emph{Viola and Jones} \cite{viola2001rapid}, \emph{Mixtures of Trees} \cite{zhu2012face}, and \emph{Active Appearance Model (AAM)} \cite{cootes2001active} have been widely used for this purpose. Recently, cascaded deep approaches with real-time performance are popular, such as \emph{Multi-Task Cascaded Convolutional Network} \cite{zhang2016joint}, \emph{Hyperface} \cite{ranjan2017hyperface}, and \emph{Supervision by Registration and Triangulation} \cite{dong2020supervision}. Please refer to \cite{bulat2017far,masi2018deep,wang2018facial} for more specific information.

Compared to other existing methods, the graph-based FAA pays more attention to representing facial affects with graphs and obtaining affective features from such representation by graph reasoning.  

In mathematical terms, a graph can be denoted as $G=(V,E)$. The node set $V$ contains all the representations of the entities in the graph, and the edge set $E$ contains all the structure information between two entities. Thus, when $E$ is empty, $G$ becomes an unstructured collection of entities (e.g., independent local facial areas \cite{zhang2017facial}). Meanwhile, we could also define some initial graph structure ahead of the relational model, which is a general practice in many affective graph representations \cite{kumar2021micro,dapogny2018confidence,song2021dynamic}. 

Given this unstructured collection, performing relational reasoning requires the model to infer the structure of these entities before predicting the property or category of an object. Naturally, generic approaches need to be adjusted depending on affective graph representations or propose new graph-based approaches to infer the latent relationship and extract the final affective feature. 

The two components can perform separately or arranged as an end-to-end framework. They are expected to exhibit better performance and generalization capability by manually or automatically providing richer information through prior knowledge. Hence, the advantages and limitations of different graph generation methods and their relational reasoning approaches are two main topics of this survey.

\section{Graph-based Affective Representations}\label{sec:gbar}
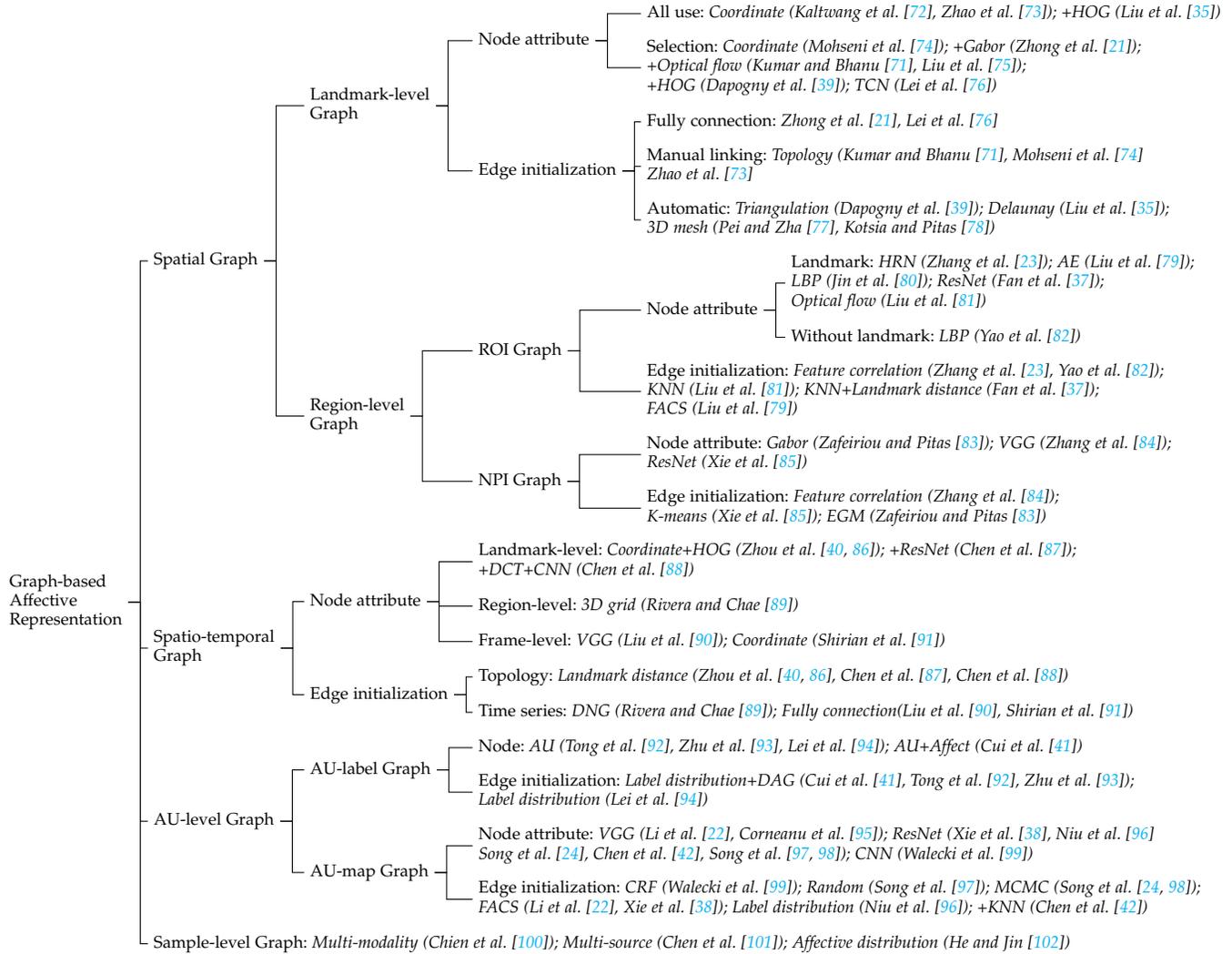
\begin{figure*}[htb]
  \scriptsize
  \caption{Taxonomy for Graph-based Facial Representation.}
  \label{fig:tree}
\begin{tikzpicture}[grow'=right]
  \tikzset{execute at begin node=\strut}
  % \tikzset{interior/.style={level distance=80pt}}
  % \tikzset{level 0/.style={level distance=50pt}}
  \tikzset{level 1/.style={level distance=60pt}}
  \tikzset{level 2/.style={level distance=65pt}}
  \tikzset{level 3/.style={level distance=70pt}}
  \tikzset{level 4/.style={level distance=70pt}}
  \tikzset{level 5/.style={level distance=60pt}}

  \tikzset{every tree node/.style={align=left, anchor=west}}

  \tikzset{edge from parent/.style={draw,edge from parent path={(\tikzparentnode.east) -- +(5pt,0) |- (\tikzchildnode)}}}
  \Tree [.{Graph-based\\Affective\\Representation} [.{Spatial Graph} [.{Landmark-level\\Graph} [.{Node attribute} {All use: \emph{Coordinate (\citet{kaltwang2015latent,zhao2021geometry}); +HOG (\citet{liu2019facial})} } 
                                                                                                                  {Selection: \emph{Coordinate (\citet{mohseni2014facial}); +Gabor (\citet{zhong2019graph});}\\ \emph{+Optical flow (\citet{liu2015main,kumar2021micro}); }\\ \emph{+HOG (\citet{dapogny2018confidence}); TCN (\citet{lei2020novel})}} ]
                                                                                               [.{Edge initialization} {Fully connection: \emph{\citet{zhong2019graph,lei2020novel}} } 
                                                                                                                       {Manual linking: \emph{Topology (\citet{mohseni2014facial,kumar2021micro}}\\ \emph{\citet{zhao2021geometry}} } 
                                                                                                                       {Automatic: \emph{Triangulation (\citet{dapogny2018confidence}); Delaunay (\citet{liu2019facial});}\\ \emph{3D mesh (\citet{pei20093d,kotsia2006facial})}} ] ]
                                                                     [.{Region-level\\Graph} [.{ROI Graph} [.{Node attribute} {Landmark: \emph{HRN (\citet{zhang2020region}); AE (\citet{liu2020relation}); }\\ \emph{LBP (\citet{jin2021learning}); ResNet (\citet{fan2020facial}); }\\ \emph{Optical flow (\citet{liu2018sparse})}} 
                                                                                                                              {Without landmark: \emph{LBP (\citet{yao2015capturing})}} ]
                                                                                                             {Edge initialization: \emph{Feature correlation (\citet{zhang2020region,yao2015capturing}); }\\ \emph{KNN (\citet{liu2018sparse}); KNN+Landmark distance (\citet{fan2020facial}); }\\ \emph{FACS (\citet{liu2020relation})}} ]
                                                                                             [.{NPI Graph} {Node attribute: \emph{Gabor (\citet{zafeiriou2008discriminant}); VGG (\citet{zhang2019context}); }\\ \emph{ResNet (\citet{xie2020adversarial})}} 
                                                                                                           {Edge initialization: \emph{Feature correlation (\citet{zhang2019context}); }\\ \emph{K-means (\citet{xie2020adversarial}); EGM (\citet{zafeiriou2008discriminant})}} ] ] ]
                                                   [.{Spatio-temporal\\Graph} [.{Node attribute} {Landmark-level: \emph{Coordinate+HOG (\citet{zhou2020facial,zhou2020learning}); +ResNet (\citet{chen2019efficient}); }\\ \emph{+DCT+CNN (\citet{chen2021cafgraph})} } 
                                                                                                 {Region-level: \emph{3D grid (\citet{rivera2015spatiotemporal})}} 
                                                                                                 {Frame-level: \emph{VGG (\citet{liu2021video}); Coordinate (\citet{shirian2021dynamic})}} ]
                                                                              [.{Edge initialization} {Topology: \emph{Landmark distance (\citet{zhou2020facial,zhou2020learning,chen2019efficient}, \citet{chen2021cafgraph})}} 
                                                                                                 {Time series: \emph{DNG (\citet{rivera2015spatiotemporal}); Fully connection(\citet{liu2021video,shirian2021dynamic})}} ] ]
                                                   [.{AU-level Graph} [.{AU-label Graph} {Node: \emph{AU (\citet{tong2007facial,zhu2014multiple,lei2021micro}); AU+Affect (\citet{cui2020label})}}
                                                                                         {Edge initialization: \emph{Label distribution+DAG (\citet{zhu2014multiple,cui2020label,tong2007facial}); }\\ \emph{Label distribution (\citet{lei2021micro})}} ]
                                                                      [.{AU-map Graph} {Node attribute: \emph{VGG (\citet{li2019semantic,corneanu2018deep}); ResNet (\citet{niu2019multi,xie2020assisted}}\\ \emph{\citet{chen2020label,song2021uncertain,song2021dynamic,song2021hybrid}); CNN (\citet{walecki2017deep})} } 
                                                                                       {Edge initialization: \emph{CRF (\citet{walecki2017deep}); Random (\citet{song2021uncertain}); MCMC (\citet{song2021dynamic,song2021hybrid}); }\\ \emph{FACS (\citet{li2019semantic,xie2020assisted}); Label distribution (\citet{niu2019multi}); +KNN (\citet{chen2020label})}} ] ]
                                                   [.{Sample-level Graph: \emph{Multi-modality (\citet{chien2020cross}); Multi-source (\citet{chen2021learning}); Affective distribution (\citet{he2019image})}} ]
        ]
\end{tikzpicture}\vspace{-10pt}
\end{figure*}

Affective representation is a crucial procedure for most graph-based FAA methods. Depending on the domain that an affective graph models, we categorize the strategy as Spatial graphs, Spatio-temporal graphs, AU-level graphs, and Sample-level graphs. Fig. \ref{fig:tree} illustrates a detailed summary of the literature using different graph representations. Note that many graph-based representations contain pre-extracted geometric or/and appearance features. Whether hand-crafted or learned, these feature descriptors are not essentially different from those used in non-graph-based affective representations. Interested readers can refer to \cite{sariyanidi2014automatic,corneanu2016survey,li2020deep} for a systematic understanding of this topic.

\subsection{Spatial Graph Representations}
\begin{figure}[tb]
  \centering
  \includegraphics[width=0.85\columnwidth]{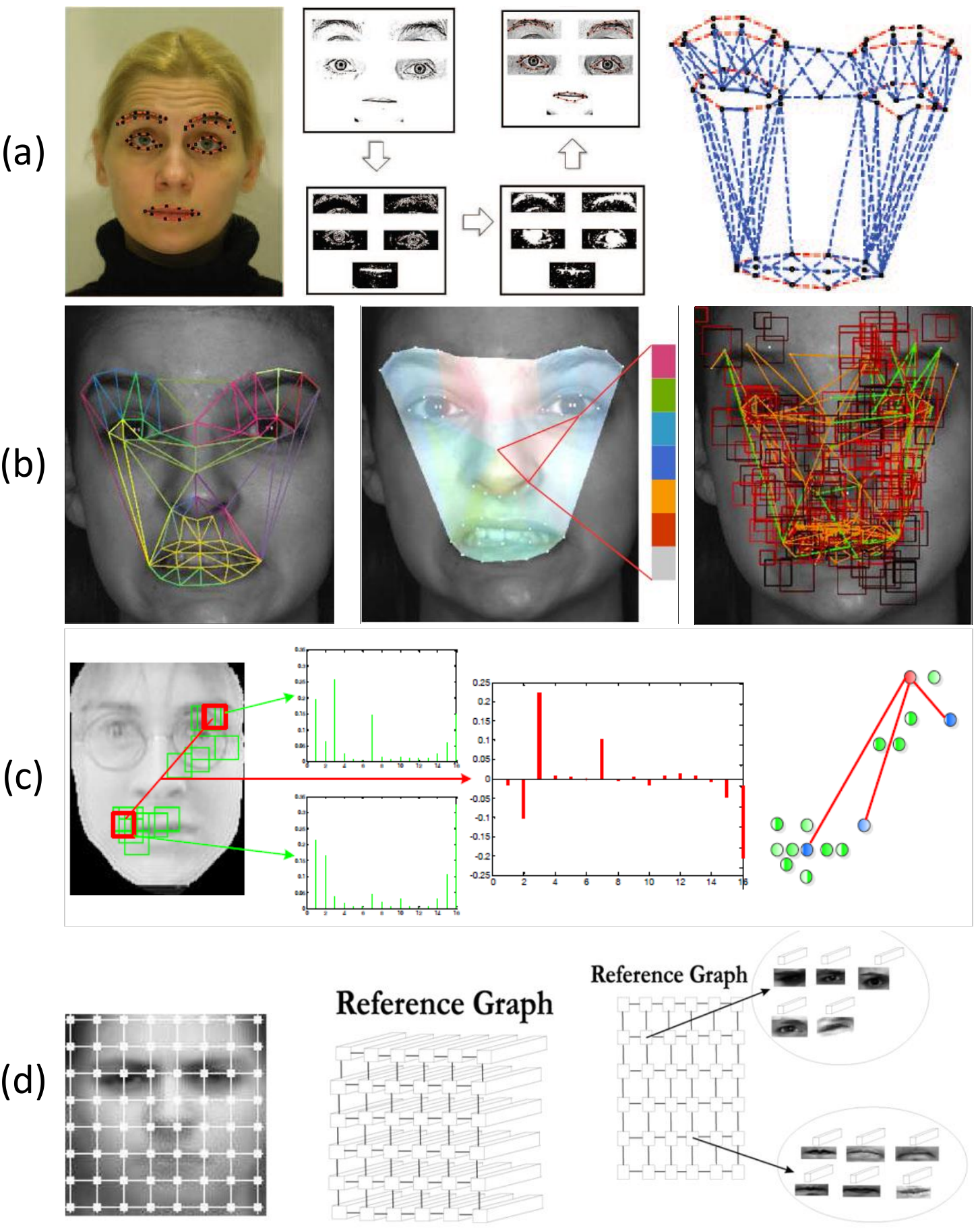}
  \caption{Spatial graphs. (a) Landmark-level graph with FACS-based edges \cite{mohseni2014facial}; (b) Landmark-level graph with automatic triangle edges \cite{dapogny2018confidence}; (c) ROI graph based on correlation without landmarks \cite{yao2015capturing}; (d) NPI graph with individual nodes \cite{zafeiriou2008discriminant}. Zoom in for better view.}
  \label{fig:spa}\vspace{-10pt}
\end{figure}

Non-graph-based spatial methods usually treat a facial affect as a whole representation or pay attention to variations among main face components or crucial facial parts \cite{zhao2016deep,zhang2017facial, oh2018survey}. For spatial affective graphs, facial changes are considered while their co-occurring relationships and affective semantics are represented as essential cues \cite{yao2015capturing,zhong2019graph, lei2020novel}. These approaches can be divided into landmark-level graphs and region-level graphs. Fig. \ref{fig:spa} illustrates frameworks of different spatial graph representations.

\subsubsection{Landmark-level graphs}
Facial landmarks are one of the most critical geometry that reflects the shape of face components and the structure of facial anatomy \cite{zhang2014facial}. Thus, it is natural to use facial landmarks as base nodes to generate a graph representation.

Limited by the detection performance, only a few landmarks that locate basic face components were applied in early graph representations \cite{kakumanu2006local}. Recently, graphs using more facial landmarks (e.g., 68 landmarks \cite{kazemi2014one}) are proposed to depict fine-grained facial shapes. For example, in \cite{liu2019facial} and \cite{zhao2021geometry}, the authors associated 68 landmarks with the AUs in FACS and made graph-based representations. The difference is that the former additionally employed local appearance features extracted by \emph{Histograms of Oriented Gradients (HOG)} \cite{dalal2005histograms} as node attributes; the latter proposed three landmark knowledge encoding strategies for enhanced geometric representations. Alternatively, \cite{kaltwang2015latent} formulated a \emph{Latent Tree (LT)} where 66 landmarks were set as parts of leaf nodes accompanied by several other leaf nodes of AU targets and hidden variables, which reflected the joint distribution of targets and features.

% \begin{figure}[h]
%   \centering
%   \includegraphics[width=0.85\columnwidth]{fig/lm.pdf}
%   \vspace{-0.3cm}
%   \caption{Facial muscle anatomy and related landmarks \cite{mohseni2014facial}.}
%   \label{fig:lm}
%   \vspace{-0.2cm}
% \end{figure}

Furthermore, some current methods select landmarks with significant contributions to avoid redundant information \cite{lei2020novel,rao2021facial}. Landmarks locating external contour and nose are frequently discarded \cite{mohseni2014facial, dapogny2018confidence} (see Figs. \ref{fig:spa}a, b) because they are considered irrelevant to facial affects. \cite{zhong2019graph} chose to remove the landmarks of the facial outline and applied a small window around each remaining landmark as one graph node, while the local features were extracted by \emph{Gabor} filter \cite{liu2003independent}. Since these local areas were segmented to introduce facial appearance into the graph representation rather than as independent nodes, similar to \cite{liu2019facial}, these methods are still classified in landmark-level graphs. On the other hand, adding extra reasonable landmarks was designed to generate comprehensive graph representations \cite{kumar2021micro,liu2021sg}, which could keep an appropriate dimension and represent sufficient affective information. 

A fully connected graph is the most intuitive way to form edges \cite{zhong2019graph,lei2020novel}. However, the number of edges is $n(n-1)/2$ for a complete graph with $n$ nodes, which means the complexity of the spatial relationship will increase as the number of nodes increases. This positive correlation is not helpful because landmarks in a facial component mostly move in concert rather than arbitrarily when conveying facial affects \cite{afzal2009perception}. Studies of point-light displays in emotion perception also show that more complex representations seem to be redundant \cite{baltruvsaitis2010synthesizing}. To this end, work like \cite{mohseni2014facial,kumar2021micro,liu2021sg,zhao2021geometry} manually reduced edges based on muscle anatomy and FACS. Another type of approach is exploiting triangulation algorithms \cite{dapogny2018confidence}, such as \emph{Delaunay} triangulation \cite{liu2019facial}, to generate graph edges consistent with true facial muscle distribution and uniform for different subjects. Similarly, the landmark-level graph with triangulation is also utilized in generating a sparse or dense facial mesh for 3D FAA \cite{pei20093d, kotsia2006facial}. The Euclidean distance is the simplest and most dominant metric for edge attributes of the above facial graphs, even with multiple normalization methods like inner-eyes distance \cite{liu2019facial,zhao2021geometry}. The \emph{Hop} distance has also been explored as edge attributes to model spatial relationships \cite{zhou2020facial,zhou2020learning}. Besides, several learning-based edge generation methods (e.g., \emph{LT} \cite{kaltwang2015latent}, \emph{Conditional Random Field (CRF)} \cite{walecki2017deep}) have been proposed to extract semantic information from facial graphs automatically. This part is discussed in detail in Sec. \ref{sec:rr-dbn} and \ref{sec:rr-td}.

\subsubsection{Region-level graphs}
Like geometric information, appearance information, especially in local facial regions, can also contribute to FAA \cite{lee2019context, wang2020region}. Using graph structures is an excellent choice to encode spatial relationships while representing texture changes in facial components \cite{yao2015capturing, zafeiriou2008discriminant}. There are two categories of region-level affective graphs: region of interest (ROI) graphs and non-prior information (NPI) graphs.

ROI graphs partition a set of facial areas as graph nodes related to affective display. Coordinates of facial landmarks are commonly applied to locate and segment ROIs. Unlike a few landmark-level graphs that only use texture near all landmarks as supplementary information, ROI graphs explicitly select meaningful areas as graph nodes, and edges do not entirely depend on established landmark relationships. \cite{zhang2020region} employed a \emph{High-Resolution Network (HRN)} \cite{wang2020deep} to regress ROI maps spotted by representative landmarks. Each spatial location in the extracted feature map was considered one graph node, while edges were induced among node pairs according to mappings between ROIs and AUs. Another example in \cite{fan2020facial} utilized feature maps of landmark-based ROIs outputted by the \emph{ResNet50} \cite{he2016deep} as nodes to construct a \emph{K-Nearest-Neighbor (KNN)} graph. For each node, its pair-wise semantic similarities were calculated, and the nodes with the closest \emph{Euclidean} distance were connected as initial edges. Similarly, \cite{liu2018sparse} also employed landmark-based ROIs, but the \emph{KNN} graph was generated in \emph{optical-flow} space to encode the local manifold structure for a sparse representation \cite{liu2015main}. Due to chained reactions among multiple AUs and the symmetrical structure of the human face, \cite{jin2021learning} proposed a parts-based graph that had manually linked edges by taking FACS and landmarks as references. The nodes were ROIs with \emph{Local Binary Pattern (LBP)} \cite{zhao2007dynamic} or deep features as attributes. In addition, the method of obtaining ROIs without relying on facial landmarks has also been studied \cite{yao2015capturing} (see Fig. \ref{fig:spa}c). 

Different from ROI graph representations, nodes in NPI graphs are evenly distributed in raw images or generated in a fully automatic manner without external knowledge. \cite{zafeiriou2008discriminant} created a reference bunch graph by evenly overlaying a rectangular graph on object images (see Fig. \ref{fig:spa}d). \emph{Gabor} filters were utilized for each graph node to compute a set of feature vectors for different facial instances. Recently, several methods have tried to introduce regions beyond facial parts or single face images as context nodes. \cite{zhang2019context} exploited the \emph{Region Proposal Network (RPN)} \cite{ren2016faster} with \emph{VGG16} \cite{simonyan2014very} to extract regions-level nodes, including the target face and its contexts, while edges were affective relationships calculated based on feature vectors. \cite{xie2020adversarial} built two NPI graphs for cross-domain FAA. First, holistic and local features were extracted as nodes for source and target domains. Then, global-to-global connection, global-to-local connection, and local-to-local connection were computed according to statistical feature distribution acquired by \emph{K-means} algorithm. 

\subsection{Spatio-Temporal Graph Representations}
\begin{figure}[tb]
  \centering
  \includegraphics[width=0.9\columnwidth]{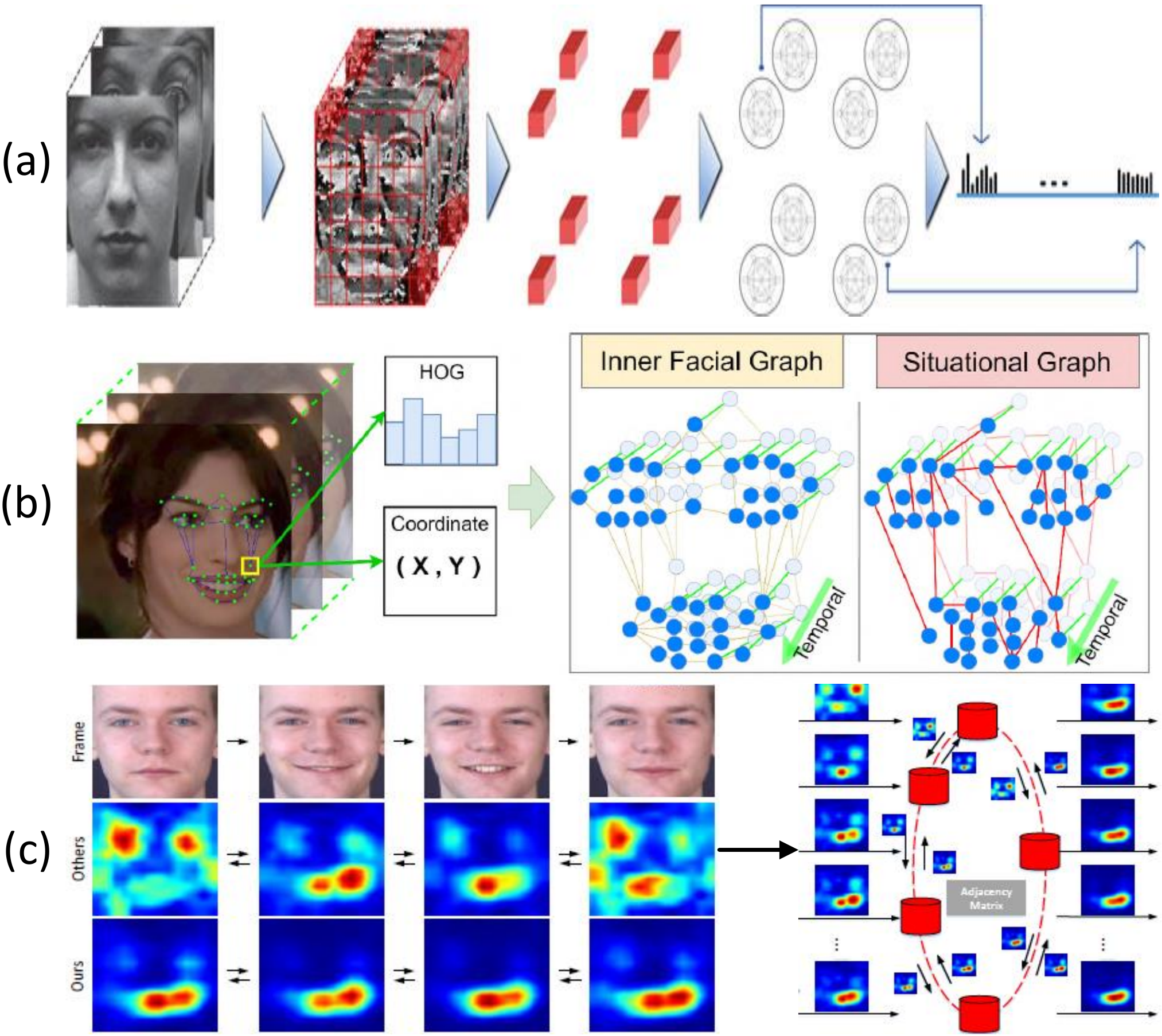}
  \caption{Spatio-temporal graphs. (a) Landmark-level graph with adaptive edges \cite{zhou2020learning}; (b) Region-level graph with edges based on transitional masks \cite{rivera2015spatiotemporal}; (c) Frame-level graph \cite{liu2021video}. Zoom in for better view.}
  \label{fig:st}\vspace{-10pt}
\end{figure}

Spatio-temporal representations deal with a sequence of frames within a temporal window and describe the dynamic evolution of facial variations \cite{liu2020phase}. In particular, introducing temporal information allows nodes to interact with each other at different times and generates a more complex affective graph. Fig. \ref{fig:st} presents frameworks of various spatio-temporal graph representations.

Extend spatial graphs to the spatio-temporal domain is currently the main route. \cite{rivera2015spatiotemporal} exploited weighted compass masks to obtain 2D directional number responses and 3D space-time directional edge responses corresponding to each of the symmetry planes of a cube. The two masks of given local neighborhoods were nodes in a spatio-temporal \emph{Directional Number Transitional Graph (DNG)}, which could represent salient facial changes and statistic frequency of affective behaviors over time (see Fig. \ref{fig:st}a). 

Several representations have been proposed to define temporal connections between landmarks, which can be seen as landmark-level spatio-temporal graphs. \cite{chen2021cafgraph} developed a context-aware facial multi-graph where intra-face edges were initialized based on morphological and muscular relationships, and inter-frame edges were created by linking the same node between consecutive frames. Similar landmark-based edge initialization in the temporal domain was also utilized in \cite{chen2019efficient, zhou2020facial}. In \cite{zhou2020learning}, authors introduced a connectivity inference block that could automatically generate dynamic edges for a spatio-temporal situational graph of part-occluded affective faces (see Fig. \ref{fig:st}b). 

Unlike landmark-level graphs, \cite{liu2021video} first extracted a holistic feature of each frame and set them as individual nodes to establish a fully connected graph (see Fig. \ref{fig:st}c), which could be seen as a frame-level spatio-temporal graph. Similar work includes \cite{shirian2021dynamic} that took \emph{Discrete Cosine Transform (DCT)} features as node attributes. Edge connections of these methods would be established by learning the long-term dependency of nodes in time series (discussed in Sec. \ref{sec:agrr}). 

\subsection{AU-level Graph Representations}
\begin{figure}[tb]
  \centering
  \includegraphics[width=0.9\columnwidth]{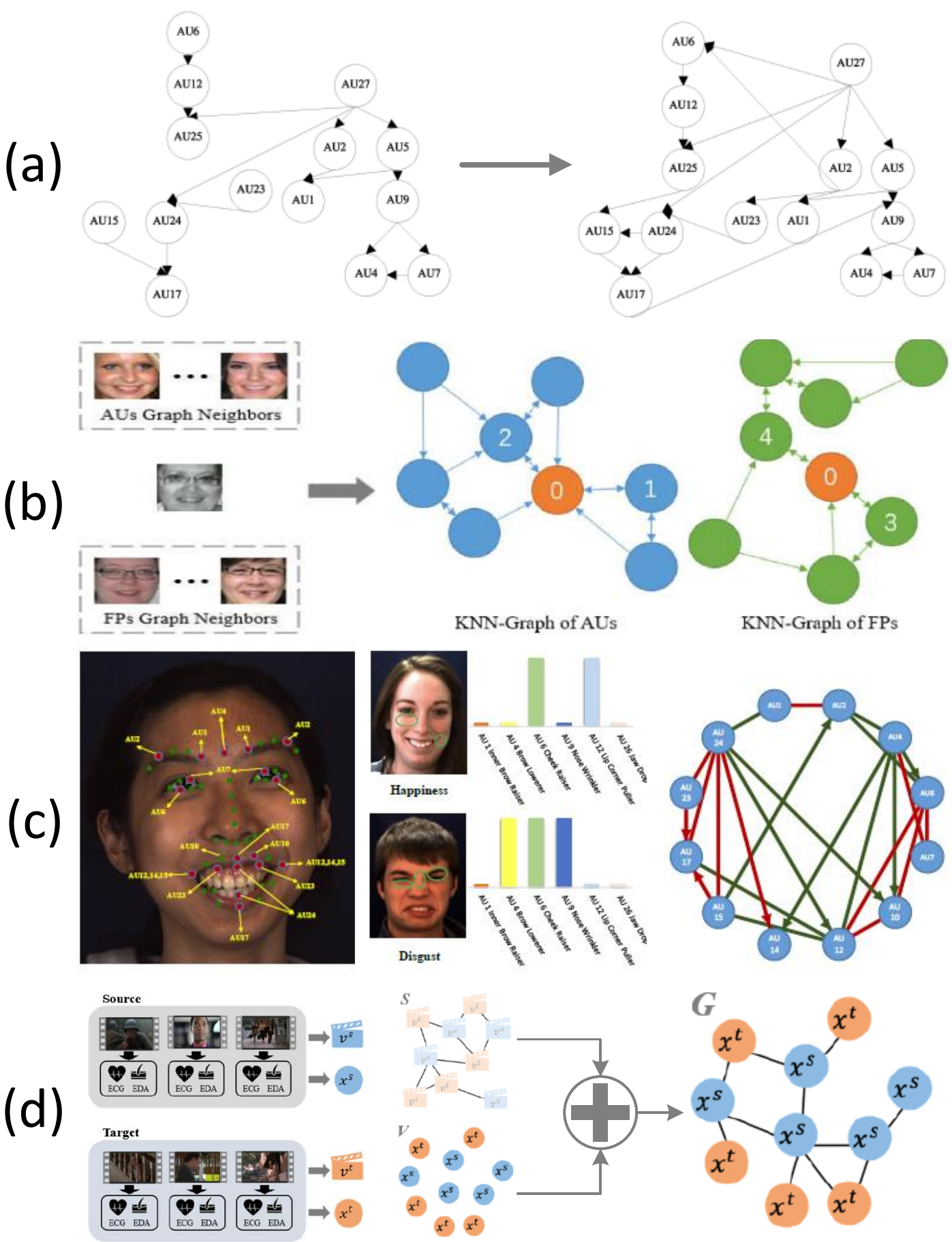}
  \caption{AU-level and Sample-level graph representations. (a) AU-label graph with edges generated from training data \cite{tong2007facial}; (b) AU-map graph with FACS based edges \cite{li2019semantic}; (c) Auxiliary graphs of AUs and landmarks \cite{chen2020label}; (d) Sample-level multi-modal graph of visual and physiological signals \cite{chien2020cross}. Zoom in for better view.}
  \label{fig:au}\vspace{-10pt}
\end{figure}

Apart from using knowledge of AUs and FACS in the above two types of affective graphs, many graph-based representations have been proposed to model affective information from the perspective of AUs themselves. We divide these approaches into two categories: AU-label graph and AU-map graph. Fig. \ref{fig:au} shows frameworks of different AU-level graph representations.

\subsubsection{AU-label graphs}
Unlike spatial and spatio-temporal graphs, AU-label graphs were built from the label distribution of training data \cite{zhu2014multiple,cui2020knowledge}. \cite{tong2007facial} computed the co-occurrence and co-absence dependency between every AU pair from the existing database (see Fig. \ref{fig:au}a). Since the dependency is not always symmetric, these AU label relationships were used as edges to construct a \emph{Directed Acyclic Graph (DAG)}. In \cite{lei2021micro}, an AU-label graph was built with a data-driven asymmetrical adjacency matrix that denoted the conditional probability of co-occurring AU pairs. AU labels were transformed into high-dimensional node vectors as node attributes \cite{mikolov2013distributed}. On the other hand, \cite{cui2020label} established a \emph{DAG} where object-level labels (affect categories) and property-level labels (AUs) were regarded as parent nodes and child nodes, respectively. The conditional probability distribution of each node to its parents was measured to obtain graph edges for correcting existing labels and generating unknown labels. A similar idea was achieved in \cite{chen2020label} to boost affective feature learning in large-scale FAA databases (see Fig. \ref{fig:au}b).

\subsubsection{AU-map graphs}
AU-map graphs are intuitively close to region-level spatial graphs, especially ROI graphs, because they both employ local feature maps as graph nodes. \cite{li2019semantic} is an example in between. Twelve AUs features were learned through landmark-based ROI features cropped from a multi-scale global appearance feature \cite{simonyan2014very}. These AU features and the AU relationships gathered from training data and manually pre-defined edge connections \cite{tian2001recognizing} were combined to construct a knowledge graph (see Fig. \ref{fig:au}c). However, the significant difference is that AUs define a set of facial muscle actions, which means there might be multiple AUs in the same ROI. Like in Fig. \ref{fig:fa}, AU12 and AU15 co-occur at lip corners but refer to '\emph{puller}' and '\emph{depressor}', respectively. Therefore, for many AU graphs, their definition of nodes is independent of those in ROI graphs, even though they are similar in feature map extraction. For instance, graph nodes in \cite{xie2020assisted} were AU features directly obtained by \emph{ResNet} without defining ROIs. The homologous protocol was also conducted in \cite{song2021hybrid}. 

Some special AU-map graphs have been proposed to introduce structure learning for more complex FAA tasks. For AU intensity estimation, \cite{walecki2017deep} trained a \emph{Convolutional Neural Network (CNN)} to learn deep AU features from multiple databases jointly. The \emph{copula functions} \cite{berkes2008characterizing} were applied to model pair-wise AU dependencies in a \emph{CRF} graph. In addition, \emph{Bayesian networks (BNs)} are also used to capture the AU inherent dependencies for this task \cite{song2021hybrid,song2021dynamic}. To account for indistinguishable affective faces, \cite{corneanu2018deep} designed a \emph{VGG}-like patch prediction module plus a fusion module to predict the probability of each AU. A prior knowledge taken from the given databases and a mutual gating strategy were used simultaneously to generate initial edge connections. To model uncertainty samples in real-world databases, \cite{song2021uncertain} established an uncertain graph, in which a weighted probabilistic mask that followed \emph{Gaussian} distribution was imposed on each AU feature map. By doing this, the importance of edges and the underlying uncertain information could be encoded in the graph representation. Another attempt in \cite{niu2019multi} boosted semi-supervised AU recognition for labeled and unlabeled face images. The parameters of two AU classifiers were used as graph nodes to share the latent relationships among AUs.

\subsection{Sample-level Graph Representations}
Recently, several graph representations beyond a single sample have been proposed, which indicates that this is still an open research field. In \cite{he2019image}, a correlation graph with word-embedded affective labels as nodes was built for distribution learning. Its edges could be generated either by psychologically normalized \emph{Gaussian} function or conditional probabilities. To combine signals from multiple corpora, \cite{chien2020cross} proposed a dual-branch framework, in which the visual semantic features were extracted in source and target sets. These features were then retrieved with correlation coefficients to generate positive edge connections for a learnable visual semantic graph (see Fig. \ref{fig:au}d). Besides, \cite{chen2021learning} constructed a \emph{KNN} graph with edges of binary weights to preserve the intrinsic geometrical structure of source and target data, which can seek more latent common information to reduce the distribution difference and make representations more discriminative. 

\subsection{Discussion}
As a significant part of the graph-based FAA method, different affective graph representations have their merits, shortcomings, and requirements (see Table \ref{tab:agad}).

\begin{table*}[!htbp]
  \centering
  \scriptsize
  \caption{An Overview of Affective Graph Representations}
  \label{tab:agad}
  \setlength{\tabcolsep}{1.5mm}{
  \begin{tabular}{|l|l|l|l|l|}
    \toprule[1.5pt]
    \makecell[c]{Category} & \makecell[c]{Branch} & \makecell[c]{Strength} & \makecell[c]{Limitation} & \makecell[c]{Demand} \\ \midrule[0.5pt]
    \multirow{3}*{Spatial} & Landmark-level & \makecell[l]{Effective facial geometry embedding;\\ Flexible structural relationships} & Sensitive to the landmark detection accuracy & High-quality face registration\\ \cmidrule[0.5pt](r){2-5} 
     & Region-level & \makecell[l]{Versatile local texture extraction;\\ Underlying correlation beyond locations} & \makecell[l]{Landmark sensitive for related ROI graphs;\\ Redundant or missing regions for NPI graphs} & Suitable for various situations \\ \midrule[0.5pt]
    \multicolumn{2}{|l|}{Spatio-Temporal} & Extra dynamic evolution information & Relatively fixed edges in the time domain & Video/Sequence input \\ \midrule[0.5pt]
    \multirow{2.5}*{AU-level} & AU-label & \multirow{2.5}*{\makecell[l]{Meaningful semantic dependencies;\\ Explicit prior knowledge introduction}} & Cannot be an end-to-end framework & \multirow{2.5}*{Reliable \& sufficient AU annotations} \\ \cmidrule[0.5pt](r){2-2} \cmidrule[0.5pt](r){4-4}
     & AU-map & & Unstable co-occurring distributions & \\ \midrule[0.5pt]
     \multicolumn{2}{|l|}{Sample-level} & \makecell[l]{Modular to existing architectures;\\ Cross-corpus information} & Lack of in-face modelling & Large-scale/Multiple databases \\
    \bottomrule[1.5pt]
  \end{tabular}}
\end{table*}

\emph{Spatial graph representations:}
Conceptually, landmark-level graphs model the facial shape variations of fiducial points and easily generate the internal structural relationships of different affective displays. However, most methods are sensitive to facial landmarks' detection errors, thereby failing in uncontrolled conditions. On the other hand, the selection of landmarks and the connection of edges have not yet formed a standard rule. Their effects on the graph representation are rarely reported, even though some FACS-based strategies have been designed. Region-level graphs explicitly regard an affective face as multiple local crucial facial areas compared with landmark-level graphs. The spatial relationships among selected regions are measured through feature similarity instead of manual initialization based on facial geometry. The circumstance resulting from inaccurate or unreasonable landmarks will also impact related ROI graphs. Since most NPI graphs utilize a region searching strategy, the problem is how to avoid the loss of target face and how to exclude invalid regions. 

\emph{Spatio-temporal graph representations:}
With extra dynamic affective information, spatio-temporal graphs can help aggregate evolution features in continuous time. For landmark-level methods, the current initialization strategy of edges is to link the facial landmark with the same index frame by frame. Unfortunately, no research has been reported to learn the interaction of landmarks with different indexes in the temporal dimension. Besides, in addition to \emph{Euclidean} distance and \emph{Hop} distance, other edge attributes measurement methods should also be explored to model the semantic context both spatially and temporally. For the frame-level methods, embedding domain knowledge related to affective behaviours like the muscular activity by graph structure is not explicitly considered in recent work. Therefore, building a hybrid spatio-temporal graph is a practical way to simultaneously encode the two levels of affective information.  

\emph{AU-level graph representations:}
As a distinctive type, AU-level graphs provide certain semantics of facial affects by representing each AU and its co-occurrence dependency. The measurement criteria of AU correlations are versatile but not general. Most AU-label graphs rely on the label distributions of one or multiple given databases. Nevertheless, AU labelling requires annotators with professional certificates and is a time-consuming task that causes existing databases with AU annotations to be usually small-scale. Therefore, the distribution from limited samples may not reflect the true dependencies of individual AUs, and its impact on FAA still needs to be assessed.

\emph{Sample-level graph representations:}
Sample-level graphs are an appealing field that introduces latent relationships in data distributions. Such characteristic makes it convenient to integrate with existing FAA methods. However, it also puts forward higher requirements for the diversity and balance of samples. On the other hand, to the best of our knowledge, there is no work combining sample-level and other in-face graphs to construct a joint representation, which we think is a good topic.

\section{Affective Graph Relational Reasoning}\label{sec:agrr}
Generally, graph relational reasoning can be considered a two-step process, i.e., understanding the structure from a certain group of entities and making inferences of the system as a whole or the property within \cite{kemp2008discovery}. However, things are slightly different in the case of graph-based FAA. Depending on what kind of affective graph representation is exploited, the contribution of graph relational reasoning can be either merged before the decision level with other affective features or reflected as a collaborative way in the level of feature learning. 

In this Section, we review relational reasoning methods designed for affective graph representations in four categories: \emph{Dynamic BNs (DBNs)}, classical deep models, \emph{Graph Neural Networks (GNNs)} and non-deep machine learning techniques.

\subsection{Dynamic Bayesian Networks}\label{sec:rr-dbn}
\emph{DBNs} are often used to reason about relationships among facial displays like AUs \cite{el2005real} and, of course, for AU-label graph representations. The \emph{BN} is a \emph{DAG} that reflects a joint probability distribution among a set of variables. In the work of \cite{tong2007facial,cui2020knowledge}, a \emph{DAG} was manually initialized according to prior knowledge, and than large databases were used to perform structure learning to find the optimal probability graph structure. After that, the probabilities of different AUs were inferred by learning the \emph{DBN}. Following this idea, \cite{zhu2014multiple} additionally integrated \emph{DBN} to a multi-task feature learning framework and made the AU inference by calculating the joint probability of each category node. Sometimes \emph{DBN} is also combined with some statistical methods to explore different graph structures \cite{song2021hybrid,song2021dynamic}, such as \emph{Hidden Markov Models} \cite{baltruvsaitis2011real}. Another advanced research of \emph{DBN} is \cite{cui2020label} that modeled the inherent relationships between category labels and property labels. Its parameters were utilized to denote the conditional probability distribution of each AU given the facial affect. The wrong labels could be corrected by leveraging the dependencies after the structure optimization.

\subsection{Adjustments of Classical Deep Models} \label{sec:rr-deep}
\begin{figure}[tb]
  \centering
  \includegraphics[width=0.9\columnwidth]{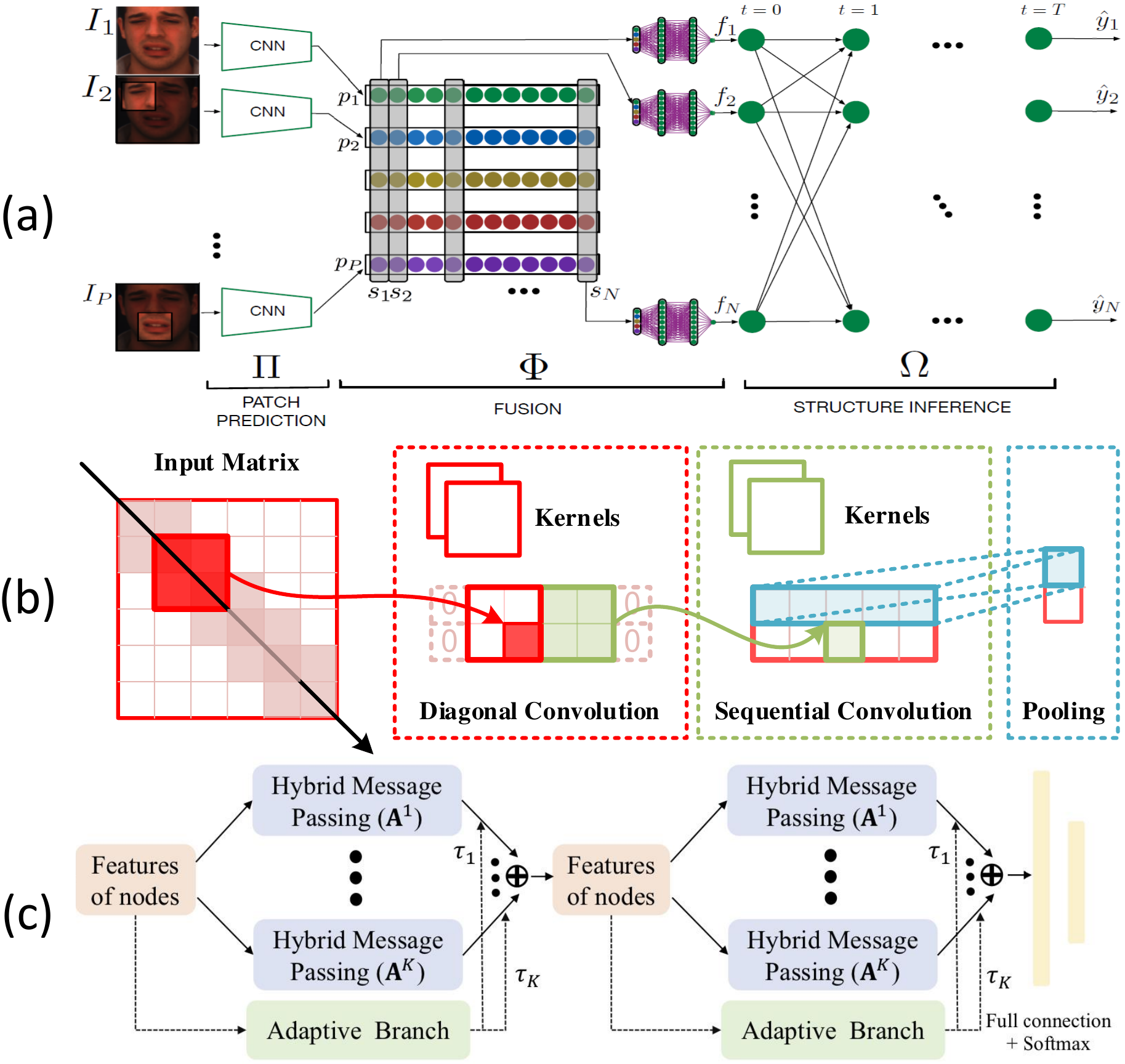}
  \caption{Classical deep models for graph relational reasoning. (a) RNN \cite{corneanu2018deep}; (b) CNN \cite{liu2019facial}; (c) MLP \cite{dapogny2018confidence}. Zoom in for better view.}
  \label{fig:dnn}
\end{figure}

Before \emph{GNNs} are widely employed, many studies have adopted conventional \emph{Deep Neural Networks (DNNs)} to process affective representations with the graph structure. These deep models are not explicitly designed but can conduct standard operations on structural graph data by adjusting the internal architecture or applying an additional transformation to the input graph representation. Fig. \ref{fig:dnn} shows examples of classical deep models for graph relational reasoning.

\subsubsection{Recurrent neural networks}
The \emph{Recurrent Neural Networks (RNNs)} variant is one of the successfully extended model types for handling graph structural inputs. Similar to random walk, \cite{zhong2019graph} applied a \emph{Bidirectional RNN} to deal with its landmark-level spatial graph representation in a rigid order. The Gabor features of each graph node was updated by multiplying with the average of the connected edges to incorporate the structural information. Subsequently, the nodes were iterated by the RNN with learnable parameters in forwarding and the backward direction. In \cite{corneanu2018deep}, the authors built a structure inference module to capture AU relationships from an AU-map graph representation. Based on a collection of interconnected recurrent structure inference units and a parameter sharing \emph{RNN}, the mutual relationship between two nodes could be updated by replicating an iterative message passing mechanism with the control of a gating strategy (see Fig. \ref{fig:dnn}a). Following the sequential idea of RNNs,\cite{li2019semantic} exploited a \emph{Gated Graph Neural Network (GGNN)} \cite{li2016gated} that calculated the hidden state of the next time-step by jointly considering the current hidden state of each node and adjacent nodes. The relational reasoning could be done through the iterative update of \emph{GGNN} over its AU-map graph.

\subsubsection{Convolutional neural networks}
Unlike the sequential networks, \cite{liu2019facial} utilized a variant \emph{CNN} to process the landmark-level spatial affective graph. Compared to standard convolution architectures, the convolution layer in this study convolved over the diagonal of a particular adjacency matrix to aggregate the information from multiple nodes. Then a list of the diagonal convolution outputs was further processed by three 1D sequential convolution layers. The corresponding pooling processes were performed behind convolution operations to integrate feature sets (see Fig. \ref{fig:dnn}b). Another attempt for landmark-level spatial graph representations is the \emph{Graph Temporal Convolutional Networks (Graph-TCN)} \cite{lei2020novel}. It followed the idea of \emph{TCNs} that consisted of residual convolution, dilated causal convolution, and weight normalization \cite{bai2018empirical}. By using different dilation factors, TCNs were applied to convolve the elements inside one node sequence and from multiple node sequences. Thus, the \emph{TCN} for a node and \emph{TCN} for an edge could be trained respectively to extract node feature and edge feature simultaneously. Besides, \cite{fan2020facial} exploited a Semantic Correspondence Convolution module to model the correlation among its region-level spatial graph. Based on the assumption that the channels of co-occurring AUs might be activated simultaneously, the \emph{Dynamic Graph CNN (DG-CNN)} \cite{wang2019dynamic} was applied on the edges of the constructed \emph{KNN} graph to connect feature maps sharing similar visual patterns. After the aggregation function, affective features were obtained to estimate AU intensities. 

\subsubsection{Multilayer perceptron networks}
As a vanilla architecture, \emph{Multilayer Perceptrons (MLPs)} has also been explored. \cite{dapogny2018confidence} employed a hierarchical \emph{Auto-Encoder (AE)} based on \emph{MLPs} to capture relationships from a landmark-level spatial graph. Specifically, the first stage learned the texture variations from the extracted \emph{HOG} features for each node. In contrast, the second stage accumulated features of multiple nodes whose appearance changes were closely related and computed the confidence scores as the triangle-wise weights over edges. Finally, a \emph{Random Forest (RF)} was used for facial affect classification and AU detection simultaneously. In \cite{song2021hybrid}, a hybrid graph network composed of different dynamic \emph{MLPs} performed multiple types of message passing, which provided more complementary information for reasoning the positive and negative dependencies among AU nodes (see Fig. \ref{fig:dnn}c).

\subsection{Graph Neural Networks}\label{sec:rr-gnn}
\begin{figure*}[htb]
  \centering
  \includegraphics[width=1.95\columnwidth]{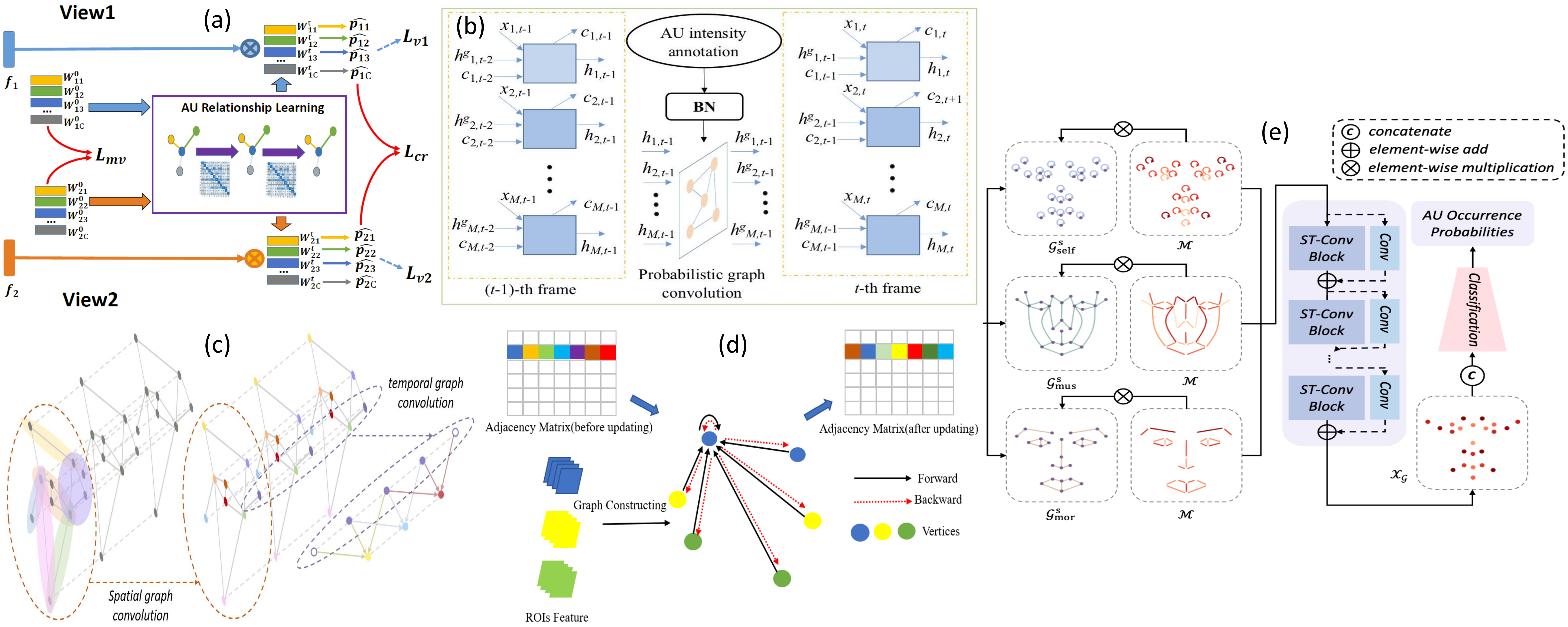}
  \caption{GNNs for graph relational reasoning. (a) GCN as an auxiliary module \cite{niu2019multi}; (b) GCN as a collaborative framework \cite{song2021dynamic}; (C) STGCN \cite{chen2021cafgraph} (d) Spectral GCN \cite{jin2021learning}; (e) GAT \cite{chen2021cafgraph}. Zoom in for better view.}
  \label{fig:gnn}\vspace{-10pt}
\end{figure*}

Unlike conventional deep learning frameworks mentioned in Sec. \ref{sec:rr-deep}, \emph{GNNs} are proposed to extend the 'depth' from 2D image to graph structure and establish an end-to-end learning framework instead of additional architecture adjustment or data transformation \cite{li2018beyond}. Several types of \emph{GNNs} have successfully addressed the relational reasoning of affective graph representations in FAA methods. Fig. \ref{fig:gnn} illustrates several \emph{GNN} architectures for graph relational reasoning.

\subsubsection{Graph convolutional networks}
\emph{Graph Convolutional Networks (GCNs)}, especially the spatial \emph{GCN} \cite{kipf2017semi}, are the most popular \emph{GNN} in graph-based FAA research. Practically, \emph{GCNs} can be set as an auxiliary module \cite{lei2021micro,zhao2021geometry} or part of the collaborative feature learning framework \cite{song2021dynamic}. 

For the auxiliary module, \emph{GCNs} are applied immediately after the graph representation. However, the outputs of relational reasoning are not directly used for facial affect classification or AU detection but are later combined with other deep features as a weighting factor (see Fig. \ref{fig:gnn}a). \cite{niu2019multi} employed a two-layer \emph{GCN} for message passing among different nodes in its AU-level graph. Both the dependency of positive and negative samples were considered and used to infer a link condition between any two nodes. The output of \emph{GCN} was formulated as a weight matrix of the pre-trained AU classifiers. Besides, \emph{GCNs} can also be utilized following the above manner to execute relational reasoning on atypical graph representations, such as multi-target graph \cite{zhang2019context}, distribution graph \cite{he2019image}, and cross-domain graph \cite{xie2020adversarial}. 

For the collaborative framework, \emph{GCNs} usually inherit the previous node feature learning model progressively (see Fig. \ref{fig:gnn}b). Like in \cite{liu2020relation}, a \emph{GCN}-based multi-label encoder was proposed to update features of each node over a region-level spatial graph representation. The reasoning process was the same as that in the auxiliary framework. Similar studies also include \cite{zhang2020region} and \cite{chien2020cross}. In addition, to incorporate the dynamic in spatio-temporal graphs, \cite{liu2021video} set \emph{GCNs} as an imitation of attention mechanism or weighting mechanism to share the most contributing features to explore the dependencies among frames. After training, the structure helped nodes update features based on messages from the peak frame and emphasize the concerned facial region. A more feasible way is to apply \emph{Spatial Temporal GCN (STGCN)} \cite{li2019spatio} on spatio-temporal graphs \cite{zhou2020facial,chen2019efficient,zhou2020learning,chen2021cafgraph} (see Fig. \ref{fig:gnn}c). In their relational reasoning, features of each node were generated with its neighbor nodes in the current frame and consecutive frames by using spatial graph convolution and temporal convolution, respectively. 

Alternatively, the approach of spectral \emph{GCN} \cite{defferrard2016convolutional} has also been studied \cite{rao2021facial}. \cite{jin2021learning} devised a lightweight GCN following the \emph{Message Passing Neural Network} \cite{gilmer2017neural}. A learnable adjacency matrix was adapted to infer the spatial dependencies of ROI nodes in different facial affects. \cite{shirian2021dynamic} extended in \emph{Inception} idea from standard \emph{CNNs} to spectral \emph{GCNs} that captured emotion dynamics at multiple temporal scales. The yielded embeddings of different dimensions were jointly learned over a classification loss and a graph learning loss for the optimal graph structure. 

\subsubsection{Graph attention networks}
\emph{Graph Attention Networks (GATs)} aim to strengthen the node connections with high contribution and offer a more flexible way to process the graph structure \cite{velivckovic2018graph}. \cite{song2021uncertain} introduced an uncertain \emph{GNN} with \emph{GAT} as the backbone. The goal is to select valuable edges, depress noisy edges, and learn AU dependencies on its AU-map graph. In addition, the underlying uncertainties were considered in a probabilistic way, close to the idea of \emph{Bayesian} methods in \emph{GNN} \cite{zhang2019bayesian}, to alleviate the data imbalance by weighting the loss function. On the other hand, \emph{GAT} collaboratively worked with \emph{GCN} in \cite{kumar2021micro} to deal with two-stream graph inputs. Compared to applying \emph{GAT} directly, \cite{xie2020assisted} proposed a GNN that added a self-attention graph pooling layer after three sequential \emph{GCN} layers. A similar block was done in \cite{liu2021sg} which revised the \emph{GCN} block with channel and node attention. It improved the reasoning process on graph representations because only important nodes would be aggregated, including affective information and facial topology. To make nodes interact more dynamically instead of using a constant graph structure, \cite{chen2021cafgraph} applied a set of learnable edge attention masks to the \emph{STGCN} for subtle adjustments of the defined spatio-temporal graph representation (see Fig. \ref{fig:gnn}e).

\subsection{Non-deep Machine Learning Methods} \label{sec:rr-td}
Although refining deep features extracted by parameterized neural networks and gradient-based methods is the mainstream, they require numerous training samples for effective learning. Due to the insufficient data in the early years or the purpose of efficient computation, many non-deep machine learning techniques have been applied for affective graph relational reasoning. Graph structure learning is one of the widely used approaches. In \cite{kaltwang2015latent}, the reasoning of its spatial graph representation was conducted by \emph{LT} learning. Parameters update and graph-edit of \emph{LT} structure were performed iteratively to maximize the marginal \emph{log}-likelihood of a set of training data. \cite{walecki2017deep} employed \emph{CRF} to infer AU dependencies in an AU-map graph. The use of \emph{copula} functions allowed it to model non-linear dependencies among nodes easily. At the same time, an iterative balanced batch learning strategy was introduced to optimize the most representative graph structure by updating each set of parameters with batches. Approaches of graph feature selection are also exploited in this part, such as \emph{Graph Sparse Coding (GSC)} \cite{liu2018sparse,chen2021learning} and \emph{Elastic Graph Matching (EGM)} \cite{zafeiriou2008discriminant}. These methods have provided a more diverse concept for graph relational reasoning.

\subsection{Discussion}
Although all the methods above can achieve affective graph relational reasoning, the choice has a causal relationship with the type of graph representation (see Table \ref{tab:agrr}). 

\begin{table}[!htbp]
  \centering
  \scriptsize
  \caption{Causal Relationships between Graphs and Reasoning methods}
  \label{tab:agrr}
  \setlength{\tabcolsep}{2.5mm}{
  \begin{tabular}{c|cccc}
    \toprule[1.5pt]
    Category & Spatial & Spatio-Temporal & AU-level & Sample-level \\ \midrule[0.5pt]
    DBNs &  &  & $\surd$ & \\ 
    DNNs & $\surd$ & $\surd$ & $\surd$ & \\ 
    GNNs & $\surd$ & $\surd$ & $\surd$ & $\surd$ \\ 
    Non-deep & $\surd$ & $\surd$ & $\surd$ & $\surd$ \\
    \bottomrule[1.5pt]
  \end{tabular}}
\end{table}

\emph{Dynamic Bayesian network}: Nearly half of AU-label graph representations employ \emph{DBNs} as their relational reasoning model. However, the representation quality highly relies on the available training data that need balanced label distribution in positive-negative samples and categories. This strong assumption will limit the effectiveness of node dependencies learned by \emph{DBNs}. Another problem is that \emph{DBNs} can only be combined with facial features as a relatively independent module and are hard to integrate into an end-to-end learning framework.

\emph{Classical deep model}: Standard deep models, including \emph{CNNs}, \emph{RNNs} and \emph{MLPs}, have been explored to conduct graph relational reasoning before the emergence of \emph{GNNs}. Even if they are suitable for more graph representations than \emph{DBNs}, these grid models focus more on local features. The additional adjustments in input format or/and network architecture cause losses of node information or let node messages only pass and update in a specific sequence, which suppresses the global property represented by the graph. Thus, we think the specifically designed networks like \emph{GNNs} will become dominant in this part. 

\emph{Graph neural network}: \emph{GNNs} are developing techniques that make full advantages of graphs. Architectures with different focuses have been proposed but have their flaws as well. For instance, \emph{GCNs} cannot handle directed edges well (e.g., AU-level graphs), while \emph{GATs} only use the node links without considering edge attributes (e.g., spatial graphs). Besides, due to the low dimension of the nodes in affective graphs, too deep \emph{GNNs} may be counterproductive. In addition, being an auxiliary block or part of the whole framework will influence the construction of \emph{GNNs}. Therefore, managing graph representation and relational reasoning using \emph{GNNs} still need to be explored.

\emph{Non-deep methods}: Non-deep machine learning has a place in early studies and is even applied in recent work because no training is required. They partly inspire advanced techniques like \emph{DBNs} and \emph{GNNs}. Nevertheless, one of the reasons they have been replaced is that these approaches need to be designed separately to cope with different graph representations, similar to hand-crafted feature extraction. Hence, it is not easy to form a general framework. On the other hand, more training data and richer computing resources allow deep models to perform more effective and higher-level relational reasoning on affective graphs.

\section{Applications and Performance}\label{sec:appe}
According to different description models of facial affects, the FAA can be subdivided into multiple applications. The typical output of FAA systems is the label of a basic facial affect or AUs. Recent research also extends the goal to predict micro-expression or affective intensity labels or continuous affects. This section compares and discusses graph-based FAA methods from four main application categories: facial expression recognition, AU detection, micro-expression recognition, and a few special applications. Due to page limitation, we select most relevant and representative papers following these standards: published in more well-known forums in the past five years; or belonging to distinct branches of graph representation and reasoning for diversity consideration. 

\subsection{Databases}
\begin{figure}[tb]
  \centering
  \includegraphics[width=0.95\columnwidth]{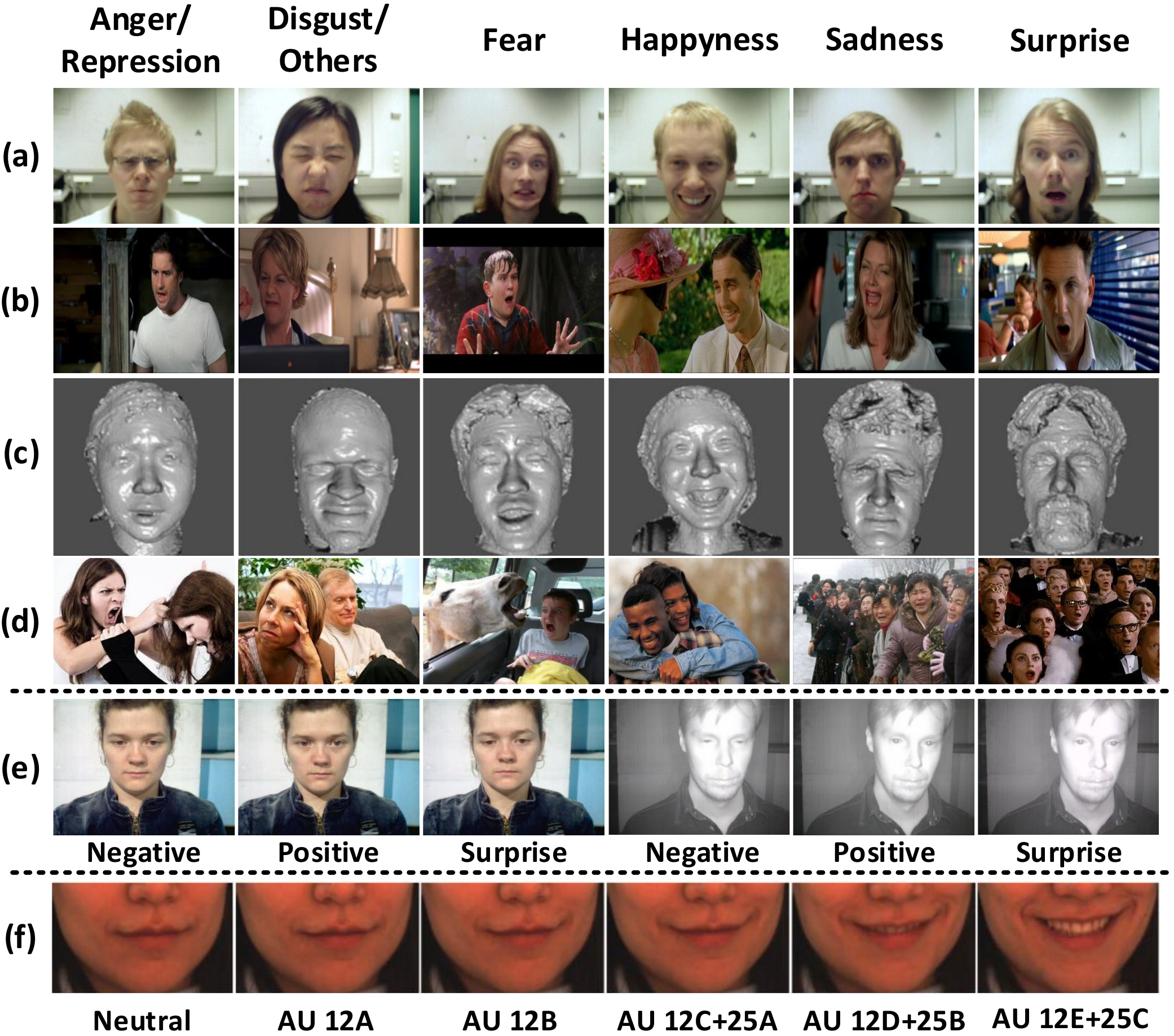}
  \caption{Facial affect databases. (a) Oulu-CASIA contains posed facial affects; (b) SFEW 2.0 has facial affects under in-the-wild scenarios; (c) BP4D provides 3D affective face images; (d) EMOTIC, multiple faces appear per image with VAD annotations; (e) The SMIC collects images of spontaneous micro facial affects in visual light and near infrared light; (f) DISFA offers frame-level AU intensity labels.}
  \label{fig:data}\vspace{-10pt}
\end{figure}

Most FAA studies apply public databases of facial affect as validation material. A comprehensive overview is presented in Table \ref{tab:db}. The characteristics of these databases are listed from four aspects: samples, attributes, graph-related properties, and certain contents. Fig. \ref{fig:data} exhibits several examples of facial affects under different conditions. In addition, for better interpreting the graph-based FAA, we summarize corresponding elements (e.g., landmark coordinates, AU labels) self-carried by databases, which are rarely considered in previous related surveys. 

\begin{table*}[!htbp]
  \centering
  \scriptsize
  \begin{threeparttable}
  \caption{An Overview of Facial Affect Databases}
  \label{tab:db}
  \setlength{\tabcolsep}{2mm}{
  \begin{tabular}{lcccccccccc}
    \toprule[1.5pt]
    \multirow{2}*{Database 'year} & \multicolumn{3}{c}{Samples} & \multicolumn{3}{c}{Attributes} & \multicolumn{3}{c}{Graph-based Properties$^3$} & \multirow{2}*{Special Contents$^4$} \\ \cmidrule[0.5pt](r){2-4} \cmidrule[0.5pt](r){5-7} \cmidrule[0.5pt](r){8-10}
     & Data Type & Subjects & Number & Eli. \& Sou.$^1$ & Affects$^2$ & B.B.$^1$ & LM & AU & Dynamic &  \\ \midrule[0.5pt]
    CK(+) '10 & Sequences & 97(123) & 486(593) & P \& Lab & 6B+N(+C) & $\bullet$ & $\bullet$ & $\bullet$ + I & $\bullet$ & / \\
    MMI '10 & Images/Videos & 75 & 740/2900 & P \& Lab & 6B+N & $\circ$ & $\circ$ & $\bullet$ & $\bullet$ + D & Head pose \\
    Oulu-CASIA '11 & Sequences & 80 & 2880 & P \& Lab & 6B & $\bullet$ & $\circ$ & $\circ$ & $\bullet$ & NIR \\
    DISFA '13 & Sequences & 27 & 130,000 & S \& Lab & 6B+N & $\bullet$ & $\bullet$ & $\bullet$ + I & $\bullet$ + F & / \\ \midrule[0.5pt]
    FER-2013 '13 & Images & / & 35887 & S \& Web & 6B+N & $\bullet$ & $\circ$ & $\circ$ & $\circ$ & Wild \\
    % FERA-2015 '15 & Videos & 41 & 328 & S \& Web & 6B+N & $\bullet$ & $\circ$ & $\circ$ & $\circ$ & Wild \\ 
    SFEW 2.0 '15 & Images & / & 1766 & S \& Movie & 6B+N & $\bullet$ & $\bullet$ & $\circ$ & $\circ$ & Wild \\
    AFEW 7.0 '17 & Videos & / & 1809 & S \& Movie & 6B+N & $\bullet$ & $\bullet$ & $\circ$ & $\bullet$ & Audio, Wild \\ \midrule[0.5pt]
    BU-3DFE '06 & Images & 100 & 2500 & P \& Lab & 6B+N & $\bullet$ & $\bullet$ & $\circ$ & $\circ$ & 3D, Multi-view, I \\
    BU-4DFE '08 & Videos & 101 & 606 & P \& Lab & 6B+N & $\bullet$ & $\bullet$ & $\bullet$ + I & $\bullet$ & 3D, Multi-view \\
    BP4D '14 & Videos & 41 & 328 & S \& Lab & 6B+E+P & $\bullet$ & $\bullet$ & $\bullet$ + I & $\bullet$ + F & 3D, Head pose \\ \midrule[0.5pt]
    SMIC '13 & Sequences & 16 & 164 & S \& Lab & 3B$^\dag$+N & $\bullet$ & $\circ$ & $\circ$ & $\bullet$ & Micro., NIR \\
    CASME II '14 & Sequences & 35 & 247 & S \& Lab & 3B$^\ddag$+R+O & $\bullet$ & $\circ$ & $\bullet$ & $\bullet$ + D & Micro. \\
    SAMM '18 & Sequences & 32 & 159 & S \& Lab & 6B+C & $\bullet$ & $\circ$ & $\bullet$ & $\bullet$ + D & Micro. \\
    CAS(ME)$^2$ '18 & Sequences & 22 & 300+57 & S \& Lab & 3B$^\dag$+O & $\bullet$ & $\circ$ & $\bullet$ & $\bullet$ + D & Macro. \& Micro. \\ \midrule[0.5pt]
    EmotioNet '16 & Images & / & 950,000 & S \& Web & 6B+17Comp. & $\bullet$ & $\bullet$ & $\bullet$ + I & $\circ$ & Wild \\
    ExpW '18 & Images & / & 91,793 & S \& Web & 6B+N & $\bullet$ & $\circ$ & $\circ$ & $\circ$ & Multi-sub., Wild\\
    RAF-DB '19 & Images & / & 29672 & S \& Web & 6B+N+11Comp. & $\bullet$ & $\bullet$ & $\circ$ & $\circ$ & Wild \\
    AffectNet '19 & Images & / & 420,299 & S \& Web & 6B+N+C+O & $\bullet$ & $\bullet$ & $\circ$ & $\circ$ & V\&A, Wild \\
    EMOTIC '19 & Images & / & 23,571 & S \& Web & 6B+N+19Comp. & $\bullet$ & $\circ$ & $\circ$ & $\circ$ & V\&A, Multi-sub., Wild \\
    \bottomrule[1.5pt]
  \end{tabular}}
  \begin{tablenotes}
    \item[1] Eli.: elicitation; Sou.: source; P: posed; S: spontaneous; B.B.: bounding boxes; LM: landmarks; $\bullet$ = Yes, $\circ$ = No.
    \item[2] 6B: six basic affects; N: neutral; C: contempt; E: embarrassment; P: pain; O: others; R: repression; 3B$^\dag$: three basic affects (positive, negative, surprise); 3B$^\ddag$: three basic affects (happiness, disgust, surprise); Comp.: compound affects.
    \item[3] I: intensity annotation; D: onset-apex-offset annotation; F: frame-level annotation.
    \item[4] NIR: near-infrared; Multi-sub.: multiple subjects per image; Micro.: micro-expression; Macro.: macro-expression; Wild: in-the-wild; V\&A: valence and arousal.
  \end{tablenotes}
  \end{threeparttable}
\end{table*}

Databases containing posed facial affects, including Extended Cohn-Kanade Dataset (CK+) \cite{kanade2000comprehensive,lucey2010extended}, M\&M Initiative Facial Expression Database (MMI) \cite{valstar2010induced}, and Oulu-CASIA NIR\&VIS Facial Expression Database (Oulu-CASIA) \cite{zhao2011facial}, are chosen by early FAA methods. More challenging databases, such as FER-2013 \cite{goodfellow2013challenges}, Static Facial Expression in the Wild (SFEW) 2.0 \cite{dhall2015video}, and Acted Facial Expression in the Wild (AFEW) 7.0 \cite{dhall2017individual}, tend to acquire spontaneous affective data from complex and wild environments. 

Some databases also contain intensity labels of facial affects and even AUs, e.g., Denver Intensity of Spontaneous Facial Action Database (DISFA) \cite{mavadati2013disfa}, Binghamton University 3D/4D Facial Expression Database (BU-3DFE/4DFE) \cite{yin20063d,zhang2013high} and Binghamton-Pittsburgh 3D Dynamic Spontaneous Facial Expression Database (BP4D) \cite{zhang2014bp4d}. 

Another type of database is for micro-expressions. Participants are required to keep a neutral face while watching videos associated with induction of specific affects \cite{zhao2019automatic}. Following this setting, Spontaneous Micro Facial Expression Database (SMIC) \cite{li2013spontaneous}, Improved Chinese Academy of Sciences Micro-Expression Database (CASME II) \cite{yan2014casme}, Spontaneous Micro-Facial Movement Database (SAMM) \cite{davison2018samm}, Chinese Academy of Sciences Macro-Expression and Micro-Expression Database (CAS(ME)$^2$) \cite{qu2018cas} have been released. However, it is hard to collect and annotate large-scale micro-expression data with uncontrolled scenarios due to its subtle, rapid, and involuntary nature.

Recently, several large-scale databases have been developed to provide massive data with spontaneous facial affects and in-the-wild conditions, such as Real-World Affective Face Database (RAF-DB) \cite{li2018reliable}, Large-Scale Face Expression in-the-Wild dataset (ExpW) \cite{zhang2018facial2}, EMOTIC \cite{kosti2019context}, AffectNet \cite{mollahosseini2017affectnet}, and EmotioNet \cite{fabian2016emotionet}. 

Concerning graph-based FAA, it is available to find and select suitable databases with corresponding metadata, such as landmarks, AU labels, and dynamics, for different graph representation purposes. However, existing databases also have some shortcomings. On the one hand, not enough accurate AU annotations are provided by in-the-wild databases, limiting AUs' role in FAA. On the other hand, there is still a blank in the dynamic large-scale affective database field, so it is hard to use temporal information to generate affective graph representation. Finally, databases about natural and spontaneous facial affects in a continuous domain need more attention instead of discrete categories.

\subsection{Facial Expression Recognition}
Facial expression recognition (FER), or macro-expression recognition, has been working on basic facial affects classification. An inevitable trend of FER is that the research focus has shifted from the early posed facial affects in controlled conditions to the recent spontaneous facial affects in real scenarios. In other words, recognizing the former is considered a solved problem for FAA methods, including graph-based FER, which can be corroborated from the results in Table \ref{tab:fer}. For example, the performance on the CK+ database is very close to $100\%$ \cite{kotsia2006facial,rivera2015spatiotemporal,liu2021video,liu2021sg}. 

\begin{table*}[!htbp]
  \centering
  \scriptsize
  \begin{threeparttable}
  \caption{Performance summary of representative graph-based FER methods}
  \label{tab:fer}
  \setlength{\tabcolsep}{0.8mm}{
  \begin{tabular}{|l|c|c|c|c|c|c|c|l|l|c|}
    \toprule[1.5pt]
    \multirow{2}*{References} & \multicolumn{2}{c|}{Prep.$^1$} & \multicolumn{3}{c|}{Representation$^2$} & \multicolumn{2}{c|}{Reasoning} & \makecell[c]{\multirow{2}*{Posed Database$^3$}} & \makecell[c]{\multirow{2}*{Wild Database}} & \multirow{2}*{Validation$^4$} \\ \cmidrule[0.5pt](r){2-3} \cmidrule[0.5pt](r){4-6} \cmidrule[0.5pt](r){7-8} 
    & B.B. & LM & Category & Node & Edge & Model & Classifier &  &  &  \\ \midrule[1.0pt]
    \citet{kotsia2006facial} & $\circ$ & 104 & $\mathcal{S:L}$ & $\mathbb{C}$ & $\bigtriangleup$ & \emph{Tracking} & \emph{SVM} & CK \emph{ar}: 0.997 & \makecell[c]{/} & LO CV \\ \midrule[0.5pt]
    \citet{zafeiriou2008discriminant} & $\bullet$ & / & $\mathcal{S:R}$ & \emph{NM} & \emph{L} & \emph{Tracking}  & \emph{EGM} & CK \emph{ar}: 0.971 & \makecell[c]{/} & LO CV \\ \midrule[0.5pt]
    \citet{mohseni2014facial} & $\circ$ & 50 & $\mathcal{S:L}$ & $\mathbb{C}$ & \emph{M}+$\mathbb{E}$ & \emph{Tracking} & \emph{Adaboost} & MMI \emph{ar}: 0.877 & \makecell[c]{/} & 10F CV \\ \midrule[0.5pt]
    \citet{rivera2015spatiotemporal} & $\bullet$ & / & $\mathcal{ST:R}$ & \emph{Histograms} & \emph{DNG} & \emph{Tracking} & \emph{SVM} & \makecell[l]{CK+ \emph{ar}: 1; MMI \emph{ar}: 0.976;\\ Oulu \emph{ar}: 0.984} & \makecell[c]{/} & 10F CV \\ \midrule[0.5pt]
    \citet{yao2015capturing} & $\bullet$ & / & $\mathcal{S:R}$ & \emph{LBP} & \emph{L} & / & \makecell[c]{\emph{SVM} \\\emph{CNN}} & \makecell[c]{/} & \makecell[l]{SFEW \emph{ar}: 0.5538;\\ AFEW \emph{ar}: 0.5380;} & HO \\ \midrule[0.5pt]
    \citet{dapogny2018confidence} & $\bullet$ & 49 & $\mathcal{S:L}$ & $\mathbb{C}$+\emph{HOG} & $\bigtriangleup$ & \emph{AE} & \emph{RF} & \makecell[l]{CK+ \emph{ar}: 0.934;\\ BU-4DFE \emph{ar}: 0.750} & SFEW \emph{ar}: 0.371 & \makecell[c]{10F-SI CV\\ /HO} \\ \midrule[0.5pt]
    \citet{zhong2019graph} & $\bullet$ & 46 & $\mathcal{S:L}$ & \emph{Gabor} & \emph{F}+$\mathbb{E}$ & \emph{RNN} & \emph{Softmax} & \makecell[l]{CK+ \emph{ar}: 0.9827; MMI \emph{ar}: 0.9444;\\ Oulu \emph{ar}: 0.9306} & \makecell[c]{/} & 10F-SI CV \\ \midrule[0.5pt]
    \citet{chen2020label} & $\bullet$ & / & $\mathcal{AU:B}$ & $\Phi$ & \emph{KNN} & / & \emph{Softmax} & \makecell[l]{CK+ \emph{ar}: 0.9308;\\ MMI \emph{ar}: 0.7049;\\ Oulu \emph{ar}: 0.6385} & \makecell[l]{SFEW \emph{ar}: 0.5650;\\ RAF \emph{ar}: 0.8553;\\ AffNet: \emph{ar}: 0.5935} & \makecell[c]{CD\\ (AffNet+RAF)} \\ \midrule[0.5pt]
    \citet{liu2019facial} & $\bullet$ & 68 & $\mathcal{S:L}$ & $\mathbb{C}$+\emph{HOG} & $\bigtriangleup$ & \emph{CNN} & \emph{Softmax} & CK+ \emph{ar}: 0.9767; MMI \emph{ar}: 0.8011 & SFEW \emph{ar}: 0.5536 & \makecell[c]{10F-SI CV\\ /HO} \\ \midrule[0.5pt]
    \citet{xie2020adversarial} & $\bullet$ & / & $\mathcal{S:R}$ & \emph{ResNet} & \emph{K-means} & \emph{GCN} & \emph{Softmax} & \makecell[l]{CK+ \emph{ar}: 0.8527;\\ JAFEE \cite{lyons1999automatic} \emph{ar}: 0.6150} & \makecell[l]{SFEW \emph{ar}: 0.5643;\\ FER2013 \emph{ar}: 0.5895\\ ExpW \emph{ar}: 68.50\%} & CD (RAF) \\ \midrule[0.5pt]
    \citet{zhou2020facial} & $\bullet$ & 34 & $\mathcal{ST:L}$ & $\mathbb{C}$+\emph{HOG} & \emph{M}+$\mathbb{H}$ & \emph{GCN} & \emph{Softmax} & CK+ \emph{ar}: 0.9863; Oulu \emph{ar}: 0.8723 & \makecell[c]{/} & 10F-SI CV \\ \midrule[0.5pt]
    \citet{liu2021video} & $\bullet$ & / & $\mathcal{ST:F}$ & \emph{VGG} & \emph{F} & \emph{GCN} & \emph{LSTM} & \makecell[l]{CK+ \emph{ar}: 0.9954; MMI \emph{ar}: 0.8589;\\ Oulu \emph{ar}: 0.9104} & \makecell[c]{/} & 10F-SI CV \\ \midrule[0.5pt]
    \citet{zhou2020learning} & $\bullet$ & 44 & $\mathcal{ST:L}$ & $\mathbb{C}$+\emph{HOG} & \emph{L}+$\mathbb{H}$ & \emph{GCN} & \emph{Softmax} & CK+ \emph{ar}: 0.9892; Oulu \emph{ar}: 0.8750 & AFEW \emph{ar}: 0.4512 & \makecell[c]{10F-SI CV\\ /HO} \\ \midrule[0.5pt]
    \citet{cui2020knowledge} & $\circ$ & / & $\mathcal{AU:B}$ & \emph{CNN} & $\Phi$ & \emph{DBN} & \emph{Softmax} & \makecell[l]{CK+ \emph{ar}: 0.9759; BP4D \emph{ar}: 0.8382;\\ MMI \emph{ar}: 0.8490} & EmotioNet \emph{ar}: 0.9555 & 5F-SI CV\\ \midrule[0.5pt]
    \citet{liu2021sg} & $\bullet$ & 40 & $\mathcal{S:L}$ & \makecell[c]{$\mathbb{C}$+\emph{HOG}+\emph{Gabor}\\ /$\mathbb{C}$+CNN} & \makecell[c]{\emph{L}+$\mathbb{H}$\\ /\emph{L}+$\mathbb{E}$} & \emph{GCN} & \emph{Softmax} & \makecell[l]{CK+ \emph{ar}: 0.9923; MMI \emph{ar}: 0.8575;\\ Oulu \emph{ar}: 0.9088} & \makecell[l]{SFEW \emph{ar}: 0.5742\\ RAF \emph{ar}: 0.8713} & \makecell[c]{10F-SI CV\\ /HO} \\ \midrule[0.5pt]
    \citet{chen2021learning} & $\circ$ & / & $\mathcal{X}$ & \emph{HOG} & \emph{KNN} & \emph{GSC} & \emph{SVM} & \makecell[l]{CK+ (JAFEE) \emph{ar}: 0.7171;\\ JAFEE (CK+) \emph{ar}: 0.5667;\\ Oulu (CK+) \emph{ar}: 0.4834;\\ CK+ (Oulu) \emph{ar}: 0.7756} & \makecell[c]{/} & CD \\ \midrule[0.5pt]
    \citet{shirian2021dynamic} & $\circ$ & 68 & $\mathcal{ST:F}$ & $\mathbb{C}$ & \emph{F} & \emph{GCN} & \emph{Softmax} & \makecell[l]{RML \cite{wang2008recognizing} \emph{ar}: 0.9411;\\ eNTERFACE \cite{martin2006enterface} \emph{ar}: 0.8749;\\ RAVDESS \cite{livingstone2018ryerson} \emph{ar}: 0.8565} & \makecell[c]{/} & 10F CV \\ \midrule[0.5pt]
    \citet{jin2021learning} & $\bullet$ & 68 & $\mathcal{S:R}$ & \emph{LBP/AE} & \emph{F/M} & \emph{GCN} & \emph{Softmax} & CK+ \emph{ar}: 0.9432; Oulu \emph{ar}: 0.7328 & RAF \emph{ar}: 0.5825 & \makecell[c]{10F-SI CV\\ /HO} \\ \midrule[0.5pt]
    \citet{zhao2021geometry} & $\bullet$ & 68 & $\mathcal{S:L}$ & $\mathbb{C}$+\emph{CNN} & \emph{M} & \emph{GCN} & \emph{Softmax} & \makecell[l]{CK+ (RAF) \emph{ar}: 0.9320;\\ MMI (RAF) \emph{ar}: 0.7439;\\ Oulu (RAF) \emph{ar}: 0.6302} & \makecell[l]{SFEW \emph{ar}: 0.5711\\ RAF \emph{ar}: 0.8752} & \makecell[c]{10F-SI CV\\ /HO} \\ \midrule[0.5pt]
    \citet{rao2021facial} & $\bullet$ & 68 & $\mathcal{S:L}$ & $\mathbb{C}$ & \emph{M} & \emph{GCN} & \emph{Softmax} & \makecell[l]{CK+ \emph{ar}: 0.9868;\\ JAFEE \emph{ar}: 0.9664} & \makecell[l]{FER2013 \emph{ar}: 0.7806\\ RAF \emph{ar}: 0.8695} & \makecell[c]{10F/LO CV\\ /HO} \\
    \bottomrule[1.5pt]
  \end{tabular}}
  \begin{tablenotes}
    \item[1] Prep.: processing; B.B.: bounding boxes; $\bullet$ = Yes, $\circ$ = No; LM: landmarks.
    \item[2] $\mathcal{S/ST/AU/X}$: \{spatial; spatio-temporal; AU-level; sample-level\} representation; $\mathcal{:L/R/F}$: \{landmark; region; frame\}-level graph; $\mathbb{C}$: landmark coordinates; $\Phi$: label distributions; $\bigtriangleup$: triangulation; \emph{L/M/F}: \{learning; manual; full\} connections; $\mathbb{E}$: Euclidean distance; $\mathbb{H}$: Hop distance.
    \item[3] \emph{ar}: average accuracy rate; DB1 (DB2): train on database 2, test on database 1.
    \item[4] CV: cross validation; LO: leave-one-subject-out; HO: holdout validation; 10F: 10-flods; SI: subject independent; CD (DB): cross database validation (training database).
  \end{tablenotes}
  \end{threeparttable}
\end{table*}

From the view of the representation, spatial graphs and spatio-temporal graphs are dominant. Specifically, hand-crafted features (e.g., \emph{LBP} \cite{yao2015capturing,jin2021learning}, \emph{Gabor} \cite{zhong2019graph,liu2021sg}, \emph{HOG} \cite{liu2019facial,zhou2020facial,chen2021learning}) or deep-based features (e.g., \emph{CNN} \cite{liu2021sg,zhao2021geometry}, \emph{VGG} \cite{liu2021video}, \emph{ResNet} \cite{xie2020assisted}) are employed to enhance the node representation similar to many non-graph FER methods \cite{zhao2007dynamic,zhang2017facial}. For reasoning approaches, early studies prefer to capture the relations of an individual node from predefined graph structures using tracking strategies \cite{kotsia2006facial,zafeiriou2008discriminant,mohseni2014facial,rivera2015spatiotemporal} or general machine learning models (e.g., \emph{RF} \cite{dapogny2018confidence}, \emph{RNN} \cite{zhong2019graph}, \emph{CNN} \cite{liu2019facial}). In the latest work, \emph{GCNs} become one of the mainstream choices in the latest work and show state-of-the-art performances on posed and in-the-wild databases \cite{xie2020assisted,liu2021sg,jin2021learning,zhao2021geometry,rao2021facial}. Another observation is that the framework of combining the spatio-temporal graph representation and \emph{GNNs} is getting more attention in FER studies \cite{zhou2020facial,zhou2020learning,liu2021video,shirian2021dynamic}.  

Although many graph-based studies have shown improvements in predicting facial affects, FER still has some potential topics. One thing is that the goal of existing methods stays on classifying basic facial affects. No study of graph-based methods to recognize compound affects (or mixture affects), whose labels are provided by recent databases like RAF-DB and EmotioNet, is reported. One possible solution is introducing AU-level graph representations that can describe fine-grained macro-expressions with closer inter-class distances. The other topic is practical graph-based representations due to the big gap between the performance of current methods and the acceptable result in practice when analyzing in-the-wild facial affects. In addition, since existing databases lack sufficient dynamic annotated samples, the evaluation of spatio-temporal graphs in large-scale conditions remains explored.

\subsection{Action Unit Detection}
The AU detection (AUD) facilitates a comprehensive analysis of the facial affect and is typically formulated as a multi-task problem that learns a two-class classification model for each AU. It can expand the recognition categories of macro-expressions through the AU combination \cite{kaltwang2015latent} and can be used as a pre-step to enhance the recognition of micro-expressions \cite{lei2021micro}. Compared with graph-based FER, the wide usage of graph structures has a long history in AUD \cite{martinez2017automatic} and has played a more dominant role. Table \ref{tab:aud} summarizes graph-based AUD methods including the performance comparison.

\begin{table*}[!htbp]
  \centering
  \scriptsize
  \begin{threeparttable}
  \caption{Performance summary of representative graph-based AUD methods}
  \label{tab:aud}
  \setlength{\tabcolsep}{1.7mm}{
  \begin{tabular}{|l|c|c|c|c|c|c|c|l|c|}
    \toprule[1.5pt]
    \multirow{2}*{References} & \multicolumn{2}{c|}{Prep.$^1$} & \multicolumn{3}{c|}{Representation$^2$} & \multicolumn{2}{c|}{Reasoning} & \makecell[c]{\multirow{2}*{Database$^3$}} & \multirow{2}*{Validation$^4$} \\ \cmidrule[0.5pt](r){2-3} \cmidrule[0.5pt](r){4-6} \cmidrule[0.5pt](r){7-8} 
    & B.B. & LM & Category & Node & Edge & Model & Output &  &  \\ \midrule[1.0pt]
    \citet{tong2007facial}  & $\bullet$ & / & $\mathcal{AU:B}$ & \emph{Gabor} & $\Phi$ & \emph{DBN} & \emph{Adabst} & \makecell[l]{CK \emph{ar}: 0.9933;\\ MMI (CK) \emph{ar}: 0.939} & \makecell[c]{LO CV\\ /CD} \\ \midrule[0.5pt] 
    \citet{zhu2014multiple} & $\bullet$ & 49 & $\mathcal{AU:B}$ & $\mathbb{C}$+\emph{Gabor} & $\Phi$ & \emph{DBN} & \emph{MPE} & \makecell[l]{CK+ \emph{ar}: 90.48\%, \emph{f}$_1$: 0.7072; \\ DISFA \emph{ar}: 93.56\%, \emph{f}$_1$: 0.7095} & 2F, 10F CV \\ \midrule[0.5pt] 
    \citet{kaltwang2015latent} & $\circ$ & 66 & $\mathcal{S:L}$ & $\mathbb{C}$ & \emph{L} & \multicolumn{2}{c|}{\emph{LT}} & DISFA \emph{corr}: 0.43, \emph{mse}: 0.39, \emph{icc}: 0.36 & 9F CV \\ \midrule[0.5pt] 
    \citet{walecki2017deep}  & $\circ$ & / & $\mathcal{AU:M}$ & \emph{CNN} & \emph{L} & \emph{CRF} & \emph{\makecell[c]{Ordinal\\ Regression}} & \makecell[l]{FERA 2015 \cite{valstar2015fera} \emph{icc}: 0.63, \emph{mae}: 1.23; \\ DISFA \emph{icc}: 0.45, \emph{mae}: 0.61} & HO \\ \midrule[0.5pt] 
    \citet{corneanu2018deep} & $\bullet$ & / & $\mathcal{AU:M}$ & \emph{VGG} & $\Phi$ & \emph{RNN} & \emph{Sigmoid} & \makecell[l]{BP4D \emph{f}$_1$: 0.617; DISFA \emph{f}$_1$: 0.567} & 3F-SI CV \\ \midrule[0.5pt]
    \citet{dapogny2018confidence} & $\bullet$ & 49 & $\mathcal{S:L}$ & $\mathbb{C}$+\emph{HOG} & $\bigtriangleup$ & \emph{AE} & \emph{RF} & \makecell[l]{CK+ \emph{auc}: 0.953, \emph{f}$_1$: 0.788, \emph{nf}$_1$: 0.865;\\ BP4D: \emph{auc}: 0.727, \emph{f}$_1$: 0.557, \emph{nf}$_1$: 0.636;\\ DISFA \emph{auc}: 0.824, \emph{f}$_1$: 0.491} & 10F-SI CV \\ \midrule[0.5pt]
    \citet{li2019semantic} & $\bullet$ & 20 & $\mathcal{AU:M}$ & $\mathbb{C}$+\emph{VGG} & \emph{M+L} & \emph{GGNN} & \emph{FC} & \makecell[l]{BP4D \emph{auc}: 0.741, \emph{f}$_1$: 0.629;\\ DISFA \emph{auc}: 0.807, \emph{f}$_1$: 0.559} & No report \\ \midrule[0.5pt]
    \citet{niu2019multi} & $\bullet$ & / & $\mathcal{AU:M}$ & $\mathbb{W}$ & \emph{L} & \emph{GCN} & \emph{FC} & BP4D \emph{f}$_1$: 0.598; EmotioNet \emph{f}$_1$: 0.681 & 3F-SI CV/ HO \\ \midrule[0.5pt]
    \citet{liu2020relation} & $\bullet$ & 19 & $\mathcal{S:R}$ & \makecell[c]{\emph{CNN}\\+\emph{AE}} & \emph{M} & \emph{GCN} & \emph{FC} & \makecell[l]{BP4D \emph{auc}: 0.873, \emph{f}$_1$: 0.628;\\ DISFA \emph{auc}: 0.746, \emph{f}$_1$: 0.550} & 3F-SI CV \\ \midrule[0.5pt]
    \citet{fan2020facial}  & $\bullet$ & 20 & $\mathcal{S:R}$ & \emph{ResNet} & KNN+$\mathbb{E}$ & \emph{DG-CNN} & \emph{\makecell[c]{Heatmap\\ Regression}} & \makecell[l]{BP4D \emph{icc}: 0.72, \emph{mae}: 0.58;\\ DISFA \emph{icc}: 0.47, \emph{mae}: 0.20} & 3F-SI CV \\ \midrule[0.5pt]
    \citet{liu2019facial} & $\bullet$ & 68 & $\mathcal{S:L}$ & $\mathbb{C}$+\emph{HOG} & $\bigtriangleup$ & \emph{CNN} & \emph{FC} & CK+ \emph{auc}: 0.929 & 10F-SI CV \\ \midrule[0.5pt]
    \citet{zhang2020region} & $\bullet$ & 18 & $\mathcal{S:R}$ & \emph{HRN} & \emph{L} & \emph{GCN} & \emph{\makecell[c]{Heatmap\\ Regression}} & BP4D \emph{f}$_1$: 0.635; DISFA \emph{f}$_1$: 0.620 & 3F-SI CV \\ \midrule[0.5pt]
    \citet{cui2020label} & $\circ$ & 51 & $\mathcal{AU:B}$ & \emph{LBP} & $\Phi$ & \emph{DBN} & \makecell[c]{\emph{LR}\\\emph{/CNN}\\\emph{/SVM}} & \makecell[l]{CK+ \emph{f}$_1$: 0.830; BP4D \emph{f}$_1$: 0.687;\\ EmotionNet \emph{f}$_1$: 0.626;\\ MMI (CK+) \emph{f}$_1$: 0.532} & \makecell[c]{5F-SI CV\\ /CD} \\ \midrule[0.5pt]
    \citet{cui2020knowledge} & $\circ$ & / & $\mathcal{AU:B}$ & \emph{VGG} & $\Phi$ & \emph{DBN} & \emph{FC} & \makecell[l]{CK+ \emph{f}$_1$: 0.74; BP4D \emph{f}$_1$: 0.57;\\ MMI \emph{f}$_1$: 0.58} & 5F-SI CV \\ \midrule[0.5pt]
    \citet{song2021uncertain} & $\bullet$ & / & $\mathcal{AU:M}$ & \emph{ResNet} & \emph{Mask} & \emph{GAT} & \emph{Softmax} & \makecell[l]{BP4D \emph{ar}: 0.782, \emph{f}$_1$: 0.633;\\ DISFA \emph{ar}: 0.934, \emph{f}$_1$: 0.600} & 3F-SI CV \\ \midrule[0.5pt]
    \citet{song2021hybrid} & $\bullet$ & / & $\mathcal{AU:M}$ & \emph{ResNet} & $\Phi$ & \emph{Hybrid GNN} & \emph{Softmax} & BP4D \emph{f}$_1$: 0.634; DISFA \emph{f}$_1$: 0.610 & 3F-SI CV \\ \midrule[0.5pt] 
    \citet{song2021dynamic} & $\bullet$ & / & $\mathcal{AU:M}$ & \emph{ResNet} & $\Phi$ & \emph{GCN+LSTM} & \emph{FC} & \makecell[l]{FERA 2015 \emph{icc}: 0.72, \emph{mae}: 0.57;\\ DISFA \emph{icc}: 0.56, \emph{mae}: 0.22} & 3F-SI CV \\ \midrule[0.5pt] 
    \citet{chen2021cafgraph} & $\bullet$ & 68 & $\mathcal{ST:L}$ & \emph{DCT+CNN} & \emph{M+Mask} & \emph{GCN} & \emph{Softmax} & BP4D \emph{f}$_1$: 0.649; DISFA \emph{f}$_1$: 0.658 & 3F-SI CV \\
    \bottomrule[1.5pt]
  \end{tabular}}
  \begin{tablenotes}
    \item[1] Prep.: processing; B.B.: bounding boxes; $\bullet$ = Yes, $\circ$ = No; LM: landmarks.
    \item[2] $\mathcal{S/ST/AU}$: \{spatial; spatio-temporal; AU-level\} representation; $\mathcal{:L/R/B/M}$: \{landmark; region; label; map\}-level graph; $\mathbb{C}$: landmark coordinates; $\Phi$: label distributions; $\bigtriangleup$: triangulation; \emph{L/M}: \{learning; manual\} connections; $\mathbb{E}$: Euclidean distance.
    \item[3] \emph{ar}: average accuracy rate; \emph{f}$_1$: F1 score; \emph{nf}$_1$: F1-norm score; \emph{corr}: Pearson correlation coefficient; \emph{mae}: mean absolute error; \emph{mes}: mean squared error; \emph{icc}: intra-class correlation coefficient, ICC(3,1); \emph{auc}: area under the receiver operating characteristic curve; DB1 (DB2): train on database 2, test on database 1.
    \item[4] CV: cross validation; LO: leave-one-subject-out; (K)F: k-folds; SI: subject independent; HO: holdout validation; CD: cross database validation.
  \end{tablenotes}
  \end{threeparttable}
\end{table*}

Specifically, spatial graphs and AU-level graphs are equally popular in the representation part of AUD. Interestingly, no matter landmark-level or region-level, all the spatial graphs constructed in the listed AUD methods employed facial landmarks \cite{kaltwang2015latent,dapogny2018confidence,liu2020relation,fan2020facial,liu2019facial,zhang2020region}, even for the spatio-temporal graph \cite{chen2021cafgraph}. The possible reason is that the landmark information is helpful and practical for locating the facial areas where AUs may occur. In this setting, their node representations were close to that in spatial graphs of FER methods, which usually combined geometric coordinates with appearance features (e.g., \emph{HOG} \cite{dapogny2018confidence,liu2019facial}). Although some AUD methods using AU-level graphs also exploited traditional features (e.g., \emph{Gabor} \cite{tong2007facial,zhu2014multiple}, \emph{LBP} \cite{cui2020label}) or deep features (e.g., \emph{VGG} \cite{corneanu2018deep}) to introduce appearance information, their graph representations were initialized from the AU label distribution of the training set. Thus, the \emph{DBN} model has become popular in the relational reasoning stage \cite{tong2007facial,zhu2014multiple,cui2020label}. Another similar trend to graph-based FER is that \emph{GNNs} have been widely utilized to learn the latent dependency among individual AUs in recent studies, such as \emph{GCN} \cite{niu2019multi,liu2020relation,zhang2020region,chen2021cafgraph}, \emph{GAT} \cite{song2021uncertain}, \emph{GGNN} \cite{li2019semantic}, and \emph{DG-CNN} \cite{fan2020facial}. But the difference is that \emph{fully-connected (FC)} layers \cite{li2019semantic,niu2019multi,liu2020relation,liu2019facial} or regression models \cite{walecki2017deep,fan2020facial,zhang2020region} are often applied for predicting labels instead of \emph{softmax} classifier \cite{song2021uncertain,chen2021cafgraph}. 

A particular line of AUD research analyzes the facial affects by estimating the AU intensities, which could have greater information value in understanding complex affective states \cite{zhi2020comprehensive}. Even though a few attempts in estimating AU intensities based on graph structures have existed \cite{kaltwang2015latent,corneanu2018deep,fan2020facial}, the study of using the latest spatio-temporal graph representations and \emph{GNNs} has not been reported. Another big challenge in AUD is few and imbalanced samples. Recent graph-based methods using transfer learning \cite{niu2019multi,cui2020label} or uncertainty learning \cite{song2021uncertain} were proposed to address this problem. They showed an advantage of the graph-based method in this topic and are helpful to implement AUD in large-scale unlabeled data.

\subsection{Micro-Expression Recognition}
Micro-expressions are fleeting and involuntary facial affects that people usually exhibit in high stake situations when attempting to conceal or mask their true feelings \cite{zhao2019automatic}. The earliest well-known studies came from \cite{haggard1966micromomentary} as well as \cite{ekman1969nonverbal}. Generally, a micro-expression lasts only 1/25 to 1/2 seconds long and is too subtle and fleeting for an untrained person to perceive. Therefore, developing an automatic micro-expression recognition (MER) system is valuable in reading human hidden affective states. Besides the short duration, low intensity and localization characteristics also make it challenging. 

To this end, graph-based MER methods have been designed to address the above challenges and have become appealing in the past two years \cite{liu2018sparse}, especially in 2020 \cite{lei2020novel,xie2020assisted}. Table \ref{tab:mer} lists the reported performance of a few representative recent studies of graph-based MER. These methods fall into the landmark-level spatial graph \cite{liu2018sparse,lei2020novel} and the AU-level graph \cite{xie2020assisted} in terms of representation types. For the former, their idea is to use landmarks to locate and analyze specific facial areas to deal with the local response and the subtleness of micro-expressions. The latter aims to infer the AU relationship to improve the final performance. The difference in processing ideas is also reflected in the reasoning procedure. Approaches like \emph{GSC} \cite{liu2018sparse} and variant \emph{CNNs} \cite{lei2020novel} are exploited in the landmark-level graph to integrate the individual node feature representations. In comparison, \emph{GCNs} are employed to learn an optimal graph structure of the AU dependency knowledge from training data and make predictions. Nevertheless, one common thing is that all the methods consider the local appearance in a spatio-temporal way by using \emph{optical-flow} or \emph{DNNs}. 

\begin{table*}[!htbp]
  \centering
  \scriptsize
  \begin{threeparttable}
  \caption{Performance summary of representative graph-based MER methods}
  \label{tab:mer}
  \setlength{\tabcolsep}{1.5mm}{
  \begin{tabular}{|l|c|c|c|c|c|c|c|l|c|}
    \toprule[1.5pt]
    \multirow{2}*{References} & \multicolumn{2}{c|}{Prep.$^1$} & \multicolumn{3}{c|}{Representation$^2$} & \multicolumn{2}{c|}{Reasoning} & \makecell[c]{\multirow{2}*{Database$^3$}} & \multirow{2}*{Validation$^4$} \\ \cmidrule[0.5pt](r){2-3} \cmidrule[0.5pt](r){4-6} \cmidrule[0.5pt](r){7-8} 
    & B.B. & LM & Category & Node & Edge & Model & Classifier &  &  \\ \midrule[1.0pt]
    \citet{liu2018sparse}  & $\bullet$ & 66 & $\mathcal{S:R}$ & \emph{Optical-flow} & \emph{KNN} & \emph{\makecell[c]{GSC}} & \emph{SVM} & \makecell[l]{SMIC (3 cl.) \emph{ar}: 0.6795, \emph{f}$_1$: 0.6844;\\ CASME I \cite{yan2013casme} (4 cl.) \emph{ar}: 0.7219, \emph{f}$_1$: 0.7236;\\ CASME II (5 cl.) \emph{ar}: 0.6356, \emph{f}$_1$: 0.6364} & LO CV \\ \midrule[0.5pt]
    \citet{lei2020novel} & $\bullet$ & 28 & $\mathcal{S:L}$ & \emph{TCN} & \emph{F} & \emph{Graph-TCN} & \emph{Softmax} & \makecell[l]{CASME II (5 cl.) \emph{ar}: 0.7398, \emph{f}$_1$: 0.7246;\\ SAMM (5 cl.) \emph{ar}: 0.7500, \emph{f}$_1$: 0.6985;\\ SAMM (4 cl.) \emph{ar}: 0.8050, \emph{f}$_1$: 0.7657} & LO CV \\ \midrule[0.5pt]
    \citet{xie2020assisted} & $\bullet$ & / & $\mathcal{AU:M}$ & \emph{CNN} & \emph{L} & \emph{GCN} & \emph{Softmax} & \makecell[l]{CASME II (3 cl.) \emph{ar}: 0.712, \emph{f}$_1$: 0.355;\\ CASME II (7 cl.) \emph{ar}: 0.561, \emph{f}$_1$: 0.394;\\ SAMM (3 cl.) \emph{ar}: 0.702, \emph{f}$_1$: 0.433;\\ SAMM (8 cl.) \emph{ar}: 0.523, \emph{f}$_1$: 0.357\\ SMIC (CASME II) (3 cl.) \emph{ar}: 0.344, \emph{f}$_1$: 0.319;\\ SMIC (SAMM) (3 cl.) \emph{ar}: 0.451, \emph{f}$_1$: 0.309} & \makecell[c]{LO CV\\ /CD} \\ \midrule[0.5pt]
    \citet{kumar2021micro} & $\bullet$ & 51 & $\mathcal{S:L}$ & $\mathbb{C}$+\emph{Optical-flow} & \emph{M+L} & \emph{GCN+GAT} & \emph{Softmax} & \makecell[l]{CASME II (3 cl.) \emph{ar}: 0.8966, \emph{f}$_1$: 0.8695;\\ CASME II (5 cl.) \emph{ar}: 0.8130, \emph{f}$_1$: 0.7090;\\ SAMM (3 cl.) \emph{ar}: 0.8872, \emph{f}$_1$: 0.8118;\\ SAMM (5 cl.) \emph{ar}: 0.8824, \emph{f}$_1$: 0.8279} & LO CV \\ \midrule[0.5pt]
    \citet{lei2021micro} & $\bullet$ & 30 & $\mathcal{AU:B}$ & \emph{Embedding} & $\Phi$ & \emph{GCN} & \emph{Softmax} & \makecell[l]{CASME II (4 cl.) \emph{ar}: 0.8080, \emph{f}$_1$: 0.7871;\\ CASME II (5 cl.) \emph{ar}: 0.7427, \emph{f}$_1$: 0.7047;\\ SAMM (4 cl.) \emph{ar}: 0.8239, \emph{f}$_1$: 0.7735;\\ SAMM (5 cl.) \emph{ar}: 0.7426, \emph{f}$_1$: 0.7045} & LO CV \\
    \bottomrule[1.5pt]
  \end{tabular}}
  \begin{tablenotes}
    \item[1] Prep.: processing; B.B.: bounding boxes; $\bullet$ = Yes, $\circ$ = No; LM: landmarks.
    \item[2] $\mathcal{S/AU}$: \{spatial; AU-level\} representation; $\mathcal{:L/B/M}$: \{landmark; label; map\}-level graph; $\mathbb{C}$: landmark coordinates; $\Phi$: label distributions; \emph{L/F}: \{learning; fully\} connections; $\mathbb{E}$: Euclidean distance; $\mathbb{G}$: Gradient using slope equation.
    \item[3] \emph{ar}: average accuracy rate; \emph{f}$_1$: F1 score; \emph{war}:weighted average recall; \emph{wf}$_1$: weighted F1 score; \emph{uar}: unweighted average recall; (N) cl.: (N) affective classes; DB1 (DB2): train on database 2, test on database 1.
    \item[4] CV: cross validation; LO: leave-one-subject-out; CD: cross database validation.
  \end{tablenotes}
  \end{threeparttable}
\end{table*}

A problem in graph-based MER is the lack of large-scale in-the-wild data. The small sample size limits the AU-level graph representation that relies on initializing the AU relationship from the AU label distribution of the training set. The lab-controlled data make it difficult to follow the trend in FER studies, which generalizes the graph-based FAA methods in real-world scenarios. However, the analysis of uncontrolled micro-expressions is fundamental because micro-expressions and macro-expressions can co-occur in many real cases. For example, the furrowing on the forehead slightly and quickly when smiling indicates the true feeling \cite{ekman1969nonverbal}. Since the evolutionary appearance information is crucial for the micro-expression analysis, building a spatio-temporal graph representation that can model the duration and the dynamic of micro-expressions is also a helpful but unexplored topic.

\subsection{Special Tasks}
The graph-based methods also play a vital role in several special FAA tasks, such as pain detection \cite{kaltwang2015latent}, non-basic affect recognition \cite{zhang2019context,he2019image}, occluded FER \cite{dapogny2018confidence,zhou2020learning}, and multi-modal affect recognition \cite{chen2019efficient,chien2020cross}. Table \ref{tab:other} summarizes the latest graph-based FAA methods for special tasks. Their node representations and edge initialization strategies for graph constructions in this field are similar to those in graph-based FER, MER, and AUD methods. While for the reasoning step, \emph{GCN} is the top-1 option. This observation implies that the framework of the graph-based method discussed in this paper can be easily extended to many other FAA tasks and promote performance improvement.

\begin{table*}[!htbp]
  \centering
  \scriptsize
  \begin{threeparttable}
  \caption{Performance summary of graph-based methods for special FAA tasks}
  \label{tab:other}
  \setlength{\tabcolsep}{1.2mm}{
  \begin{tabular}{|l|c|c|c|c|c|c|c|l|c|}
    \toprule[1.5pt]
    \multirow{2}*{References} & \multicolumn{2}{c|}{Prep.$^1$} & \multicolumn{3}{c|}{Representation$^2$} & \multicolumn{2}{c|}{Reasoning} & \makecell[c]{\multirow{2}*{Database$^3$}} & \multirow{2}*{Validation$^4$} \\ \cmidrule[0.5pt](r){2-3} \cmidrule[0.5pt](r){4-6} \cmidrule[0.5pt](r){7-8} 
    & B.B. & LM & Category & Node & Edge & Model & Output &  &  \\ \midrule[1.0pt]
    \citet{kaltwang2015latent}  & $\circ$ & 66 & $\mathcal{S:L}$ & $\mathbb{C}$ & \emph{L} & \multicolumn{2}{c|}{\emph{LT}} & \makecell[l]{ShoulderPain \cite{lucey2011painful} \emph{corr}: 0.23, \emph{mse}: 0.60} & 8F CV \\ \midrule[0.5pt]
    \citet{dapogny2018confidence} & $\bullet$ & 49 & $\mathcal{S:L}$ & $\mathbb{C}$+\emph{HOG} & $\bigtriangleup$ & \emph{AE} & \emph{RF} & \makecell[l]{CK+ (eyes occluded) \emph{ar}: 0.879;\\ CK+ (mouth occluded) \emph{ar}: 0.727} & 10F-SI CV \\ \midrule[0.5pt]
    \citet{zhang2019context} & $\bullet$ & / & $\mathcal{S:R}$ & \emph{VGG} & \emph{L} & \emph{GCN} & \emph{Softmax}/\emph{FC} & \makecell[l]{EMOTIC (26 cl.) \emph{prc}: 0.2842; EMOTIC (VAD) \emph{er}: 0.9} & HO \\ \midrule[0.5pt]
    \citet{chen2019efficient} & $\bullet$ & 68 & $\mathcal{ST:L}$ & $\mathbb{C}$ & $\bigtriangleup$+\emph{M} & \emph{GCN} & \emph{FC} & \makecell[l]{CES \cite{ringeval2019avec} \emph{val\_ccc}: 0.515, \emph{aro\_ccc}: 0.513} & HO \\ \midrule[0.5pt]
    \citet{he2019image} & $\circ$ & / & $\mathcal{X}$ & \emph{Embedding} & \emph{M+L} & \emph{GCN} & \emph{Softmax} & \makecell[l]{FlickLDL \cite{borth2013large} \emph{ar}: 0.691; TwiterLDL \cite{yang2017joint} \emph{ar}: 0.758} & HO \\ \midrule[0.5pt]
    \citet{zhou2020learning} & $\bullet$ & 44 & $\mathcal{ST:L}$ & $\mathbb{C}$+\emph{HOG} & \emph{L}+$\mathbb{H}$ & \emph{GCN} & \emph{SoftMax} & \makecell[l]{CK+ (random occlusion) \emph{ar}: 0.9551;\\ Oulu (random occlusion) \emph{ar}: 0.8121;\\ AFEW (random occlusion) \emph{ar}: 0.4047} & \makecell[c]{10F-SI CV\\ /HO} \\ \midrule[0.5pt]
    \citet{chien2020cross} & $\circ$ & / & $\mathcal{X}$ & \emph{CNN} & \emph{\makecell[c]{Transfer\\ Knowledge}} & \emph{GCN} & \emph{FC} & \makecell[l]{Amigos \cite{correa2018amigos} (Ascertain) \emph{val\_uar}: 0.798, \emph{aro\_uar}: 0.679;\\ Ascertain \cite{subramanian2016ascertain} (Amigos) \emph{val\_uar}: 0.704, \emph{aro\_uar}: 0.569} & CD \\
    \bottomrule[1.5pt]
  \end{tabular}}
  \begin{tablenotes}
    \item[1] Prep.: processing; B.B.: bounding boxes; $\bullet$ = Yes, $\circ$ = No; LM: landmarks.
    \item[2] $\mathcal{S/ST/X}$: \{spatial; spatio-temporal; sample-level\} representation; $\mathcal{:L/R}$: \{landmark; region\}-level graph; $\mathbb{C}$: landmark coordinates; $\Phi$: label distributions; $\bigtriangleup$: triangulation; \emph{M/L}: \{manual/learning\} connections.
    \item[3] \emph{corr}: Pearson correlation coefficient; \emph{mes}: mean squared error; \emph{prc}: area under the precision recall curves; \emph{er}: average error rate; \emph{val/aro}: valance/arousal; \emph{ccc}: concordance correlation coefficient; \emph{uar}: unweighted average recall; (N) cl.: (N) affective classes; VAD: valence, arousal, dominance; DB1 (DB2): train on database 2, test on database 1; \emph{ar}: average accuracy rate; \emph{f}$_1$: F1 score; \emph{mif}$_1$: micro F1 score; \emph{maf}$_1$: macro F1 score.
    \item[4] CV: cross validation; (K)F: k-folds; SI: subject independent; HO: holdout validation; CD: cross database validation.
  \end{tablenotes}
  \end{threeparttable}
\end{table*}

\section{Open Directions}\label{sec:od}
Graph-based FAA methods have been dissected into fundamental components for elaboration and discussion in this review. When encoding facial affect into graphs, strategies vary according to node and edge elements. Relational reasoning approaches infer latent relationships or inherent dependencies of graph nodes in terms of space, time, and semantics. The category of graph representations will affect the technique choice of relational reasoning to a certain extent. 

Despite significant advances and numerous work, the graph-based FAA is still an appealing field with many open directions. Due to advantages in modeling and reasoning latent relationships of facial affects, graph-based methods may provide complementary information to help solve some challenges that non-graph-based approaches face. Also, the graph-based method has natural advantages or unexplored research potential in other topics. 

% Finally, the recent trend of developing FAA methods is to combine every procedure as an end-to-end learning pipeline. Thus, integrating the process of establishing landmark-level graphs into existing frameworks still needs to be studied.
% incorporating context regions or multiple face regions into graph nodes is an emerging topic.
% which we think is a promising direction.

\subsection{In-the-wild Scenarios}
Although many efforts have been made for graph-based FAA in natural conditions \cite{yao2015capturing,dapogny2018confidence,chen2020label,liu2019facial,xie2020assisted,zhou2020learning,niu2019multi,cui2020label}, even the state-of-the-art performance is far from actual applications. Factors like illumination, head pose, and part occlusion are challenging in constructing an effective graph representation. For one thing, significant illumination changes and head pose variations will impair the accuracy of face detection and registration, which is vital for establishing landmark-level graphs. ROI graphs without landmarks or NPI graphs \cite{yao2015capturing, xie2020adversarial} should be a possible direction to avoid this problem. Also, missing face parts resulting from camera view or context occlusion make it challenging to encode enough facial information and obtain meaningful connections in an affective graph. Pilot work \cite{dapogny2018confidence,zhou2020learning} has tried to exploit a sub-graph without masked facial parts or generate adaptive edge links to alleviate the influence. Unfortunately, there has still been a considerable performance decrease compared to normal conditions. Proposing more effective spatio-temporal graphs can account for these problems based on evolutional affective information.

\subsection{3D and 4D Facial Affects}
Using 3D and 4D face images might be another good topic because the 3D face shape provides additional depth information and dynamically contains subtle facial deformations. They are intuitively insensitive to pose and light changes. Some studies have transformed 3D faces into 2D images and generated graph representations \cite{dapogny2018confidence,corneanu2018deep,li2019semantic,fan2020facial}, but they have not fully taken advantage of the 3D data. Alternatively, non-graph-based \cite{behzad2020landmarks} and graph-based methods \cite{pei20093d} have been explored to conduct FAA directly on 3D or 4D faces. Since the 3D face mesh structure is naturally close to the graph structure, employing the graph representation and reasoning to handle 3D face images will promote the improvement of in-the-wild FAA. Besides, there is also a potential topic of using 3D and 4D data with graph-based methods, especially landmark-level graphs and \emph{GNNs}, in micro-expression recognition.

\subsection{Valence and Arousal}
Estimating the continuous dimension is a rising topic in FAA. Unlike discrete labels, Valence-Arousal (V-A) annotations describe a wider range of facial affects that are consistent with those in the real world. Large-scale FAA databases (Aff-Wild I \cite{zafeiriou2017aff}, II \cite{kollias2020analysing}) containing V-A annotations have been released to support the continuous FAA. Existing graph-based methods mainly perform the V-A measurement \cite{kaltwang2015latent,walecki2017deep,fan2020facial} on lab-controlled databases except for a few studies like \cite{zhang2019context}. Recent graph-based methods have studied multi-label learning according to intrinsic mappings between facial affect categories and other annotations \cite{li2019semantic,cui2020label}. Such underlying assumptions can also be extended to the V-A measurement task, where AU-level graphs and \emph{DBNs}, as well as sample-level graphs, are potential directions.

\subsection{Context and Multi-modality}
Most current FAA methods only consider a single face in one image or sequence. However, people usually have affective behaviors, including facial expressions, body gestures, and emotionally speaking in real cases \cite{huang2018multimodal}. These facial affective displays are highly associated with context surroundings that include but are not limited to the affective behavior of other people in social interactions or inanimate objects. Existing studies like \cite{zhang2019context} and \cite{xie2020adversarial} have employed graph reasoning to infer relationships between the target face and other objects in the same image. Facial affects and other helpful contexts can be combined in a graph representation to perform the analysis on a fuller scope, such as gesture \cite{kaliouby2005real,liu2021imigue}. Another valuable topic is to introduce additional data channels that are multi-modality. Sample-level and spatio-temporal have also been successfully extended to process multi-modal affect analysis tasks with audio \cite{chen2019efficient} and physiological signal \cite{chien2020cross}, respectively, which shows a good research prospect. 

\subsection{Cross-database and Transfer Learning}
Insufficient annotations and imbalanced labels are two problems that limit the development of FAA research. One possible solution is to use graph-based transfer learning. Efforts like \cite{niu2019multi,cui2020label,song2021uncertain} have exploited the graph structure to solve this challenge in semi-supervision, label correction, generation, or uncertainty measurement. On the other hand, the performance of affective features extracted using graph-based representation and reasoning has been proved through cross-database validation in all FER \cite{chen2020label,xie2020adversarial}, AUD \cite{tong2007facial,cui2020label}, MER \cite{xie2020assisted}, and cross-corpus analysis \cite{chen2021learning,chien2020cross}. Specifically, the strength of distribution modeling of AU-level and sample-level graphs is valuable in improving the generalization capability of affective features. 

% or


% \appendices

% \section{}

% % use section* for acknowledgment

\ifCLASSOPTIONcompsoc
  \section*{Acknowledgments}
% \else
%   % regular IEEE prefers the singular form
% \fi

\textbf{Funding:} This work was supported by the China Scholarship Council [CSC, No.202006150091], and the Ministry of Education and
Culture of Finland for AI forum project.

The authors would like to thank Muzammil Behzad and Tuomas Varanka for providing materials and suggestions for the figures used in this paper.

\ifCLASSOPTIONcaptionsoff
  \newpage
\fi



%\bibliographystyle{IEEEtran}
%\begin{thebibliography}{1}
%\end{thebibliography}

{
\footnotesize
\bibliographystyle{IEEEtranN}
\bibliography{IEEEabrv, ref}

% Generated by IEEEtranN.bst, version: 1.14 (2015/08/26)
\begin{thebibliography}{174}
\providecommand{\natexlab}[1]{#1}
\providecommand{\url}[1]{#1}
\csname url@samestyle\endcsname
\providecommand{\newblock}{\relax}
\providecommand{\bibinfo}[2]{#2}
\providecommand{\BIBentrySTDinterwordspacing}{\spaceskip=0pt\relax}
\providecommand{\BIBentryALTinterwordstretchfactor}{4}
\providecommand{\BIBentryALTinterwordspacing}{\spaceskip=\fontdimen2\font plus
\BIBentryALTinterwordstretchfactor\fontdimen3\font minus
  \fontdimen4\font\relax}
\providecommand{\BIBforeignlanguage}[2]{{%
\expandafter\ifx\csname l@#1\endcsname\relax
\typeout{** WARNING: IEEEtranN.bst: No hyphenation pattern has been}%
\typeout{** loaded for the language `#1'. Using the pattern for}%
\typeout{** the default language instead.}%
\else
\language=\csname l@#1\endcsname
\fi
#2}}
\providecommand{\BIBdecl}{\relax}
\BIBdecl

\bibitem[Mehrabian et~al.(1971)]{mehrabian1971silent}
A.~Mehrabian \emph{et~al.}, \emph{Silent messages}.\hskip 1em plus 0.5em minus
  0.4em\relax Wadsworth Belmont, CA, 1971, vol.~8, no. 152.

\bibitem[Mehrabian and Russell(1974)]{mehrabian1974approach}
A.~Mehrabian and J.~A. Russell, \emph{An approach to environmental
  psychology.}\hskip 1em plus 0.5em minus 0.4em\relax the MIT Press, 1974.

\bibitem[Darwin and Prodger(1998)]{darwin1998expression}
C.~Darwin and P.~Prodger, \emph{The expression of the emotions in man and
  animals}.\hskip 1em plus 0.5em minus 0.4em\relax Oxford University Press,
  USA, 1998.

\bibitem[Gazzaniga et~al.(2014)Gazzaniga, Ivry, and
  Mangun]{gazzaniga2014cognitive}
M.~S. Gazzaniga, R.~B. Ivry, and G.~Mangun, \emph{Cognitive Neuroscience. The
  Biology of the Mind, (2014)}.\hskip 1em plus 0.5em minus 0.4em\relax Norton:
  New York, 2014.

\bibitem[Picard et~al.(2001)Picard, Vyzas, and Healey]{picard2001toward}
R.~W. Picard, E.~Vyzas, and J.~Healey, ``Toward machine emotional intelligence:
  Analysis of affective physiological state,'' \emph{IEEE Transactions on
  Pattern Analysis and Machine Intelligence}, vol.~23, no.~10, pp. 1175--1191,
  2001.

\bibitem[Jack et~al.(2012)Jack, Garrod, Yu, Caldara, and
  Schyns]{jack2012facial}
R.~E. Jack, O.~G. Garrod, H.~Yu, R.~Caldara, and P.~G. Schyns, ``Facial
  expressions of emotion are not culturally universal,'' \emph{National Academy
  of Sciences}, vol. 109, no.~19, pp. 7241--7244, 2012.

\bibitem[Sariyanidi et~al.(2014)Sariyanidi, Gunes, and
  Cavallaro]{sariyanidi2014automatic}
E.~Sariyanidi, H.~Gunes, and A.~Cavallaro, ``Automatic analysis of facial
  affect: A survey of registration, representation, and recognition,''
  \emph{IEEE Transactions on Pattern Analysis and Machine Intelligence},
  vol.~37, no.~6, pp. 1113--1133, 2014.

\bibitem[Calvo et~al.(2018)Calvo, Guti{\'e}rrez-Garc{\'\i}a, and
  Del~L{\'\i}bano]{calvo2018makes}
M.~G. Calvo, A.~Guti{\'e}rrez-Garc{\'\i}a, and M.~Del~L{\'\i}bano, ``What makes
  a smiling face look happy? visual saliency, distinctiveness, and affect,''
  \emph{Psychological Research}, vol.~82, no.~2, pp. 296--309, 2018.

\bibitem[Tavakolian and Hadid(2019)]{tavakolian2019spatiotemporal}
M.~Tavakolian and A.~Hadid, ``A spatiotemporal convolutional neural network for
  automatic pain intensity estimation from facial dynamics,''
  \emph{International Journal of Computer Vision}, vol. 127, no.~10, pp.
  1413--1425, 2019.

\bibitem[Zhang et~al.(2018{\natexlab{a}})Zhang, Luo, Loy, and
  Tang]{zhang2018facial2}
Z.~Zhang, P.~Luo, C.~C. Loy, and X.~Tang, ``From facial expression recognition
  to interpersonal relation prediction,'' \emph{International Journal of
  Computer Vision}, vol. 126, no.~5, pp. 550--569, 2018.

\bibitem[Zhao et~al.(2020)Zhao, Peng, Tian, Kapadia, and
  Metaxas]{zhao2020towards}
L.~Zhao, X.~Peng, Y.~Tian, M.~Kapadia, and D.~N. Metaxas, ``Towards
  image-to-video translation: A structure-aware approach via multi-stage
  generative adversarial networks,'' \emph{International Journal of Computer
  Vision}, vol. 128, no.~10, pp. 2514--2533, 2020.

\bibitem[Valstar et~al.(2017)Valstar, S{\'a}nchez-Lozano, Cohn, Jeni, Girard,
  Zhang, Yin, and Pantic]{valstar2017fera}
M.~F. Valstar, E.~S{\'a}nchez-Lozano, J.~F. Cohn, L.~A. Jeni, J.~M. Girard,
  Z.~Zhang, L.~Yin, and M.~Pantic, ``Fera 2017-addressing head pose in the
  third facial expression recognition and analysis challenge,'' in \emph{IEEE
  International Conference on Automatic Face \& Gesture Recognition
  (FG)}.\hskip 1em plus 0.5em minus 0.4em\relax IEEE, 2017, pp. 839--847.

\bibitem[Dhall et~al.(2020)Dhall, Sharma, Goecke, and Gedeon]{dhall2020emotiw}
A.~Dhall, G.~Sharma, R.~Goecke, and T.~Gedeon, ``Emotiw 2020: Driver gaze,
  group emotion, student engagement and physiological signal based
  challenges,'' in \emph{International Conference on Multimodal Interaction
  (ICMI)}, 2020, pp. 784--789.

\bibitem[Zafeiriou et~al.(2017)Zafeiriou, Kollias, Nicolaou, Papaioannou, Zhao,
  and Kotsia]{zafeiriou2017aff}
S.~Zafeiriou, D.~Kollias, M.~A. Nicolaou, A.~Papaioannou, G.~Zhao, and
  I.~Kotsia, ``Aff-wild: valence and arousal'in-the-wild'challenge,'' in
  \emph{IEEE Conference on Computer Vision and Pattern Recognition Workshops
  (CVPRW)}, 2017, pp. 34--41.

\bibitem[Kollias et~al.(2021)Kollias, Kotsia, Hajiyev, and
  Zafeiriou]{kollias2021analysing}
D.~Kollias, I.~Kotsia, E.~Hajiyev, and S.~Zafeiriou, ``Analysing affective
  behavior in the second abaw2 competition,'' \emph{arXiv preprint
  arXiv:2106.15318}, 2021.

\bibitem[Benitez-Quiroz et~al.(2017)Benitez-Quiroz, Srinivasan, Feng, Wang, and
  Martinez]{benitez2017emotionet}
C.~F. Benitez-Quiroz, R.~Srinivasan, Q.~Feng, Y.~Wang, and A.~M. Martinez,
  ``Emotionet challenge: Recognition of facial expressions of emotion in the
  wild,'' \emph{arXiv preprint arXiv:1703.01210}, 2017.

\bibitem[Ringeval et~al.(2019)Ringeval, Schuller, Valstar, Cummins, Cowie,
  Tavabi, Schmitt, Alisamir, Amiriparian, Messner, et~al.]{ringeval2019avec}
F.~Ringeval, B.~Schuller, M.~Valstar, N.~Cummins, R.~Cowie, L.~Tavabi,
  M.~Schmitt, S.~Alisamir, S.~Amiriparian, E.-M. Messner \emph{et~al.}, ``Avec
  2019 workshop and challenge: state-of-mind, detecting depression with ai, and
  cross-cultural affect recognition,'' in \emph{International on Audio/Visual
  Emotion Challenge and Workshop}, 2019, pp. 3--12.

\bibitem[Stappen et~al.(2020)Stappen, Baird, Rizos, Tzirakis, Du, Hafner,
  Schumann, Mallol-Ragolta, Schuller, Lefter, et~al.]{stappen2020muse}
L.~Stappen, A.~Baird, G.~Rizos, P.~Tzirakis, X.~Du, F.~Hafner, L.~Schumann,
  A.~Mallol-Ragolta, B.~W. Schuller, I.~Lefter \emph{et~al.}, ``Muse 2020
  challenge and workshop: Multimodal sentiment analysis, emotion-target
  engagement and trustworthiness detection in real-life media: Emotional car
  reviews in-the-wild,'' in \emph{International on Multimodal Sentiment
  Analysis in Real-life Media Challenge and Workshop}, 2020, pp. 35--44.

\bibitem[Zhao and Pietikainen(2007)]{zhao2007dynamic}
G.~Zhao and M.~Pietikainen, ``Dynamic texture recognition using local binary
  patterns with an application to facial expressions,'' \emph{IEEE Transactions
  on Pattern Analysis and Machine Intelligence}, vol.~29, no.~6, pp. 915--928,
  2007.

\bibitem[Friesen and Ekman(1978)]{friesen1978facial}
E.~Friesen and P.~Ekman, ``Facial action coding system: a technique for the
  measurement of facial movement,'' \emph{Palo Alto}, vol.~3, no.~2, p.~5,
  1978.

\bibitem[Zhong et~al.(2019)Zhong, Bai, Li, Chen, Li, and Liu]{zhong2019graph}
L.~Zhong, C.~Bai, J.~Li, T.~Chen, S.~Li, and Y.~Liu, ``A graph-structured
  representation with brnn for static-based facial expression recognition,'' in
  \emph{IEEE International Conference on Automatic Face and Gesture Recognition
  (FG)}.\hskip 1em plus 0.5em minus 0.4em\relax IEEE, 2019, pp. 1--5.

\bibitem[Li et~al.(2019{\natexlab{a}})Li, Zhu, Zeng, Wang, and
  Lin]{li2019semantic}
G.~Li, X.~Zhu, Y.~Zeng, Q.~Wang, and L.~Lin, ``Semantic relationships guided
  representation learning for facial action unit recognition,'' in \emph{AAAI
  Conference on Artificial Intelligence (AAAI)}, vol.~33, no.~01, 2019, pp.
  8594--8601.

\bibitem[Zhang et~al.(2020)Zhang, Wang, and Yin]{zhang2020region}
Z.~Zhang, T.~Wang, and L.~Yin, ``Region of interest based graph convolution: A
  heatmap regression approach for action unit detection,'' in \emph{ACM
  International Conference on Multimedia (ACM MM)}, 2020, pp. 2890--2898.

\bibitem[Song et~al.(2021{\natexlab{a}})Song, Cui, Wang, Zheng, and
  Ji]{song2021dynamic}
T.~Song, Z.~Cui, Y.~Wang, W.~Zheng, and Q.~Ji, ``Dynamic probabilistic graph
  convolution for facial action unit intensity estimation,'' in
  \emph{Proceedings of the IEEE/CVF Conference on Computer Vision and Pattern
  Recognition (CVPR)}, 2021, pp. 4845--4854.

\bibitem[Ekman(1994)]{ekman1994strong}
P.~Ekman, ``Strong evidence for universals in facial expressions: a reply to
  russell's mistaken critique.'' \emph{Psychological Bulletin}, vol. 115,
  no.~2, pp. 268--287, 1994.

\bibitem[Cohn and De~la Torre(2015)]{cohn2015automated}
J.~F. Cohn and F.~De~la Torre, \emph{Automated face analysis for affective
  computing.}\hskip 1em plus 0.5em minus 0.4em\relax Oxford University Press,
  2015.

\bibitem[Zarins and Kondrats(2015)]{zarins2015anatomy}
U.~Zarins and S.~Kondrats, \emph{Anatomy for sculptors: understanding the human
  figure}.\hskip 1em plus 0.5em minus 0.4em\relax Anatomy Next, Incorporated,
  2015.

\bibitem[Bimler and Paramei(2006)]{bimler2006facial}
D.~L. Bimler and G.~V. Paramei, ``Facial-expression affective attributes and
  their configural correlates: components and categories,'' \emph{Spanish
  Journal of Psychology}, vol.~9, no.~1, p.~19, 2006.

\bibitem[Ekman(2002)]{ekman2002facial}
P.~Ekman, ``Facial action coding system (facs),'' \emph{A Human Face}, 2002.

\bibitem[Zhang et~al.(2017)Zhang, Huang, Du, and Wang]{zhang2017facial}
K.~Zhang, Y.~Huang, Y.~Du, and L.~Wang, ``Facial expression recognition based
  on deep evolutional spatial-temporal networks,'' \emph{IEEE Transactions on
  Image Processing}, vol.~26, no.~9, pp. 4193--4203, 2017.

\bibitem[Li et~al.(2020)Li, Lu, Li, Zhang, and Zhang]{li2020facial}
Y.~Li, G.~Lu, J.~Li, Z.~Zhang, and D.~Zhang, ``Facial expression recognition in
  the wild using multi-level features and attention mechanisms,'' \emph{IEEE
  Transactions on Affective Computing}, 2020.

\bibitem[Jacob and Stenger(2021)]{jacob2021facial}
G.~M. Jacob and B.~Stenger, ``Facial action unit detection with transformers,''
  in \emph{Proceedings of the IEEE/CVF Conference on Computer Vision and
  Pattern Recognition (CVPR)}, 2021, pp. 7680--7689.

\bibitem[Martinez et~al.(2017)Martinez, Valstar, Jiang, and
  Pantic]{martinez2017automatic}
B.~Martinez, M.~F. Valstar, B.~Jiang, and M.~Pantic, ``Automatic analysis of
  facial actions: A survey,'' \emph{IEEE Transactions on Affective Computing},
  vol.~10, no.~3, pp. 325--347, 2017.

\bibitem[Barrett(2017)]{barrett2017emotions}
L.~F. Barrett, \emph{How emotions are made: The secret life of the
  brain}.\hskip 1em plus 0.5em minus 0.4em\relax Houghton Mifflin Harcourt,
  2017.

\bibitem[Liu et~al.(2019)Liu, Zhang, Lin, and Wang]{liu2019facial}
Y.~Liu, X.~Zhang, Y.~Lin, and H.~Wang, ``Facial expression recognition via deep
  action units graph network based on psychological mechanism,'' \emph{IEEE
  Transactions on Cognitive and Developmental Systems}, vol.~12, no.~2, pp.
  311--322, 2019.

\bibitem[Cui et~al.(2020{\natexlab{a}})Cui, Song, Wang, and
  Ji]{cui2020knowledge}
Z.~Cui, T.~Song, Y.~Wang, and Q.~Ji, ``Knowledge augmented deep neural networks
  for joint facial expression and action unit recognition,'' \emph{Advances in
  Neural Information Processing Systems (NeurIPS)}, vol.~33, 2020.

\bibitem[Fan et~al.(2020)Fan, Lam, and Li]{fan2020facial}
Y.~Fan, J.~Lam, and V.~Li, ``Facial action unit intensity estimation via
  semantic correspondence learning with dynamic graph convolution,'' in
  \emph{AAAI Conference on Artificial Intelligence (AAAI)}, vol.~34, no.~07,
  2020, pp. 12\,701--12\,708.

\bibitem[Xie et~al.(2020{\natexlab{a}})Xie, Lo, Shuai, and
  Cheng]{xie2020assisted}
H.-X. Xie, L.~Lo, H.-H. Shuai, and W.-H. Cheng, ``Au-assisted graph attention
  convolutional network for micro-expression recognition,'' in \emph{ACM
  International Conference on Multimedia (ACM MM)}, 2020, pp. 2871--2880.

\bibitem[Dapogny et~al.(2018)Dapogny, Bailly, and
  Dubuisson]{dapogny2018confidence}
A.~Dapogny, K.~Bailly, and S.~Dubuisson, ``Confidence-weighted local expression
  predictions for occlusion handling in expression recognition and action unit
  detection,'' \emph{International Journal of Computer Vision}, vol. 126,
  no.~2, pp. 255--271, 2018.

\bibitem[Zhou et~al.(2020{\natexlab{a}})Zhou, Zhang, and Liu]{zhou2020learning}
J.~Zhou, X.~Zhang, and Y.~Liu, ``Learning the connectivity: Situational graph
  convolution network for facial expression recognition,'' in \emph{IEEE
  International Conference on Visual Communications and Image
  Processing}.\hskip 1em plus 0.5em minus 0.4em\relax IEEE, 2020, pp. 230--234.

\bibitem[Cui et~al.(2020{\natexlab{b}})Cui, Zhang, and Ji]{cui2020label}
Z.~Cui, Y.~Zhang, and Q.~Ji, ``Label error correction and generation through
  label relationships,'' in \emph{AAAI Conference on Artificial Intelligence
  (AAAI)}, vol.~34, no.~04, 2020, pp. 3693--3700.

\bibitem[Chen et~al.(2020)Chen, Wang, Chen, Shi, Geng, and Rui]{chen2020label}
S.~Chen, J.~Wang, Y.~Chen, Z.~Shi, X.~Geng, and Y.~Rui, ``Label distribution
  learning on auxiliary label space graphs for facial expression recognition,''
  in \emph{IEEE Conference on Computer Vision and Pattern Recognition (CVPR)},
  2020, pp. 13\,984--13\,993.

\bibitem[Corneanu et~al.(2016)Corneanu, Sim{\'o}n, Cohn, and
  Guerrero]{corneanu2016survey}
C.~A. Corneanu, M.~O. Sim{\'o}n, J.~F. Cohn, and S.~E. Guerrero, ``Survey on
  rgb, 3d, thermal, and multimodal approaches for facial expression
  recognition: History, trends, and affect-related applications,'' \emph{IEEE
  Transactions on Pattern Analysis and Machine Intelligence}, vol.~38, no.~8,
  pp. 1548--1568, 2016.

\bibitem[Kollias et~al.(2019)Kollias, Tzirakis, Nicolaou, Papaioannou, Zhao,
  Schuller, Kotsia, and Zafeiriou]{kollias2019deep}
D.~Kollias, P.~Tzirakis, M.~A. Nicolaou, A.~Papaioannou, G.~Zhao, B.~Schuller,
  I.~Kotsia, and S.~Zafeiriou, ``Deep affect prediction in-the-wild: Aff-wild
  database and challenge, deep architectures, and beyond,'' \emph{International
  Journal of Computer Vision}, vol. 127, no.~6, pp. 907--929, 2019.

\bibitem[Li and Deng(2020)]{li2020deep}
S.~Li and W.~Deng, ``Deep facial expression recognition: A survey,'' \emph{IEEE
  Transactions on Affective Computing}, 2020.

\bibitem[Goh et~al.(2020)Goh, Ng, Lim, and Sheikh]{goh2020micro}
K.~M. Goh, C.~H. Ng, L.~L. Lim, and U.~U. Sheikh, ``Micro-expression
  recognition: an updated review of current trends, challenges and solutions,''
  \emph{The Visual Computer}, vol.~36, no.~3, pp. 445--468, 2020.

\bibitem[Zhang et~al.(2018{\natexlab{b}})Zhang, Verma, Tjondronegoro, and
  Chandran]{zhang2018facial}
L.~Zhang, B.~Verma, D.~Tjondronegoro, and V.~Chandran, ``Facial expression
  analysis under partial occlusion: A survey,'' \emph{ACM Computing Surveys},
  vol.~51, no.~2, pp. 1--49, 2018.

\bibitem[Rouast et~al.(2019)Rouast, Adam, and Chiong]{rouast2019deep}
P.~V. Rouast, M.~Adam, and R.~Chiong, ``Deep learning for human affect
  recognition: Insights and new developments,'' \emph{IEEE Transactions on
  Affective Computing}, 2019.

\bibitem[Ben et~al.(2021)Ben, Ren, Zhang, Wang, Kpalma, Meng, and
  Liu]{ben2021video}
X.~Ben, Y.~Ren, J.~Zhang, S.-J. Wang, K.~Kpalma, W.~Meng, and Y.-J. Liu,
  ``Video-based facial micro-expression analysis: A survey of datasets,
  features and algorithms,'' \emph{IEEE Transactions on Pattern Analysis and
  Machine Intelligence}, 2021.

\bibitem[Ekman and Friesen(1971)]{ekman1971constants}
P.~Ekman and W.~V. Friesen, ``Constants across cultures in the face and
  emotion.'' \emph{Journal of Personality and Social Psychology}, vol.~17,
  no.~2, pp. 124--129, 1971.

\bibitem[Du et~al.(2014)Du, Tao, and Martinez]{du2014compound}
S.~Du, Y.~Tao, and A.~M. Martinez, ``Compound facial expressions of emotion,''
  \emph{Proceedings of the National Academy of Sciences}, vol. 111, no.~15, pp.
  E1454--E1462, 2014.

\bibitem[Sethu et~al.(2019)Sethu, Provost, Epps, Busso, Cummins, and
  Narayanan]{sethu2019ambiguous}
V.~Sethu, E.~M. Provost, J.~Epps, C.~Busso, N.~Cummins, and S.~Narayanan, ``The
  ambiguous world of emotion representation,'' \emph{arXiv preprint
  arXiv:1909.00360}, 2019.

\bibitem[Mehrabian(1995)]{mehrabian1995framework}
A.~Mehrabian, ``Framework for a comprehensive description and measurement of
  emotional states.'' \emph{Genetic, social, and general psychology
  monographs}, 1995.

\bibitem[Plutchik(1980)]{plutchik1980general}
R.~Plutchik, ``A general psychoevolutionary theory of emotion,'' in
  \emph{Theories of Emotion}.\hskip 1em plus 0.5em minus 0.4em\relax Elsevier,
  1980, pp. 3--33.

\bibitem[Greenwald et~al.(1989)Greenwald, Cook, and
  Lang]{greenwald1989affective}
M.~K. Greenwald, E.~W. Cook, and P.~J. Lang, ``Affective judgment and
  psychophysiological response: dimensional covariation in the evaluation of
  pictorial stimuli.'' \emph{Journal of Psychophysiology}, vol.~3, no.~1, pp.
  51--64, 1989.

\bibitem[Russell(1978)]{russell1978evidence}
J.~A. Russell, ``Evidence of convergent validity on the dimensions of affect.''
  \emph{Journal of Personality and Social Psychology}, vol.~36, no.~10, p.
  1152, 1978.

\bibitem[Soleymani et~al.(2015)Soleymani, Asghari-Esfeden, Fu, and
  Pantic]{soleymani2015analysis}
M.~Soleymani, S.~Asghari-Esfeden, Y.~Fu, and M.~Pantic, ``Analysis of eeg
  signals and facial expressions for continuous emotion detection,'' \emph{IEEE
  Transactions on Affective Computing}, vol.~7, no.~1, pp. 17--28, 2015.

\bibitem[{Yin} et~al.(2008){Yin}, {Chen}, {Sun}, {Worm}, and
  {Reale}]{yin2008high}
L.~{Yin}, X.~{Chen}, Y.~{Sun}, T.~{Worm}, and M.~{Reale}, ``A high-resolution
  3d dynamic facial expression database,'' in \emph{IEEE International
  Conference on Automatic Face and Gesture Recognition (FG)}, 2008, pp. 1--6.

\bibitem[Baltru{\v{s}}aitis et~al.(2016)Baltru{\v{s}}aitis, Robinson, and
  Morency]{baltruvsaitis2016openface}
T.~Baltru{\v{s}}aitis, P.~Robinson, and L.-P. Morency, ``Openface: an open
  source facial behavior analysis toolkit,'' in \emph{2016 IEEE Winter
  Conference on Applications of Computer Vision (WACV)}.\hskip 1em plus 0.5em
  minus 0.4em\relax IEEE, 2016, pp. 1--10.

\bibitem[Baltrusaitis et~al.(2018)Baltrusaitis, Zadeh, Lim, and
  Morency]{baltrusaitis2018openface}
T.~Baltrusaitis, A.~Zadeh, Y.~C. Lim, and L.-P. Morency, ``Openface 2.0: Facial
  behavior analysis toolkit,'' in \emph{2018 13th IEEE international conference
  on automatic face \& gesture recognition (FG 2018)}.\hskip 1em plus 0.5em
  minus 0.4em\relax IEEE, 2018, pp. 59--66.

\bibitem[Viola and Jones(2001)]{viola2001rapid}
P.~Viola and M.~Jones, ``Rapid object detection using a boosted cascade of
  simple features,'' in \emph{IEEE Conference on Computer Vision and Pattern
  Recognition (CVPR)}, vol.~1.\hskip 1em plus 0.5em minus 0.4em\relax IEEE,
  2001, pp. I--I.

\bibitem[Zhu and Ramanan(2012)]{zhu2012face}
X.~Zhu and D.~Ramanan, ``Face detection, pose estimation, and landmark
  localization in the wild,'' in \emph{IEEE Conference on Computer Vision and
  Pattern Recognition (CVPR)}.\hskip 1em plus 0.5em minus 0.4em\relax IEEE,
  2012, pp. 2879--2886.

\bibitem[Cootes et~al.(2001)Cootes, Edwards, and Taylor]{cootes2001active}
T.~F. Cootes, G.~J. Edwards, and C.~J. Taylor, ``Active appearance models,''
  \emph{IEEE Transactions on Pattern Analysis and Machine Intelligence},
  vol.~23, no.~6, pp. 681--685, 2001.

\bibitem[Zhang et~al.(2016)Zhang, Zhang, Li, and Qiao]{zhang2016joint}
K.~Zhang, Z.~Zhang, Z.~Li, and Y.~Qiao, ``Joint face detection and alignment
  using multitask cascaded convolutional networks,'' \emph{IEEE Signal
  Processing Letters}, vol.~23, no.~10, pp. 1499--1503, 2016.

\bibitem[Ranjan et~al.(2017)Ranjan, Patel, and Chellappa]{ranjan2017hyperface}
R.~Ranjan, V.~M. Patel, and R.~Chellappa, ``Hyperface: A deep multi-task
  learning framework for face detection, landmark localization, pose
  estimation, and gender recognition,'' \emph{IEEE Transactions on Pattern
  Analysis and Machine Intelligence}, vol.~41, no.~1, pp. 121--135, 2017.

\bibitem[Dong et~al.(2020)Dong, Yang, Wei, Weng, Sheikh, and
  Yu]{dong2020supervision}
X.~Dong, Y.~Yang, S.-E. Wei, X.~Weng, Y.~Sheikh, and S.-I. Yu, ``Supervision by
  registration and triangulation for landmark detection,'' \emph{IEEE
  Transactions on Pattern Analysis and Machine Intelligence}, 2020.

\bibitem[Bulat and Tzimiropoulos(2017)]{bulat2017far}
A.~Bulat and G.~Tzimiropoulos, ``How far are we from solving the 2d and 3d face
  alignment problem?(and a dataset of 230,000 3d facial landmarks),'' in
  \emph{IEEE International Conference on Computer Vision (ICCV)}, 2017, pp.
  1021--1030.

\bibitem[Masi et~al.(2018)Masi, Wu, Hassner, and Natarajan]{masi2018deep}
I.~Masi, Y.~Wu, T.~Hassner, and P.~Natarajan, ``Deep face recognition: A
  survey,'' in \emph{SIBGRAPI Conference on Graphics, Patterns and Images
  (SIBGRAPI)}.\hskip 1em plus 0.5em minus 0.4em\relax IEEE, 2018, pp. 471--478.

\bibitem[Wang et~al.(2018)Wang, Gao, Tao, Yang, and Li]{wang2018facial}
N.~Wang, X.~Gao, D.~Tao, H.~Yang, and X.~Li, ``Facial feature point detection:
  A comprehensive survey,'' \emph{Neurocomputing}, vol. 275, pp. 50--65, 2018.

\bibitem[Lucey et~al.(2010)Lucey, Cohn, Kanade, Saragih, Ambadar, and
  Matthews]{lucey2010extended}
P.~Lucey, J.~F. Cohn, T.~Kanade, J.~Saragih, Z.~Ambadar, and I.~Matthews, ``The
  extended cohn-kanade dataset (ck+): A complete dataset for action unit and
  emotion-specified expression,'' in \emph{IEEE Conference on Computer Vision
  and Pattern Recognition-Workshops}.\hskip 1em plus 0.5em minus 0.4em\relax
  IEEE, 2010, pp. 94--101.

\bibitem[Kumar and Bhanu(2021)]{kumar2021micro}
A.~J.~R. Kumar and B.~Bhanu, ``Micro-expression classification based on
  landmark relations with graph attention convolutional network,'' in
  \emph{Proceedings of the IEEE/CVF Conference on Computer Vision and Pattern
  Recognition (CVPR)}, 2021, pp. 1511--1520.

\bibitem[Kaltwang et~al.(2015)Kaltwang, Todorovic, and
  Pantic]{kaltwang2015latent}
S.~Kaltwang, S.~Todorovic, and M.~Pantic, ``Latent trees for estimating
  intensity of facial action units,'' in \emph{IEEE Conference on Computer
  Vision and Pattern Recognition (CVPR)}, 2015, pp. 296--304.

\bibitem[Zhao et~al.(2021)Zhao, Liu, Huang, Lun, and Lam]{zhao2021geometry}
R.~Zhao, T.~Liu, Z.~Huang, D.~P.-K. Lun, and K.~K. Lam, ``Geometry-aware facial
  expression recognition via attentive graph convolutional networks,''
  \emph{IEEE Transactions on Affective Computing}, 2021.

\bibitem[Mohseni et~al.(2014)Mohseni, Zarei, and Ramazani]{mohseni2014facial}
S.~Mohseni, N.~Zarei, and S.~Ramazani, ``Facial expression recognition using
  anatomy based facial graph,'' in \emph{IEEE International Conference on
  Systems, Man, and Cybernetics (SMC)}.\hskip 1em plus 0.5em minus 0.4em\relax
  IEEE, 2014, pp. 3715--3719.

\bibitem[Liu et~al.(2015)Liu, Zhang, Yan, Wang, Zhao, and Fu]{liu2015main}
Y.-J. Liu, J.-K. Zhang, W.-J. Yan, S.-J. Wang, G.~Zhao, and X.~Fu, ``A main
  directional mean optical flow feature for spontaneous micro-expression
  recognition,'' \emph{IEEE Transactions on Affective Computing}, vol.~7,
  no.~4, pp. 299--310, 2015.

\bibitem[Lei et~al.(2020)Lei, Li, Chen, and Li]{lei2020novel}
L.~Lei, J.~Li, T.~Chen, and S.~Li, ``A novel graph-tcn with a graph structured
  representation for micro-expression recognition,'' in \emph{ACM International
  Conference on Multimedia (ACM MM)}, 2020, pp. 2237--2245.

\bibitem[Pei and Zha(2009)]{pei20093d}
Y.~Pei and H.~Zha, ``3d facial expression editing based on the dynamic graph
  model,'' in \emph{IEEE International Conference on Multimedia and Expo
  (ICME)}.\hskip 1em plus 0.5em minus 0.4em\relax IEEE, 2009, pp. 1354--1357.

\bibitem[Kotsia and Pitas(2006)]{kotsia2006facial}
I.~Kotsia and I.~Pitas, ``Facial expression recognition in image sequences
  using geometric deformation features and support vector machines,''
  \emph{IEEE Transactions on Image Processing}, vol.~16, no.~1, pp. 172--187,
  2006.

\bibitem[Liu et~al.(2020{\natexlab{a}})Liu, Dong, Zhang, Wang, and
  Dang]{liu2020relation}
Z.~Liu, J.~Dong, C.~Zhang, L.~Wang, and J.~Dang, ``Relation modeling with graph
  convolutional networks for facial action unit detection,'' in
  \emph{International Conference on Multimedia Modeling (MMM)}.\hskip 1em plus
  0.5em minus 0.4em\relax Springer, 2020, pp. 489--501.

\bibitem[Jin et~al.(2021)Jin, Lai, and Jin]{jin2021learning}
X.~Jin, Z.~Lai, and Z.~Jin, ``Learning dynamic relationships for facial
  expression recognition based on graph convolutional network,'' \emph{IEEE
  Transactions on Image Processing}, vol.~30, pp. 7143--7155, 2021.

\bibitem[Liu et~al.(2021{\natexlab{a}})Liu, Li, and Lai]{liu2018sparse}
Y.-J. Liu, B.-J. Li, and Y.-K. Lai, ``Sparse mdmo: Learning a discriminative
  feature for spontaneous micro-expression recognition,'' \emph{IEEE
  Transactions on Affective Computing}, vol.~12, no.~1, pp. 254--261, 2021.

\bibitem[Yao et~al.(2015)Yao, Shao, Ma, and Chen]{yao2015capturing}
A.~Yao, J.~Shao, N.~Ma, and Y.~Chen, ``Capturing au-aware facial features and
  their latent relations for emotion recognition in the wild,'' in \emph{ACM on
  International Conference on Multimodal Interaction (ICMI)}, 2015, pp.
  451--458.

\bibitem[Zafeiriou and Pitas(2008)]{zafeiriou2008discriminant}
S.~Zafeiriou and I.~Pitas, ``Discriminant graph structures for facial
  expression recognition,'' \emph{IEEE Transactions on Multimedia}, vol.~10,
  no.~8, pp. 1528--1540, 2008.

\bibitem[Zhang et~al.(2019{\natexlab{a}})Zhang, Liang, and
  Ma]{zhang2019context}
M.~Zhang, Y.~Liang, and H.~Ma, ``Context-aware affective graph reasoning for
  emotion recognition,'' in \emph{IEEE International Conference on Multimedia
  and Expo (ICME)}.\hskip 1em plus 0.5em minus 0.4em\relax IEEE, 2019, pp.
  151--156.

\bibitem[Xie et~al.(2020{\natexlab{b}})Xie, Chen, Pu, Wu, and
  Lin]{xie2020adversarial}
Y.~Xie, T.~Chen, T.~Pu, H.~Wu, and L.~Lin, ``Adversarial graph representation
  adaptation for cross-domain facial expression recognition,'' in \emph{ACM
  International Conference on Multimedia (ACM MM)}, 2020, pp. 1255--1264.

\bibitem[Zhou et~al.(2020{\natexlab{b}})Zhou, Zhang, Liu, and
  Lan]{zhou2020facial}
J.~Zhou, X.~Zhang, Y.~Liu, and X.~Lan, ``Facial expression recognition using
  spatial-temporal semantic graph network,'' in \emph{IEEE International
  Conference on Image Processing (ICIP)}.\hskip 1em plus 0.5em minus
  0.4em\relax IEEE, 2020, pp. 1961--1965.

\bibitem[Chen et~al.(2019)Chen, Deng, Cheng, Wang, Jiang, and
  Sahli]{chen2019efficient}
H.~Chen, Y.~Deng, S.~Cheng, Y.~Wang, D.~Jiang, and H.~Sahli, ``Efficient
  spatial temporal convolutional features for audiovisual continuous affect
  recognition,'' in \emph{International on Audio/Visual Emotion Challenge and
  Workshop (AVEC)}, 2019, pp. 19--26.

\bibitem[Chen et~al.(2021{\natexlab{a}})Chen, Chen, Wang, Wang, and
  Liang]{chen2021cafgraph}
Y.~Chen, D.~Chen, Y.~Wang, T.~Wang, and Y.~Liang, ``Cafgraph: Context-aware
  facial multi-graph representation for facial action unit recognition,'' in
  \emph{Proceedings of the 29th ACM International Conference on Multimedia (ACM
  MM)}, 2021, pp. 1029--1037.

\bibitem[Rivera and Chae(2015)]{rivera2015spatiotemporal}
A.~R. Rivera and O.~Chae, ``Spatiotemporal directional number transitional
  graph for dynamic texture recognition,'' \emph{IEEE Transactions on Pattern
  Analysis and Machine Intelligence}, vol.~37, no.~10, pp. 2146--2152, 2015.

\bibitem[Liu et~al.(2021{\natexlab{b}})Liu, Zhang, and Zhou]{liu2021video}
D.~Liu, H.~Zhang, and P.~Zhou, ``Video-based facial expression recognition
  using graph convolutional networks,'' in \emph{International Conference on
  Pattern Recognition (ICPR)}.\hskip 1em plus 0.5em minus 0.4em\relax IEEE,
  2021, pp. 607--614.

\bibitem[Shirian et~al.(2021)Shirian, Tripathi, and Guha]{shirian2021dynamic}
A.~Shirian, S.~Tripathi, and T.~Guha, ``Dynamic emotion modeling with learnable
  graphs and graph inception network,'' \emph{IEEE Transactions on Multimedia},
  2021.

\bibitem[Tong et~al.(2007)Tong, Liao, and Ji]{tong2007facial}
Y.~Tong, W.~Liao, and Q.~Ji, ``Facial action unit recognition by exploiting
  their dynamic and semantic relationships,'' \emph{IEEE Transactions on
  Pattern Analysis and Machine Intelligence}, vol.~29, no.~10, pp. 1683--1699,
  2007.

\bibitem[Zhu et~al.(2014)Zhu, Wang, Yue, and Ji]{zhu2014multiple}
Y.~Zhu, S.~Wang, L.~Yue, and Q.~Ji, ``Multiple-facial action unit recognition
  by shared feature learning and semantic relation modeling,'' in
  \emph{International Conference on Pattern Recognition (ICPR)}.\hskip 1em plus
  0.5em minus 0.4em\relax IEEE, 2014, pp. 1663--1668.

\bibitem[Lei et~al.(2021)Lei, Chen, Li, and Li]{lei2021micro}
L.~Lei, T.~Chen, S.~Li, and J.~Li, ``Micro-expression recognition based on
  facial graph representation learning and facial action unit fusion,'' in
  \emph{Proceedings of the IEEE/CVF Conference on Computer Vision and Pattern
  Recognition (CVPR)}, 2021, pp. 1571--1580.

\bibitem[Corneanu et~al.(2018)Corneanu, Madadi, and Escalera]{corneanu2018deep}
C.~Corneanu, M.~Madadi, and S.~Escalera, ``Deep structure inference network for
  facial action unit recognition,'' in \emph{European Conference on Computer
  Vision (ECCV)}, 2018, pp. 298--313.

\bibitem[Niu et~al.(2019)Niu, Han, Shan, and Chen]{niu2019multi}
X.~Niu, H.~Han, S.~Shan, and X.~Chen, ``Multi-label co-regularization for
  semi-supervised facial action unit recognition,'' in \emph{International
  Conference on Neural Information Processing Systems (NeurIPS)}, vol.~32,
  2019, pp. 1--11.

\bibitem[Song et~al.(2021{\natexlab{b}})Song, Chen, Zheng, and
  Ji]{song2021uncertain}
T.~Song, L.~Chen, W.~Zheng, and Q.~Ji, ``Uncertain graph neural networks for
  facial action unit detection,'' in \emph{AAAI Conference on Artificial
  Intelligence (AAAI)}, 2021, pp. 1--10.

\bibitem[Song et~al.(2021{\natexlab{c}})Song, Cui, Zheng, and
  Ji]{song2021hybrid}
T.~Song, Z.~Cui, W.~Zheng, and Q.~Ji, ``Hybrid message passing with
  performance-driven structures for facial action unit detection,'' in
  \emph{Proceedings of the IEEE/CVF Conference on Computer Vision and Pattern
  Recognition (CVPR)}, 2021, pp. 6267--6276.

\bibitem[Walecki et~al.(2017)Walecki, Rudovic, Pavlovic, Schuller, and
  Pantic]{walecki2017deep}
R.~Walecki, O.~Rudovic, V.~Pavlovic, B.~Schuller, and M.~Pantic, ``Deep
  structured learning for facial expression intensity estimation,'' \emph{Image
  and Vision Computting}, vol. 259, pp. 143--154, 2017.

\bibitem[Chien et~al.(2020)Chien, Yang, and Lee]{chien2020cross}
W.-S. Chien, H.-C. Yang, and C.-C. Lee, ``Cross corpus physiological-based
  emotion recognition using a learnable visual semantic graph convolutional
  network,'' in \emph{ACM International Conference on Multimedia (ACM MM)},
  2020, pp. 2999--3006.

\bibitem[Chen et~al.(2021{\natexlab{b}})Chen, Song, and
  Zheng]{chen2021learning}
D.~Chen, P.~Song, and W.~Zheng, ``Learning transferable sparse representations
  for cross-corpus facial expression recognition,'' \emph{IEEE Transactions on
  Affective Computing}, 2021.

\bibitem[He and Jin(2019)]{he2019image}
T.~He and X.~Jin, ``Image emotion distribution learning with graph
  convolutional networks,'' in \emph{International Conference on Multimedia
  Retrieval (ICMR)}, 2019, pp. 382--390.

\bibitem[Zhao et~al.(2016)Zhao, Chu, and Zhang]{zhao2016deep}
K.~Zhao, W.-S. Chu, and H.~Zhang, ``Deep region and multi-label learning for
  facial action unit detection,'' in \emph{IEEE Conference on Computer Vision
  and Pattern Recognition (CVPR)}, 2016, pp. 3391--3399.

\bibitem[Oh et~al.(2018)Oh, See, Le~Ngo, Phan, and Baskaran]{oh2018survey}
Y.-H. Oh, J.~See, A.~C. Le~Ngo, R.~C.-W. Phan, and V.~M. Baskaran, ``A survey
  of automatic facial micro-expression analysis: databases, methods, and
  challenges,'' \emph{Frontiers in Psychology}, vol.~9, p. 1128, 2018.

\bibitem[Zhang et~al.(2014{\natexlab{a}})Zhang, Luo, Loy, and
  Tang]{zhang2014facial}
Z.~Zhang, P.~Luo, C.~C. Loy, and X.~Tang, ``Facial landmark detection by deep
  multi-task learning,'' in \emph{European conference on computer vision
  (ECCV)}.\hskip 1em plus 0.5em minus 0.4em\relax Springer, 2014, pp. 94--108.

\bibitem[Kakumanu and Bourbakis(2006)]{kakumanu2006local}
P.~Kakumanu and N.~Bourbakis, ``A local-global graph approach for facial
  expression recognition,'' in \emph{IEEE International Conference on Tools
  with Artificial Intelligence (ICTAI)}.\hskip 1em plus 0.5em minus 0.4em\relax
  IEEE, 2006, pp. 685--692.

\bibitem[Kazemi and Sullivan(2014)]{kazemi2014one}
V.~Kazemi and J.~Sullivan, ``One millisecond face alignment with an ensemble of
  regression trees,'' in \emph{IEEE Conference on Computer Vision and Pattern
  Recognition (CVPR)}, 2014, pp. 1867--1874.

\bibitem[Dalal and Triggs(2005)]{dalal2005histograms}
N.~Dalal and B.~Triggs, ``Histograms of oriented gradients for human
  detection,'' in \emph{IEEE Conference on Computer Vision and Pattern
  Recognition (CVPR)}, vol.~1.\hskip 1em plus 0.5em minus 0.4em\relax Ieee,
  2005, pp. 886--893.

\bibitem[Rao et~al.(2021)Rao, Li, Wang, Sun, and Chen]{rao2021facial}
T.~Rao, J.~Li, X.~Wang, Y.~Sun, and H.~Chen, ``Facial expression recognition
  with multi-sale graph convolutional networks,'' \emph{IEEE MultiMedia}, 2021.

\bibitem[Liu and Wechsler(2003)]{liu2003independent}
C.~Liu and H.~Wechsler, ``Independent component analysis of gabor features for
  face recognition,'' \emph{IEEE Transactions on Neural Networks}, vol.~14,
  no.~4, pp. 919--928, 2003.

\bibitem[Liu et~al.(2021{\natexlab{c}})Liu, Zhang, Zhou, and Fu]{liu2021sg}
Y.~Liu, X.~Zhang, J.~Zhou, and L.~Fu, ``Sg-dsn: A semantic graph-based
  dual-stream network for facial expression recognition,''
  \emph{Neurocomputing}, vol. 462, pp. 320--330, 2021.

\bibitem[Afzal et~al.(2009)Afzal, Sezgin, Gao, and
  Robinson]{afzal2009perception}
S.~Afzal, T.~M. Sezgin, Y.~Gao, and P.~Robinson, ``Perception of emotional
  expressions in different representations using facial feature points,'' in
  \emph{International Conference on Affective Computing and Intelligent
  Interaction and Workshops}.\hskip 1em plus 0.5em minus 0.4em\relax IEEE,
  2009, pp. 1--6.

\bibitem[Baltru{\v{s}}aitis et~al.(2010)Baltru{\v{s}}aitis, Riek, and
  Robinson]{baltruvsaitis2010synthesizing}
T.~Baltru{\v{s}}aitis, L.~D. Riek, and P.~Robinson, ``Synthesizing expressions
  using facial feature point tracking: how emotion is conveyed,'' in
  \emph{Proceedings of the 3rd international workshop on Affective interaction
  in natural environments}, 2010, pp. 27--32.

\bibitem[Lee et~al.(2019)Lee, Kim, Kim, Park, and Sohn]{lee2019context}
J.~Lee, S.~Kim, S.~Kim, J.~Park, and K.~Sohn, ``Context-aware emotion
  recognition networks,'' in \emph{Proceedings of the IEEE/CVF International
  Conference on Computer Vision (CVPR)}, 2019, pp. 10\,143--10\,152.

\bibitem[Wang et~al.(2020{\natexlab{a}})Wang, Peng, Yang, Meng, and
  Qiao]{wang2020region}
K.~Wang, X.~Peng, J.~Yang, D.~Meng, and Y.~Qiao, ``Region attention networks
  for pose and occlusion robust facial expression recognition,'' \emph{IEEE
  Transactions on Image Processing}, vol.~29, pp. 4057--4069, 2020.

\bibitem[Wang et~al.(2020{\natexlab{b}})Wang, Sun, Cheng, Jiang, Deng, Zhao,
  Liu, Mu, Tan, Wang, et~al.]{wang2020deep}
J.~Wang, K.~Sun, T.~Cheng, B.~Jiang, C.~Deng, Y.~Zhao, D.~Liu, Y.~Mu, M.~Tan,
  X.~Wang \emph{et~al.}, ``Deep high-resolution representation learning for
  visual recognition,'' \emph{IEEE Transactions on Pattern Analysis and Machine
  Intelligence}, 2020.

\bibitem[He et~al.(2016)He, Zhang, Ren, and Sun]{he2016deep}
K.~He, X.~Zhang, S.~Ren, and J.~Sun, ``Deep residual learning for image
  recognition,'' in \emph{IEEE Conference on Computer Vision and Pattern
  Recognition (CVPR)}, 2016, pp. 770--778.

\bibitem[Ren et~al.(2016)Ren, He, Girshick, and Sun]{ren2016faster}
S.~Ren, K.~He, R.~Girshick, and J.~Sun, ``Faster r-cnn: towards real-time
  object detection with region proposal networks,'' \emph{IEEE Transactions on
  Pattern Analysis and Machine Intelligence}, vol.~39, no.~6, pp. 1137--1149,
  2016.

\bibitem[Simonyan and Zisserman(2014)]{simonyan2014very}
K.~Simonyan and A.~Zisserman, ``Very deep convolutional networks for
  large-scale image recognition,'' \emph{arXiv preprint arXiv:1409.1556}, 2014.

\bibitem[Liu et~al.(2020{\natexlab{b}})]{liu2020phase}
Q.~Liu \emph{et~al.}, ``Phase space reconstruction driven spatio-temporal
  feature learning for dynamic facial expression recognition,'' \emph{IEEE
  Transactions on Affective Computing}, 2020.

\bibitem[Mikolov et~al.(2013)Mikolov, Sutskever, Chen, Corrado, and
  Dean]{mikolov2013distributed}
T.~Mikolov, I.~Sutskever, K.~Chen, G.~S. Corrado, and J.~Dean, ``Distributed
  representations of words and phrases and their compositionality,'' in
  \emph{Advances in neural information processing systems (NeurIPS)}, 2013, pp.
  3111--3119.

\bibitem[Tian et~al.(2001)Tian, Kanade, and Cohn]{tian2001recognizing}
Y.-I. Tian, T.~Kanade, and J.~F. Cohn, ``Recognizing action units for facial
  expression analysis,'' \emph{IEEE Transactions on Pattern Analysis and
  Machine Intelligence}, vol.~23, no.~2, pp. 97--115, 2001.

\bibitem[Berkes et~al.(2008)Berkes, Wood, and Pillow]{berkes2008characterizing}
P.~Berkes, F.~Wood, and J.~Pillow, ``Characterizing neural dependencies with
  copula models,'' in \emph{International Conference on Neural Information
  Processing Systems (NeurIPS)}, vol.~21, 2008, pp. 129--136.

\bibitem[Kemp and Tenenbaum(2008)]{kemp2008discovery}
C.~Kemp and J.~B. Tenenbaum, ``The discovery of structural form,''
  \emph{National Academy of Sciences}, vol. 105, no.~31, pp. 10\,687--10\,692,
  2008.

\bibitem[El~Kaliouby and Robinson(2005)]{el2005real}
R.~El~Kaliouby and P.~Robinson, ``Real-time inference of complex mental states
  from facial expressions and head gestures,'' in \emph{Real-time vision for
  human-computer interaction}.\hskip 1em plus 0.5em minus 0.4em\relax Springer,
  2005, pp. 181--200.

\bibitem[Baltru{\v{s}}aitis et~al.(2011)Baltru{\v{s}}aitis, McDuff, Banda,
  Mahmoud, El~Kaliouby, Robinson, and Picard]{baltruvsaitis2011real}
T.~Baltru{\v{s}}aitis, D.~McDuff, N.~Banda, M.~Mahmoud, R.~El~Kaliouby,
  P.~Robinson, and R.~Picard, ``Real-time inference of mental states from
  facial expressions and upper body gestures,'' in \emph{2011 IEEE
  International Conference on Automatic Face \& Gesture Recognition
  (FG)}.\hskip 1em plus 0.5em minus 0.4em\relax IEEE, 2011, pp. 909--914.

\bibitem[Li et~al.(2016)Li, Tarlow, Brockschmidt, and Zemel]{li2016gated}
Y.~Li, D.~Tarlow, M.~Brockschmidt, and R.~Zemel, ``Gated graph sequence neural
  networks,'' in \emph{International Conference on Learning Representations
  (ICLR)}, 2016, pp. 1--20.

\bibitem[Bai et~al.(2018)Bai, Kolter, and Koltun]{bai2018empirical}
S.~Bai, J.~Z. Kolter, and V.~Koltun, ``An empirical evaluation of generic
  convolutional and recurrent networks for sequence modeling,'' \emph{arXiv
  preprint arXiv:1803.01271}, 2018.

\bibitem[Wang et~al.(2019)Wang, Sun, Liu, Sarma, Bronstein, and
  Solomon]{wang2019dynamic}
Y.~Wang, Y.~Sun, Z.~Liu, S.~E. Sarma, M.~M. Bronstein, and J.~M. Solomon,
  ``Dynamic graph cnn for learning on point clouds,'' \emph{ACM Transactions On
  Graphics}, vol.~38, no.~5, pp. 1--12, 2019.

\bibitem[Li and Gupta(2018)]{li2018beyond}
Y.~Li and A.~Gupta, ``Beyond grids: Learning graph representations for visual
  recognition,'' in \emph{International Conference on Neural Information
  Processing Systems (NeurIPS)}, 2018, pp. 9245--9255.

\bibitem[Kipf and Welling(2017)]{kipf2017semi}
T.~N. Kipf and M.~Welling, ``Semi-supervised classification with graph
  convolutional networks,'' in \emph{International Conference on Learning
  Representations (ICLR)}, 2017.

\bibitem[Li et~al.(2019{\natexlab{b}})Li, Li, Zhang, and Wu]{li2019spatio}
B.~Li, X.~Li, Z.~Zhang, and F.~Wu, ``Spatio-temporal graph routing for
  skeleton-based action recognition,'' in \emph{AAAI Conference on Artificial
  Intelligence (AAAI)}, vol.~33, no.~01, 2019, pp. 8561--8568.

\bibitem[Defferrard et~al.(2016)Defferrard, Bresson, and
  Vandergheynst]{defferrard2016convolutional}
M.~Defferrard, X.~Bresson, and P.~Vandergheynst, ``Convolutional neural
  networks on graphs with fast localized spectral filtering,'' in
  \emph{International Conference on Neural Information Processing Systems
  (NeurIPS)}, 2016, pp. 3837--3845.

\bibitem[Gilmer et~al.(2017)Gilmer, Schoenholz, Riley, Vinyals, and
  Dahl]{gilmer2017neural}
J.~Gilmer, S.~S. Schoenholz, P.~F. Riley, O.~Vinyals, and G.~E. Dahl, ``Neural
  message passing for quantum chemistry,'' in \emph{International Conference on
  Machine Learning (ICML)}.\hskip 1em plus 0.5em minus 0.4em\relax PMLR, 2017,
  pp. 1263--1272.

\bibitem[Veli{\v{c}}kovi{\'c} et~al.(2018)Veli{\v{c}}kovi{\'c}, Cucurull,
  Casanova, Romero, Li{\`o}, and Bengio]{velivckovic2018graph}
P.~Veli{\v{c}}kovi{\'c}, G.~Cucurull, A.~Casanova, A.~Romero, P.~Li{\`o}, and
  Y.~Bengio, ``Graph attention networks,'' in \emph{International Conference on
  Learning Representations}, 2018.

\bibitem[Zhang et~al.(2019{\natexlab{b}})Zhang, Pal, Coates, and
  Ustebay]{zhang2019bayesian}
Y.~Zhang, S.~Pal, M.~Coates, and D.~Ustebay, ``Bayesian graph convolutional
  neural networks for semi-supervised classification,'' in \emph{AAAI
  Conference on Artificial Intelligence (AAAI)}, vol.~33, no.~01, 2019, pp.
  5829--5836.

\bibitem[Kanade et~al.(2000)Kanade, Cohn, and Tian]{kanade2000comprehensive}
T.~Kanade, J.~F. Cohn, and Y.~Tian, ``Comprehensive database for facial
  expression analysis,'' in \emph{IEEE International Conference on Automatic
  Face and Gesture Recognition (FG)}.\hskip 1em plus 0.5em minus 0.4em\relax
  IEEE, 2000, pp. 46--53.

\bibitem[Valstar and Pantic(2010)]{valstar2010induced}
M.~Valstar and M.~Pantic, ``Induced disgust, happiness and surprise: an
  addition to the mmi facial expression database,'' in \emph{International
  Workshop on EMOTION: Corpora for Research on Emotion and Affect}.\hskip 1em
  plus 0.5em minus 0.4em\relax Paris, France, 2010, p.~65.

\bibitem[Zhao et~al.(2011)Zhao, Huang, Taini, Li, and
  Pietik{\"a}Inen]{zhao2011facial}
G.~Zhao, X.~Huang, M.~Taini, S.~Z. Li, and M.~Pietik{\"a}Inen, ``Facial
  expression recognition from near-infrared videos,'' \emph{Image and Vision
  Computing}, vol.~29, no.~9, pp. 607--619, 2011.

\bibitem[Goodfellow et~al.(2015)Goodfellow, Erhan, Carrier, Courville, Mirza,
  Hamner, Cukierski, Tang, Thaler, Lee, et~al.]{goodfellow2013challenges}
I.~J. Goodfellow, D.~Erhan, P.~L. Carrier, A.~Courville, M.~Mirza, B.~Hamner,
  W.~Cukierski, Y.~Tang, D.~Thaler, D.-H. Lee \emph{et~al.}, ``Challenges in
  representation learning: A report on three machine learning contests,''
  \emph{Neural Networks}, vol.~64, pp. 59--63, 2015.

\bibitem[Dhall et~al.(2015)Dhall, Ramana~Murthy, Goecke, Joshi, and
  Gedeon]{dhall2015video}
A.~Dhall, O.~Ramana~Murthy, R.~Goecke, J.~Joshi, and T.~Gedeon, ``Video and
  image based emotion recognition challenges in the wild: Emotiw 2015,'' in
  \emph{ACM on International Conference on Multimodal Interaction (ICMI)},
  2015, pp. 423--426.

\bibitem[Dhall et~al.(2017)Dhall, Goecke, Ghosh, Joshi, Hoey, and
  Gedeon]{dhall2017individual}
A.~Dhall, R.~Goecke, S.~Ghosh, J.~Joshi, J.~Hoey, and T.~Gedeon, ``From
  individual to group-level emotion recognition: Emotiw 5.0,'' in \emph{ACM
  International Conference on Multimodal Interaction (ICMI)}, 2017, pp.
  524--528.

\bibitem[Mavadati et~al.(2013)Mavadati, Mahoor, Bartlett, Trinh, and
  Cohn]{mavadati2013disfa}
S.~M. Mavadati, M.~H. Mahoor, K.~Bartlett, P.~Trinh, and J.~F. Cohn, ``Disfa: A
  spontaneous facial action intensity database,'' \emph{IEEE Transactions on
  Affective Computing}, vol.~4, no.~2, pp. 151--160, 2013.

\bibitem[Yin et~al.(2006)Yin, Wei, Sun, Wang, and Rosato]{yin20063d}
L.~Yin, X.~Wei, Y.~Sun, J.~Wang, and M.~J. Rosato, ``A 3d facial expression
  database for facial behavior research,'' in \emph{International Conference on
  Automatic Face and Gesture Recognition (FG)}.\hskip 1em plus 0.5em minus
  0.4em\relax IEEE, 2006, pp. 211--216.

\bibitem[Zhang et~al.(2013)Zhang, Yin, Cohn, Canavan, Reale, Horowitz, and
  Liu]{zhang2013high}
X.~Zhang, L.~Yin, J.~F. Cohn, S.~Canavan, M.~Reale, A.~Horowitz, and P.~Liu,
  ``A high-resolution spontaneous 3d dynamic facial expression database,'' in
  \emph{IEEE International Conference and workshops on Automatic Face and
  Gesture Recognition (FG)}.\hskip 1em plus 0.5em minus 0.4em\relax IEEE, 2013,
  pp. 1--6.

\bibitem[Zhang et~al.(2014{\natexlab{b}})Zhang, Yin, Cohn, Canavan, Reale,
  Horowitz, Liu, and Girard]{zhang2014bp4d}
X.~Zhang, L.~Yin, J.~F. Cohn, S.~Canavan, M.~Reale, A.~Horowitz, P.~Liu, and
  J.~M. Girard, ``Bp4d-spontaneous: a high-resolution spontaneous 3d dynamic
  facial expression database,'' \emph{Image and Vision Computing}, vol.~32,
  no.~10, pp. 692--706, 2014.

\bibitem[Zhao and Li(2019)]{zhao2019automatic}
G.~Zhao and X.~Li, ``Automatic micro-expression analysis: open challenges,''
  \emph{Frontiers in Psychology}, vol.~10, p. 1833, 2019.

\bibitem[Li et~al.(2013)Li, Pfister, Huang, Zhao, and
  Pietik{\"a}inen]{li2013spontaneous}
X.~Li, T.~Pfister, X.~Huang, G.~Zhao, and M.~Pietik{\"a}inen, ``A spontaneous
  micro-expression database: Inducement, collection and baseline,'' in
  \emph{IEEE International Conference and Workshops on Automatic Face and
  Gesture Recognition (FG)}.\hskip 1em plus 0.5em minus 0.4em\relax IEEE, 2013,
  pp. 1--6.

\bibitem[Yan et~al.(2014)Yan, Li, Wang, Zhao, Liu, Chen, and Fu]{yan2014casme}
W.-J. Yan, X.~Li, S.-J. Wang, G.~Zhao, Y.-J. Liu, Y.-H. Chen, and X.~Fu,
  ``Casme ii: An improved spontaneous micro-expression database and the
  baseline evaluation,'' \emph{PloS One}, vol.~9, no.~1, p. e86041, 2014.

\bibitem[Davison et~al.(2018)Davison, Lansley, Costen, Tan, and
  Yap]{davison2018samm}
A.~K. Davison, C.~Lansley, N.~Costen, K.~Tan, and M.~H. Yap, ``Samm: A
  spontaneous micro-facial movement dataset,'' \emph{IEEE Transactions on
  Affective Computing}, vol.~9, no.~1, pp. 116--129, 2018.

\bibitem[Qu et~al.(2018)Qu, Wang, Yan, Li, Wu, and Fu]{qu2018cas}
F.~Qu, S.-J. Wang, W.-J. Yan, H.~Li, S.~Wu, and X.~Fu, ``Cas(me)$^2$ : A
  database for spontaneous macro-expression and micro-expression spotting and
  recognition,'' \emph{IEEE Transactions on Affective Computing}, vol.~9,
  no.~4, pp. 424--436, 2018.

\bibitem[Li and Deng(2019)]{li2018reliable}
S.~Li and W.~Deng, ``Reliable crowdsourcing and deep locality-preserving
  learning for unconstrained facial expression recognition,'' \emph{IEEE
  Transactions on Image Processing}, vol.~28, no.~1, pp. 356--370, 2019.

\bibitem[Kosti et~al.(2019)Kosti, Alvarez, Recasens, and
  Lapedriza]{kosti2019context}
R.~Kosti, J.~M. Alvarez, A.~Recasens, and A.~Lapedriza, ``Context based emotion
  recognition using emotic dataset,'' \emph{IEEE Transactions on Pattern
  Analysis and Machine Intelligence}, vol.~42, no.~11, pp. 2755--2766, 2019.

\bibitem[Mollahosseini et~al.(2019)Mollahosseini, Hasani, and
  Mahoor]{mollahosseini2017affectnet}
A.~Mollahosseini, B.~Hasani, and M.~H. Mahoor, ``Affectnet: A database for
  facial expression, valence, and arousal computing in the wild,'' \emph{IEEE
  Transactions on Affective Computing}, vol.~10, no.~1, pp. 18--31, 2019.

\bibitem[Fabian Benitez-Quiroz et~al.(2016)Fabian Benitez-Quiroz, Srinivasan,
  and Martinez]{fabian2016emotionet}
C.~Fabian Benitez-Quiroz, R.~Srinivasan, and A.~M. Martinez, ``Emotionet: An
  accurate, real-time algorithm for the automatic annotation of a million
  facial expressions in the wild,'' in \emph{IEEE Conference on Computer Vision
  and Pattern Recognition (CVPR)}, 2016, pp. 5562--5570.

\bibitem[Lyons et~al.(1999)Lyons, Budynek, and Akamatsu]{lyons1999automatic}
M.~J. Lyons, J.~Budynek, and S.~Akamatsu, ``Automatic classification of single
  facial images,'' \emph{IEEE Transactions on Pattern Analysis and Machine
  Intelligence}, vol.~21, no.~12, pp. 1357--1362, 1999.

\bibitem[Wang and Guan(2008)]{wang2008recognizing}
Y.~Wang and L.~Guan, ``Recognizing human emotional state from audiovisual
  signals,'' \emph{IEEE transactions on multimedia}, vol.~10, no.~5, pp.
  936--946, 2008.

\bibitem[Martin et~al.(2006)Martin, Kotsia, Macq, and
  Pitas]{martin2006enterface}
O.~Martin, I.~Kotsia, B.~Macq, and I.~Pitas, ``The enterface'05 audio-visual
  emotion database,'' in \emph{International Conference on Data Engineering
  Workshops}.\hskip 1em plus 0.5em minus 0.4em\relax IEEE, 2006, pp. 8--8.

\bibitem[Livingstone and Russo(2018)]{livingstone2018ryerson}
S.~R. Livingstone and F.~A. Russo, ``The ryerson audio-visual database of
  emotional speech and song (ravdess): A dynamic, multimodal set of facial and
  vocal expressions in north american english,'' \emph{PloS one}, vol.~13,
  no.~5, p. e0196391, 2018.

\bibitem[Valstar et~al.(2015)Valstar, Almaev, Girard, McKeown, Mehu, Yin,
  Pantic, and Cohn]{valstar2015fera}
M.~F. Valstar, T.~Almaev, J.~M. Girard, G.~McKeown, M.~Mehu, L.~Yin, M.~Pantic,
  and J.~F. Cohn, ``Fera 2015-second facial expression recognition and analysis
  challenge,'' in \emph{IEEE International Conference and Workshops on
  Automatic Face and Gesture Recognition (FG)}, vol.~6.\hskip 1em plus 0.5em
  minus 0.4em\relax IEEE, 2015, pp. 1--8.

\bibitem[Zhi et~al.(2020)Zhi, Liu, and Zhang]{zhi2020comprehensive}
R.~Zhi, M.~Liu, and D.~Zhang, ``A comprehensive survey on automatic facial
  action unit analysis,'' \emph{The Visual Computer}, vol.~36, no.~5, pp.
  1067--1093, 2020.

\bibitem[Haggard and Isaacs(1966)]{haggard1966micromomentary}
E.~A. Haggard and K.~S. Isaacs, ``Micromomentary facial expressions as
  indicators of ego mechanisms in psychotherapy,'' in \emph{Methods of Research
  in Psychotherapy}.\hskip 1em plus 0.5em minus 0.4em\relax Springer, 1966, pp.
  154--165.

\bibitem[Ekman and Friesen(1969)]{ekman1969nonverbal}
P.~Ekman and W.~V. Friesen, ``Nonverbal leakage and clues to deception,''
  \emph{Psychiatry}, vol.~32, no.~1, pp. 88--106, 1969.

\bibitem[Yan et~al.(2013)Yan, Wu, Liu, Wang, and Fu]{yan2013casme}
W.-J. Yan, Q.~Wu, Y.-J. Liu, S.-J. Wang, and X.~Fu, ``Casme database: a dataset
  of spontaneous micro-expressions collected from neutralized faces,'' in
  \emph{IEEE International Conference and workshops on Automatic Face and
  Gesture Recognition (FG)}.\hskip 1em plus 0.5em minus 0.4em\relax IEEE, 2013,
  pp. 1--7.

\bibitem[Lucey et~al.(2011)Lucey, Cohn, Prkachin, Solomon, and
  Matthews]{lucey2011painful}
P.~Lucey, J.~F. Cohn, K.~M. Prkachin, P.~E. Solomon, and I.~Matthews, ``Painful
  data: The unbc-mcmaster shoulder pain expression archive database,'' in
  \emph{IEEE International Conference on Automatic Face and Gesture Recognition
  (FG)}.\hskip 1em plus 0.5em minus 0.4em\relax IEEE, 2011, pp. 57--64.

\bibitem[Borth et~al.(2013)Borth, Ji, Chen, Breuel, and Chang]{borth2013large}
D.~Borth, R.~Ji, T.~Chen, T.~Breuel, and S.-F. Chang, ``Large-scale visual
  sentiment ontology and detectors using adjective noun pairs,'' in
  \emph{Proceedings of the ACM international conference on Multimedia (ACM
  MM)}, 2013, pp. 223--232.

\bibitem[Yang et~al.(2017)Yang, She, and Sun]{yang2017joint}
J.~Yang, D.~She, and M.~Sun, ``Joint image emotion classification and
  distribution learning via deep convolutional neural network.'' in
  \emph{Proceedings of the International Joint Conference on Artificial
  Intelligence (IJCAI)}, 2017, pp. 3266--3272.

\bibitem[Correa et~al.(2018)Correa, Abadi, Sebe, and Patras]{correa2018amigos}
J.~A.~M. Correa, M.~K. Abadi, N.~Sebe, and I.~Patras, ``Amigos: A dataset for
  affect, personality and mood research on individuals and groups,'' \emph{IEEE
  Transactions on Affective Computing}, 2018.

\bibitem[Subramanian et~al.(2016)Subramanian, Wache, Abadi, Vieriu, Winkler,
  and Sebe]{subramanian2016ascertain}
R.~Subramanian, J.~Wache, M.~K. Abadi, R.~L. Vieriu, S.~Winkler, and N.~Sebe,
  ``Ascertain: Emotion and personality recognition using commercial sensors,''
  \emph{IEEE Transactions on Affective Computing}, vol.~9, no.~2, pp. 147--160,
  2016.

\bibitem[Behzad et~al.(2020)Behzad, Vo, Li, and Zhao]{behzad2020landmarks}
M.~Behzad, N.~Vo, X.~Li, and G.~Zhao, ``Landmarks-assisted collaborative deep
  framework for automatic 4d facial expression recognition,'' in \emph{IEEE
  International Conference on Automatic Face and Gesture Recognition
  (FG)}.\hskip 1em plus 0.5em minus 0.4em\relax IEEE, 2020, pp. 1--5.

\bibitem[Kollias et~al.(2020)Kollias, Schulc, Hajiyev, and
  Zafeiriou]{kollias2020analysing}
D.~Kollias, A.~Schulc, E.~Hajiyev, and S.~Zafeiriou, ``Analysing affective
  behavior in the first abaw 2020 competition,'' in \emph{IEEE International
  Conference on Automatic Face and Gesture Recognition (FG)}.\hskip 1em plus
  0.5em minus 0.4em\relax IEEE Computer Society, 2020, pp. 794--800.

\bibitem[Huang et~al.(2018)Huang, Dhall, Goecke, Pietik{\"a}inen, and
  Zhao]{huang2018multimodal}
X.~Huang, A.~Dhall, R.~Goecke, M.~Pietik{\"a}inen, and G.~Zhao, ``Multimodal
  framework for analyzing the affect of a group of people,'' \emph{IEEE
  Transactions on Multimedia}, vol.~20, no.~10, pp. 2706--2721, 2018.

\bibitem[Kaliouby and Robinson(2005)]{kaliouby2005real}
R.~e. Kaliouby and P.~Robinson, ``Real-time inference of complex mental states
  from facial expressions and head gestures,'' in \emph{Real-time vision for
  human-computer interaction}.\hskip 1em plus 0.5em minus 0.4em\relax Springer,
  2005, pp. 181--200.

\bibitem[Liu et~al.(2021{\natexlab{d}})Liu, Shi, Chen, Yu, Li, and
  Zhao]{liu2021imigue}
X.~Liu, H.~Shi, H.~Chen, Z.~Yu, X.~Li, and G.~Zhao, ``imigue: An identity-free
  video dataset for micro-gesture understanding and emotion analysis,'' in
  \emph{Proceedings of the IEEE/CVF Conference on Computer Vision and Pattern
  Recognition}, 2021, pp. 10\,631--10\,642.

\end{thebibliography}
}
\vspace{-30pt}
\begin{IEEEbiography}[{\includegraphics[width=1in,height=1in,clip,keepaspectratio]{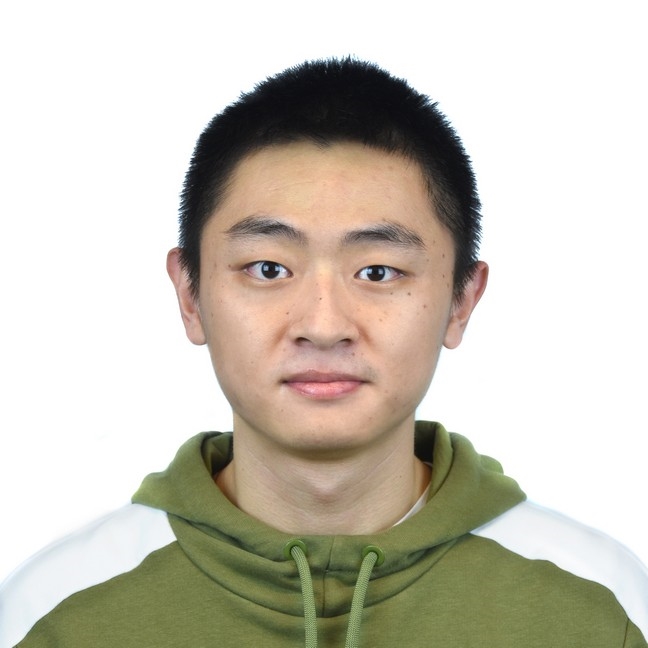}}]{Yang Liu} 
  reveived his Ph.D. degree in computer sciench and technology from the South China University of Technology, in 2021. He is currently a PostDoctoral researcher at the Center for Machine Vision and Signal Analysis, University of Oulu, Finland. His current research interests include facial expression recognition, affective computing, and deep learning.
\end{IEEEbiography}\vspace{-50pt}

\begin{IEEEbiography}[{\includegraphics[width=1in,height=1.25in,clip,keepaspectratio]{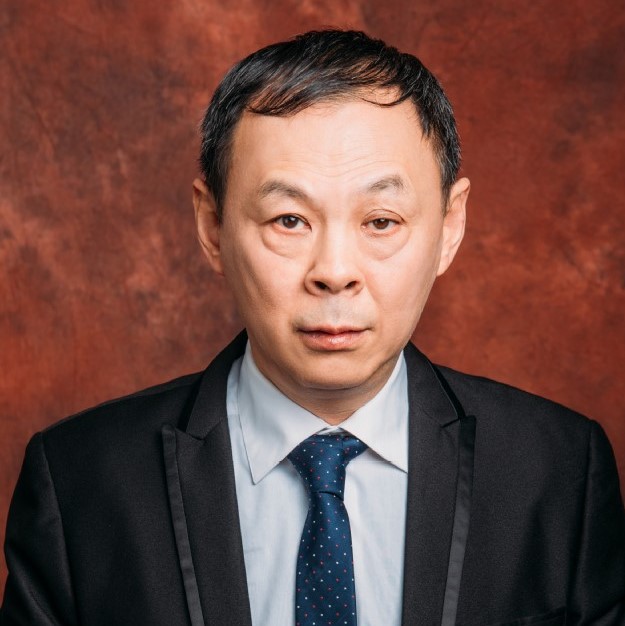}}]{Xingming Zhang}
  is currently a Professor with the School of Computer Science and Engineering, South China University of Technology, Guangzhou, China. He is a member of the Standing Committee of the Education Specialized Committee of China Computer Federation and the Standing director of the University Computer Education Research Association of China. His research focuses on video processing, big data, video surveillance, and face recognition.
\end{IEEEbiography}\vspace{-50pt}

\begin{IEEEbiography}[{\includegraphics[width=1in,height=1.25in,clip,keepaspectratio]{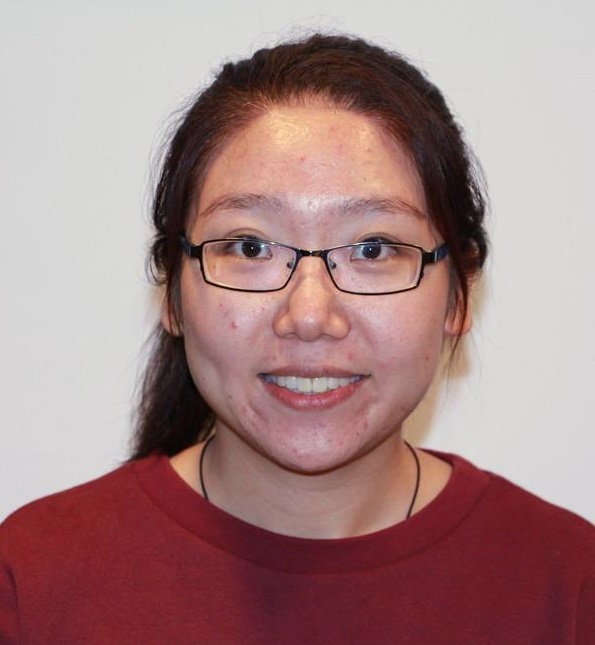}}]{Yante Li}
  received her M.S. degree in computer science and engineering from the China University of Petroleum (East China), in 2017. She is currently pursuing a Ph.D. degree with the University of Oulu, Finland. Her current research interests include micro-expression analysis and facial action unit detection.
\end{IEEEbiography}\vspace{-50pt}

\begin{IEEEbiography}[{\includegraphics[width=1in,height=1.25in,clip,keepaspectratio]{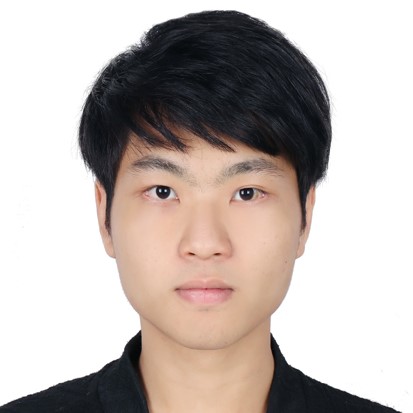}}]{Jinzhao Zhou}
  received his M.S. degree in computer technology from the South China University of Technology, in 2021. He is currently pursuing a Ph.D. at the University of Technology Sydney, Australia. His research interests include affective computing, reinforcement learning, and machine learning.
\end{IEEEbiography}\vspace{-50pt}

\begin{IEEEbiography}[{\includegraphics[width=1in,height=1.25in,clip,keepaspectratio]{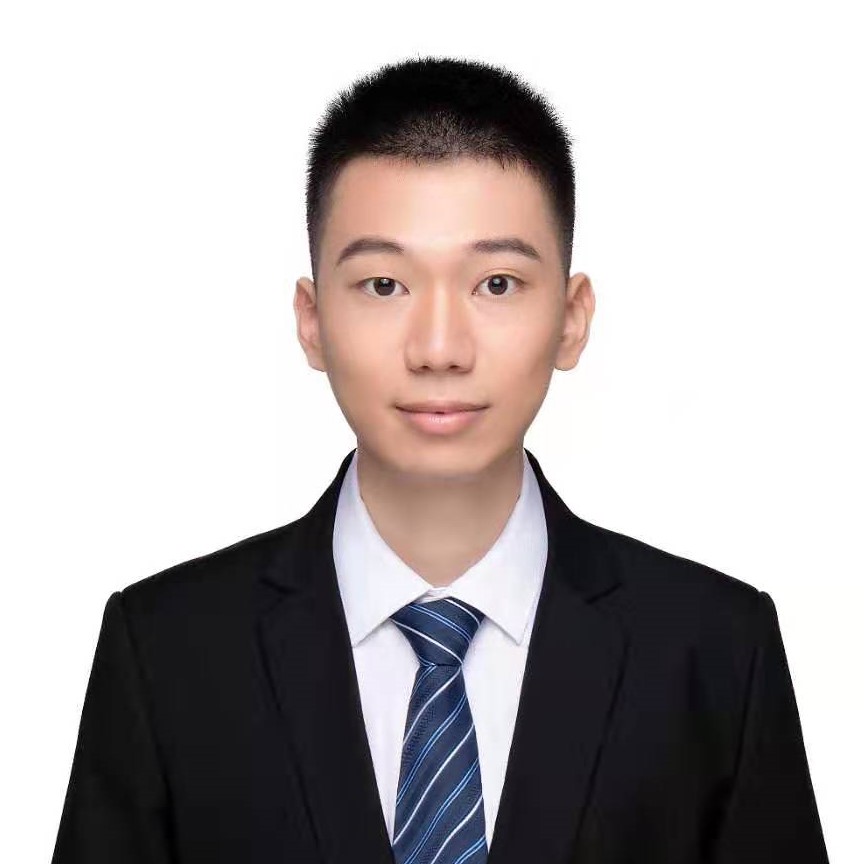}}]{Xin Li}
  received the M.S. degree in computer science and engineering from the South China University of Technology, in 2021. He is currently pursuing his Ph.D. at the Department of Electrical and Computer Engineering, Rutgers University, United States. His research interests include mobile computing and sensing.
\end{IEEEbiography}\vspace{-50pt}

\begin{IEEEbiography}[{\includegraphics[width=1in,height=1.25in,clip,keepaspectratio]{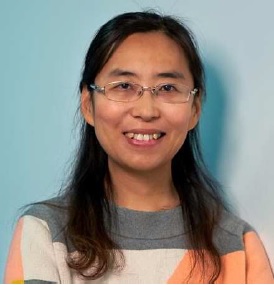}}]{Guoying Zhao}
  is currently an Academy Professor with the University of Oulu, IEEE Fellow, IAPR Fellow, and AAIA Fellow. She has authored or co-authored more than 260 papers in journals and conferences with 17900+ citations in Google Scholar and h-index 65. She is a co-program chair for ICMI 2021, has served as area chair for several conferences, and is an associate editor for Pattern Recognition, IEEE TCSVT, and Image and Vision Computing Journals. Her current research interests include image and video descriptors, facial-expression and micro-expression recognition, emotional gesture analysis, affective computing, and biometrics.
\end{IEEEbiography}
%% insert where needed to balance the two columns on the last page with
%%\newpage


%\vfill


\end{document}